\documentclass[10pt,a4paper,review]{article}

\usepackage{arxiv}

\usepackage[utf8]{inputenc}
\usepackage{amsmath}
\usepackage{amsfonts}
\usepackage{amssymb}
\usepackage{graphicx}
\usepackage{bm}
\usepackage{xcolor}
\usepackage{hyperref}
\usepackage{multirow}
\usepackage{wrapfig}
\usepackage{lscape}
\usepackage{rotating}
\usepackage{epstopdf}
\usepackage[export]{adjustbox}
\usepackage{makecell}

\usepackage{placeins}
\usepackage{physics}
\usepackage{caption}
\usepackage{subfigure}
\usepackage[sort&compress]{natbib}
\DeclareMathOperator*{\argmax}{arg\,max}
\graphicspath{{./figures/}}

\usepackage{xcolor}

\usepackage[utf8]{inputenc}

\title{Model-data-driven Constitutive Responses: Application to a Multiscale Computational Framework
}

\author{
  Jan Niklas Fuhg \\
  Sibley School of Mechanical and Aerospace Engineering \\
  Cornell University, 
   New York, USA \\
  \texttt{jf853@cornell.edu} \\
     \And
  Christoph B\"ohm \\
  Institute of Continuum Mechanics, \\
  Leibniz Universit\"at Hannover, \\ An der Universit\"at 1, 30823 Garbsen, Germany \\
  
   \And
 Nikolaos Bouklas \\
  Sibley School of Mechanical and Aerospace Engineering\\
  Cornell University,
   New York, USA \\
  
  \And
  Am\'{e}lie Fau \\
  Laboratoire de m\'{e}canique et technologie, \'{E}cole normale sup\'{e}rieure Paris-Saclay, \\
4 avenue des Sciences, 91190 Gif-sur-Yvette, France \\

  \And
  Peter Wriggers \\
    Institute of Continuum Mechanics, Leibniz Universit\"at Hannover, \\ An der Universit\"at 1, 30823 Garbsen, Germany \\
    
      \And
      Michele Marino \\
      Department of Civil Engineering and Computer Science, University of Rome Tor Vergata, \\ Via del Politecnico 1, 00133 Rome, Italy.\\
        \texttt{m.marino@ing.uniroma2.it} 
  }

\begin{document}

\maketitle

%\affil[1]{B}

\begin{abstract}
Computational multiscale methods for analyzing and deriving constitutive responses have been used as a tool in engineering problems because of their ability to combine information at different length scales. However, their application in a nonlinear framework can be limited by high computational costs, numerical difficulties, and/or inaccuracies. In this paper, a hybrid methodology is presented which combines classical constitutive laws (model-based), a data-driven correction component, and computational multiscale approaches. A model-based material representation is locally improved with data from lower scales obtained by means of a nonlinear numerical homogenization procedure leading to a model-data-driven approach. Therefore, macroscale simulations explicitly incorporate the true microscale response, maintaining the same level of accuracy that would be obtained with online micro-macro simulations but with a computational cost comparable to classical model-driven approaches. In the proposed approach, both model and data play a fundamental role allowing for the synergistic integration between a physics-based response and a machine learning black-box. Numerical applications are implemented in two dimensions for different tests investigating both  material and structural responses in large deformation.
%
%Moreover, the robustness of model-data-driven numerical simulations is significantly higher than data-driven ones.
%
%Furthermore, a consistent sampling strategy is proposed that enables the enlargement of the considered training space in finite strain setting in an iterative manner. 
%
Overall, the presented model-data-driven methodology proves more versatile and accurate than methods based on classical model-driven as well as pure data-driven techniques. In particular,  a lower number of training samples is required and robustness is higher than for simulations which solely rely only on data. 
\end{abstract}


 \maketitle    
%% maketitle must follow the abstract.
%

\section{Introduction}
Constitutive equations translate observed data into a causal relationship between input and output quantities. In problems with hyperelastic materials, they generally represent a relationship between strains (input) and stresses (output) at each material point of a continuum body. The accuracy of constitutive models plays a major role in the predictive capabilities of numerical simulations, and thus strategies for increasing their effectiveness (computational speed, robustness, ...) is a major open topic in the field. 

Constitutive descriptions in engineering applications are often based on an analytical modelling approach. Hyperelastic stress-strain relationships are obtained from the derivative of a strain energy function defined \emph{ad hoc}. For a number of different materials, huge efforts have been made by the research community in order to find the most suitable expressions which reproduce the available experimental observations. Many phenomenological and micromechanical-based constitutive models are available \cite{Ogden1972,Flory1976,Arruda1993,Boyce2000,Marino2019} (and many more will be developed in the future). These models are widely employed since they are easy to implement in numerical environments, have low computational-cost and are robust from a numerical point of view. Nevertheless, analytical constitutive models  also have limiting drawbacks: the calibration of parameters from fitting of available data might not be trivial; for certain ranges of deformation, the \emph{a priori} fixed mathematical law might result in poor fitting capabilities independent from the choice of model parameters; model parameters might not have a physical meaning and their value might not be related to microstructural properties.

Rather than introducing analytical assumptions on material behavior at the macroscale, computational multiscale models (e.g., numerical homogenization, domain decomposition) are based on the definition of numerical models with a detailed description of material microstructure and on the consistent mapping of microstructural mechanical properties to the macroscale on the basis of numerical simulations conducted at the microscale. These methods provide an unprecedented insight to details of constitutive responses, but are associated with high computational costs and numerical complexities. These approaches are therefore rarely suitable for parametric \emph{in silico} analyses, generally required in the design or the optimization of structural responses.

At the end of the 20th century, data irrupted massively in most scientific fields, leading to the fast emergence of machine learning techniques. The main objective of machine learning is to identify correlations, generally among big data, to describe general trends. In the field of computational materials science, Kirchdoerfer and Ortiz opened the way to the implementation of data-driven simulations \cite{Kirchdoerfer2016,Kirchdoerfer2018}, where constitutive equations are replaced by experimental data that could be possibly noisy \cite{Kirchdoerfer2017,Ayensa2018}. Recently, machine learning algorithms have been adopted in the framework of multiscale homogenization schemes or nonlinear material modelling \cite{Eggersmann2019,Huang2020}. An overview over machine learning solutions to open questions in multiscale modeling has been provided in \cite{peng2020multiscale} where homogenization however is not discussed in much detail. The introduction of data-driven methods for the homogenization of representative volume element has been highly driven by the works of Yvonnet and coworkers \cite{yvonnet2009numerically,le2015computational,lu2019data}. In \cite{yvonnet2009numerically}, a high-order interpolation scheme is employed to obtain a metamodel for homogenization of representative volume elements (RVEs) in small strain mechanical problems in one and two dimensions. The technique is reiterated in later works by \cite{le2015computational} to utilize neural networks for homogenization purposes considering nonlinear elastic media and small strain problems, and later extended in \cite{lu2019data} using the same approach to homogenize the nonlinear anisotropic electrical response of graphene/polymer nanocomposites. The approach proposed in \cite{liu2019deep}, which has been expanded in \cite{liu2019exploring} to three-dimensional problems, proposes a collection of mechanics building blocks which describe analytical homogenization solutions and use neural networks to connect these individual blocks to obtain the complex response of RVEs.  Different to the other reviewed approaches, image-based inputs have for example been used in \cite{yang2018deep}, where convolutional neural networks are employed to infer one effective component of the elastic stiffness tangent from images of multiscale volume elements. Here, the data for the input-output mapping was obtained from small strain finite element simulations.  Despite these remarkable applications, data-driven simulations in the context of material modelling still face severe drawbacks. Firstly, machine learning techniques often require big data, translating into expensive and time-consuming experimental/computational campaigns of tests. Moreover, the number of information required to fully characterize anisotropic materials can massively increase and available tests might not provide a full description. Furthermore, for some applications, material properties might be part of the design variables: microstructural properties are not fixed \emph{a priori} and are varied in the design process seeking for an optimal structural behaviour. In this case, large amounts of data able to span over all possible microstructural realizations are simply not available (or extremely costly to be obtained). In addition, the predicted response can be highly inaccurate and numerically unstable if required at points outside the training space. Before conducting the final simulation, the minimum width of the training space is generally unknown in real applications, requiring concurrent schemes coupling the finite element simulation and machine learning algorithms (given the availability of sufficient reference data). In general, all these drawbacks are inherited by the total lack of physics behind the machine learnt correlation. 

To overcome these issues, the synergistic integration between physics-based knowledge and machine learning models based on data is attracting great interest. Some examples and reviews can be found in \cite{Karpatne2017,Gonzales2019a,Gonzalez2019b,Ibanez2019,Rai2020}. Moreover, drawbacks related to the need of extensive training sets are solved by the emergence of Kriging, also known as Gaussian process regression \cite{kleijnen2009kriging,Lupera2018}. Kriging machine learning algorithms are supervised learning strategies which assume that the interpolated values of the response surface function follow a Gaussian process governed by prior covariances. Kriging approaches provide both a mean predictor of the response function and an indication of the accuracy at each site \cite{Rasmussen2005}. In comparison to other machine-learning techniques, such as Neural Networks \cite{haykin1994neural} or Support Vector Regression \cite{drucker1997support}, Kriging has the advantage of yielding exact interpolation at training points as well as being nonparametric. Furthermore, Kriging allows to capture the model uncertainty and shows better approximation results for smaller datasets. 

In this context, this work presents a hybrid model-data-driven methodology to conduct finite element simulations and its application in a multiscale computational framework.
The proposed approach significant improves the accuracy of model-based constitutive descriptions for the homogenization of RVEs. Compared with advanced computational multiscale schemes, it enables a significant speedup in comparison to FE$^2$ methods, while maintaining the same level of accuracy.  Moreover, it outperforms also classical data-driven schemes.
%
%which will enable signficant speedup in comparison to traditional methods. 
%
The rationale is to combine the response of well-known constitutive equations (model-driven component) with a correction obtained from machine learning techniques (data-driven component). The data-driven component is based on Ordinary Kriging, \cite{kleijnen2009kriging}, obtaining training points by means of a nonlinear computational homogenization procedure. 
%
%A small dataset allows to efficiently combine phenomenological and computational multiscale approaches: an analytical-based constitutive equation can be corrected by means of a data-driven component trained on the basis of few numerical simulations. 
%
The proposed framework is presented in Section \ref{sec::Materials}, including the description of the hybrid approach, a brief overview over the machine learning algorithm, and details on the employed nonlinear homogenization procedure. Applications are described in Section \ref{sec:Applications}, introducing test cases with different complexities. Numerical results are presented and compared with both model-driven and data-driven strategies in Section \ref{sec::resultsAndAppli}, before drawing some conclusions in Section \ref{sec:conclusions}. 

\section{Materials and Methods}\label{sec::Materials}

%The  continuum-mechanical framework presented in what follows addresses the description of macroscale responses. 

The domain of a continuum body in the reference configuration is represented by $\Omega_o$. The body is loaded by means of body forces ${\bf b}$ within $\Omega_o$ and boundary tractions ${\bf t}$ on part $\partial \Omega_o^t$ of the entire boundary $\partial \Omega_o$, i.e. $\partial \Omega_o^t\subseteq \partial \Omega_o$. The resulting displacement field ${\bf u}$ of each material point respects boundary conditions ${\bf u} = \bar{\bf u}$ on part $\partial \Omega_o^u = \partial \Omega_o \setminus \partial \Omega_o^t$. The deformation map is locally described in terms of the deformation gradient ${\bf F}= {\bf I} + \text{Grad}({\bf u})$. 
%
%When represented in a Cartesian coordinate system, the tensor component $(i,j)$ is denoted by $[ \, \bullet \,]_{ij}$.
%
The present work addresses the modelling of a hyperelastic material response, described by a strain-energy density function $\Psi_{el}({\bf C})$, with ${\bf C}={\bf F}^T{\bf F}$ being the right Cauchy-Green deformation tensor. 

%The macroscale response will be coupled with a lower scale (micro) description, which will be later introduced also in the framework of continuum mechanics.

\subsection{Model-data hybrid strain energy density}

In the model-data-driven approach, the strain-energy density function $\Psi_{el}$ is supposed to be unknown and will not be postulated \emph{ab initio}. However, a modelling component $\Psi_{mod}$, based on an analytical constitutive law, is introduced as an approximation of $\Psi_{el}$. Accounting for the (inevitable) remainder $\Psi_{rem} = \Psi_{el} - \Psi_{mod}$, the constitutive response between ${\bf C}$ and the second Piola-Kirchhoff stress tensor ${\bf S}$ reads:
\begin{equation} \label{eq:S}
{\bf C} \mapsto {\bf S} = \frac{\partial \Psi_{el}}{\partial {\bf E}} = 2 \frac{\partial (\Psi_{mod} +  \Psi_{rem})}{\partial {\bf C}} = {\bf S}_{mod}({\bf C}) + {\bf S}_{rem}({\bf C}) \, ,
\end{equation}
where ${\bf E}=({\bf C}-{\bf I})/2$ is the Green-Lagrange strain tensor and:
\begin{equation} 
 {\bf S}_{mod} =  2 \frac{\partial  \Psi_{mod}}{\partial {\bf C}} \, , \qquad {\bf S}_{rem} =  2 \frac{\partial  \Psi_{rem}}{\partial {\bf C}}\, .
\end{equation}
In other words, the modelling stress component ${\bf S}_{mod} $ is corrected by a component ${\bf S}_{rem}$. The functional expression of ${\bf S}_{rem}({\bf C})$ will not be obtained from the \emph{a priori} definition of a strain-energy component $\Psi_{rem}$, but from multiscale information of the micromechanical response.

%, but from a machine learnt function ${\bf C} \mapsto {\bf S}_{rem}$. 

Function ${\bf S}({\bf C})$ enters in the balance of linear momentum, which in weak form reads
\begin{equation} \label{eq:weak}
G({\bf u}, \delta{\bf u}) = \int_{\Omega_o} {\bf S}({\bf C}) : \delta{\bf E} \, dV - \int_{\Omega_o}  {\bf b}\cdot \delta{\bf u} \, dV - \int_{\partial \Omega_o^t} {\bf t} \cdot \delta{\bf u} \, dA = 0 \quad \forall \, \delta{\bf u} \in \mathcal{U}\, .
\end{equation}
Here, $\delta{\bf E}= ({\bf F}^T \, \text{Grad}(\delta {\bf u})+\text{Grad}^T(\delta {\bf u}) {\bf F})/2$ is the Green-Lagrange strain tensor associated with increment $\delta{\bf u}$ of the displacement field ${\bf u}$ and and $\mathcal{U}=\{ {\bf v} \in C^0(\Omega_o)\, \text{s.t.} \, {\bf v}=\bar{\bf u} \text{ on }  \partial \Omega_o^u\}$ is the set of compatible displacements.

For the solution of the set of nonlinear equations derived from Eq. \eqref{eq:weak}, iterative solution schemes based on the Newton's method are generally employed. At each iteration, an improved solution is obtained from the Taylor series expansion of the nonlinear equation at the already computed approximate solution $\bar{\bf u}$. Taylor expansion corresponds to the linearization $\Delta G$ of the weak form around $\bar{\bf u}$ along a displacement incremental direction $\Delta {\bf u}$, that is:
\begin{equation} \label{eq:linear}
\Delta G(\bar{\bf u}, \delta{\bf u}) \cdot \Delta {\bf u} =  \int_{\Omega_o} (\text{Grad} (\Delta {\bf u}) \, \bar{\bf S} : \text{Grad}(\delta {\bf u}) + \delta\bar{\bf E} :  \bar{\mathbb{D}} : \Delta \bar{\bf E}) \, dV= 0\, ,
\end{equation}
where $\bar{\bf S}={\bf S}(\bar{\bf C})$ with $\bar{\bf C}=\bar{\bf F}^T\bar{\bf F}$ the right Cauchy-Green deformation tensor associated with $\bar{\bf F} = \text{Grad}(\bar{\bf u})$ at the attempt solution\begin{footnote}{In Eq. \eqref{eq:linear}, the variations of the
Green-Lagrange strain tensor result in $\delta{\bf E}=(\bar{\bf F}^T \, \text{Grad}(\delta {\bf u})+\text{Grad}^T(\delta {\bf u}) \bar{\bf F})/2$  and $\Delta{\bf E}=(\bar{\bf F}^T \, \text{Grad}(\Delta {\bf u})+\text{Grad}^T(\Delta {\bf u}) \bar{\bf F})/2$. }\end{footnote}. Moreover, $\bar{\mathbb{D}}=\mathbb{D}(\bar{\bf C})$ is the fourth-order stiffness tensor (referred to the reference configuration $\Omega_o$) computed at ${\bf C} = \bar{\bf C}$, with:
\begin{equation} \label{eq:D}
{\bf C} \mapsto \mathbb{D} =  \frac{\partial {\bf S}}{\partial {\bf E}} = 4 \frac{\partial^2 \Psi_{el}}{\partial {\bf C} \partial {\bf C}} = \mathbb{D}_{mod}({\bf C}) + \mathbb{D}_{rem}({\bf C})\, ,
\end{equation}
where:
\begin{equation}
\mathbb{D}_{mod} = 4 \frac{\partial^2 \Psi_{mod}}{\partial {\bf C} \partial {\bf C}}\, ,\qquad \mathbb{D}_{rem} = 4 \frac{\partial^2 \Psi_{rem}}{\partial {\bf C} \partial {\bf C}}\, .
\end{equation}
Analogously to stresses, %a machine learnt function ${\bf C} \mapsto \mathbb{D}_{rem}$ 
the remainder term ${\bf C} \mapsto \mathbb{D}_{rem}$  defines the correction to the modelling component $\mathbb{D}_{mod}$, allowing for the solution of nonlinear \mbox{con\-ti\-nu\-um} mechanical problems and being strictly related to the multiscale behaviour of material micromechanics. Equilibrium balance laws which incorporate the true material micromechanical response can be thus numerically solved only by completing Eqs. \eqref{eq:S} and \eqref{eq:D} with the knowledge on the correction terms ${\bf S}_{rem}({\bf C})$ and $\mathbb{D}_{rem}({\bf C})$. 

\begin{figure}[tb]
\centering
\includegraphics[width=0.9\textwidth]{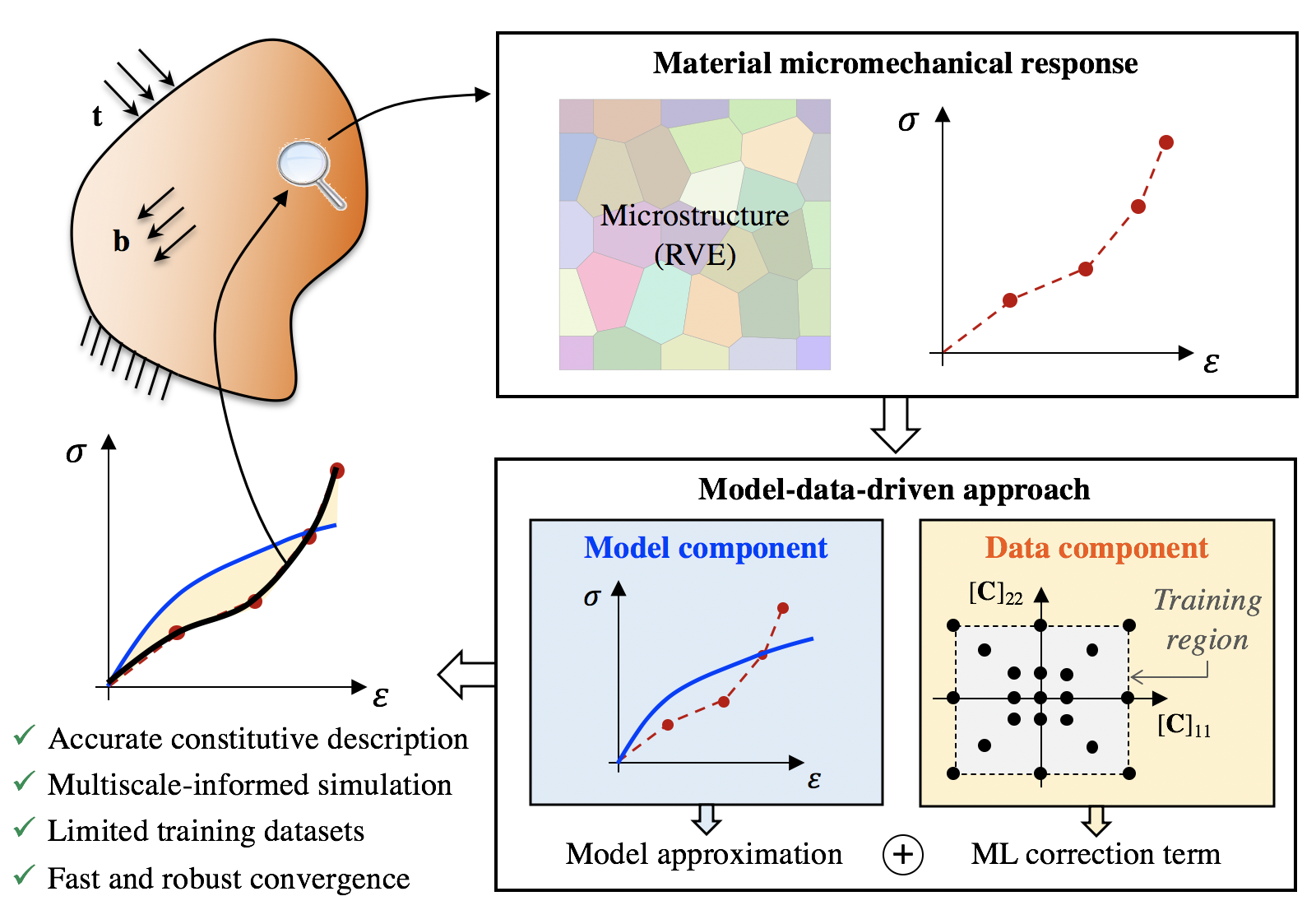}
\caption{Model-data-driven modelling strategy. Acronyms: RVE, Representative Volume Element; ML, Machine-learning.}\label{fig:intro}
\end{figure}

The model-data-driven approach, schematically summarized in Fig. \ref{fig:intro}, is built upon the knowledge of coupled values for $({\bf C},{\bf S})$ and $({\bf C},\mathbb{D})$ which represent the ground truth of the micromechanical response. This information can be obtained either from a series of experimental tests on material specimens or from a campaign of numerical simulations on representative volume elements of the microstructure. The micromechanical material response is firstly employed to conceive and calibrate the model component $\Psi_{mod}({\bf C})$ which provides an approximated constitutive response. Given ${\bf S}_{mod}({\bf C})$ and $\mathbb{D}_{mod}({\bf C})$, the remainder terms ${\bf S}_{rem}({\bf C})$ and $\mathbb{D}_{rem}({\bf C})$ are then obtained by adopting a machine-learning technique based on Kriging (see Section \ref{sec:data}), here coupled with numerical homogenization based on the finite element method (see Section \ref{sec::datasetDiscussion}). Accordingly, in the limits of the finite element approximation, the surrogate function $\mathbb{D}_{rem}({\bf C})$ will represent the exact material tangent (i.e., the true derivative of ${\bf S}_{rem}({\bf C})$) only at training points (where Kriging returns exact interpolation), while it represents an approximated relationship between the training samples.

\subsection{Data-driven component: machine learning approach} \label{sec:data}

Exploiting the symmetry of ${\bf C}$, ${\bf S}_{rem}$ and $\mathbb{D}_{rem}$, an unknown relationship is sought between 6 input values, collected in the Voigt vector representation ${\bf c}=\text{vec}({\bf C}) \in \mathbb{R}^6$ of ${\bf C}$ , and 27 output values, collected in vector ${\bf y}_{rem} \in \mathbb{R}^{27}$,
\begin{equation}
{\bf y}_{rem} = \begin{pmatrix}{\bf s}_{rem} \\ {\bf d}_{rem} \end{pmatrix}\, , 
\end{equation}
where ${\bf s}_{rem}=\text{vec}({\bf S}_{rem})\in \mathbb{R}^6$ is the Voigt vector representation of ${\bf S}_{rem}$ and ${\bf d}_{rem}=\text{vec}(\mathbb{D}_{rem})\in \mathbb{R}^{21}$ a vector collecting the independent components in $\mathbb{D}_{rem}$.

Function ${\bf c} \mapsto {\bf y}_{rem}$ is built from the knowledge of the actual constitutive response at a finite set of  $n_{tp}$ training points, associated with different values ${\bf c}_i = \text{vec}({\bf C}_i)$ where $i=1, \, \ldots  , \, n_{tp}$. Therefore, values ${\bf y}_{rem,i}={\bf y}_{rem}({\bf c}_i)$ follow:
\begin{equation}
{\bf S}_{rem}({\bf C}_i)  = {\bf S}({\bf C}_i) - {\bf S}_{mod}({\bf C}_i)\, , \quad \mathbb{D}_{rem}({\bf C}_i)  =  \mathbb{D}({\bf C}_i) -  \mathbb{D}_{mod}({\bf C}_i)\, ,
\end{equation}
since $ {\bf S}({\bf C}_i)$ and $\mathbb{D}({\bf C}_i)$ are known, and ${\bf S}_{mod}({\bf C}_i)$ and $\mathbb{D}_{mod}({\bf C}_i)$ are computable. This yields the training dataset $\mathcal{D}$ for the determination of the data-driven component of the proposed approach:
\begin{equation} \label{eq:Dset}
\mathcal{D} = \lbrace \left( {\bf c}_{i}, \, {\bf y}_{rem,i} \right), \, i=1, \, \ldots  , \, n_{tp}  \rbrace \, .
\end{equation}

In this work, Ordinary Kriging has been employed to approximate the mapping ${\bf c} \mapsto {\bf y}_{rem}$ from $\mathcal{D}$. For the sake of completeness, Ordinary Kriging is briefly summarized next, with proofs and further details that can be found in  \cite{santner2003design}.

\subsubsection{Ordinary Kriging} \label{sec:kriging}

Consider a black-box function $\mathcal{M}: \mathbb{X} \rightarrow \mathbb{Y}$ between the input ${\bf c} \in \mathbb{X} \subseteq \mathbb{R}^{n_{x}}$ and the output ${\bf y}_{rem} \in \mathbb{Y} \subseteq \mathbb{R}^{n_{y}}$. The training dataset consists of $n_{tp}$ data points, and let ${\bf y}_{rem}^{tp}$ be the associated $n_{y}n_{tp} \times 1$ stacked vector of outputs, i.e.
\begin{equation}
{\bf y}_{rem}^{tp}=\begin{bmatrix} {\bf y}_{rem,1} \\ \vdots \\ {\bf y}_{rem,n_{tp}} \end{bmatrix}\, .
\end{equation}

%${\bf y}_{rem}^{n_{y},n_{tr}}$ denote the .
 
The main assumption for the approximation of this mapping with Kriging is that the deterministic output is regarded as a realization of a stochastic process:
\begin{equation}\label{eq:Gauss_general}
{\bf Y}(\bf c) = \bm{\mu} + {\bf A} {\bf Z}\, ,
\end{equation}
with ${\bf Y}$ the $n_{y}$-dimensional output vector, $\bm{\mu}$ the $n_{y}$-dimensional vector representing the mean of the Gaussian process, ${\bf A}$ a $(n_{y}\times n_{y})$ positive-definite matrix (the first set of unknown parameters), and ${\bf Z}$ a $n_{y}$-dimensional vector of mutually independent Gaussian processes \citep{svenson2010multiobjective}. 
A required preliminary ingredient for the definition of a Kriging metamodel is a user-chosen autocorrelation function $R$ between two inputs ${\bf c}$ and ${\bf c}'$, here introduced as the Mat\'{e}rn 3/2 function  of the form \citep{matern1960spatial}:
\begin{equation}\label{eq:autocorr}
\begin{aligned}
R({\bf c}, {\bf c}', \bm{\theta}_{i})  =  \prod_{k=1}^{n_{x}}  \left( 1 + \dfrac{\sqrt{3} \abs{c_{k} - c'_{k}}}{\theta_{i,k}} \right) \exp \left(-\dfrac{\sqrt{3} \abs{c_{k} - c'_{k}} }{\theta_{i,k}}  \right) \, \text{,}
\end{aligned}
\end{equation}
where $c_k$ and $c'_k$ are the $k$-th component of ${\bf c}$ and ${\bf c}'$ (with $k=1,\ldots,n_x$), and $\boldsymbol{\theta}=\lbrace \boldsymbol{\theta}_{1}, \ldots, \boldsymbol{\theta}_{n_{y}} \rbrace$ is a vector where each $\boldsymbol{\theta}_{i}$ collects $n_x$ hyperparameters $\theta_{i,1}, \ldots, \theta_{i,n_x}$ (the second set of unknown parameters). The covariance of the process ${\bf Y}$ evaluated at some input values is given by
\begin{equation}
Cov({\bf Y}({\bf c}), {\bf Y}({\bf{c}}')) = {\bf A} {\bf R}({\bf c},{\bf c}'){\bf A}^{T}
\end{equation}
where  ${\bf R} \in \mathbb{R}^{n_{y}\times n_{y}}$ given by 
\begin{equation}
{\bf R}({\bf c},{\bf c}') = \text{diag}\lbrace R({\bf c}, {\bf c}', \boldsymbol{\theta}_{1}), \cdots  , R({\bf c}, {\bf c}', \boldsymbol{\theta}_{n_{y}}) \rbrace\, .
\end{equation}
When the corresponding input strains are equivalent the $(n_{y}\times n_{y})$ matrix
\begin{equation}
{\bf \Sigma}_{0}({\bf c}) = Cov({\bf Y}({\bf c}), {\bf Y}({\bf{c}}))
\end{equation}
can be defined.
The covariance matrix $\boldsymbol{\Sigma}$ of the process is a $(n_{y}n_{tp} \times n_{y}n_{tp})$ matrix given block-component-wise by
\begin{equation}
[\boldsymbol{\Sigma}]_{ij} = \begin{cases}
{\bf \Sigma}_{0}({\bf c}_i) & \text{for } i=j\, , \\
Cov({\bf Y}({\bf c}_{i}), {\bf Y}({\bf{c}}_{j})) & \text{else}\, .
\end{cases}
%
%\Sigma_{n_{y},n_{tr}} = \begin{bmatrix}
%\Sigma_{0}({\bf c}_1) & Cov({\bf Y}({\bf c}_{1}), {\bf Y}({\bf{c}}_{2})) &  \cdots  & Cov({\bf Y}({\bf{c}}_{1}), {\bf Y}({\bf c}_{n_{tr}})) \\
%Cov({\bf Y}({\bf c}_{1}), {\bf Y}({\bf{c}}_{2})) & \Sigma_{0}(({\bf c}_2) & \cdots & Cov({\bf Y}({\bf{c}}_{2}), {\bf Y}({\bf c}_{n_{tr}})) \\
%\vdots & \vdots & \ddots & \vdots \\
%Cov({\bf Y}({\bf c}_{1}), {\bf Y}({\bf{c}}_{n_{tr}}))  & Cov({\bf Y}({\bf{c}}_{2}), {\bf Y}({\bf c}_{n_{tr}})) & \cdots & \Sigma_{0}({\bf c}_{n_{tr}}) \\
%\end{bmatrix}.
\end{equation}

By using multivariate Ordinary Kriging, the mapping between any point ${\bf c}_{\star} \in \mathbb{X}$, in general not belonging to the training dataset (i.e., ${\bf c}_{\star} \neq {\bf c}_1, \ldots , {\bf c}_{n_{tr}}$), and the corresponding output value ${\bf y}_{rem}({\bf c}_{\star})$ is approximated by a generalized least squares estimator with mean:
\begin{equation}\label{eq:mean}
\begin{aligned}
\hat{{\bf y}}_{rem}({\bf c}_{\star}) =  \hat{\bm{\mu}} + \boldsymbol{{\Pi}}({\bf c}_{\star}) \boldsymbol{\Sigma} \big({\bf y}_{rem}^{tp}- {\bf \mathcal{F}} \hat{\bm{\mu}} \big),
\end{aligned}
\end{equation}
where $\boldsymbol{{\Pi}} \in \mathbb{R}^{n_{y} \times n_{y} n_{tp}}$ and $\hat{\bm{\mu}} \in \mathbb{R}^{n_{y}}$ are given by: 
\begin{subequations}
\begin{align}
& \boldsymbol{{\Pi}}({\bf c}_{\star}) = \begin{bmatrix}
Cov({\bf Y}({\bf c}_{\star}), {\bf Y}({\bf c}_{1})) & \cdots & Cov({\bf Y}({\bf c}_{\star}), {\bf Y}({\bf c}_{n_{tr}})) \end{bmatrix}\, ,\\
& \hat{\bm{\mu}}= ({\bf \mathcal{F}}^{T} \boldsymbol{\Sigma}^{-1} {\bf F}_{n_{y}})^{-1} {\bf \mathcal{F}}^{T} \boldsymbol{\Sigma}^{-1} {\bf y}_{rem}^{tp}\, ,
\end{align}
\end{subequations}
with ${\bf{\mathcal{F}}} = \bm{1}_{n_{tp}} \otimes {\bf{I}}_{n_{y}}$ obtained
%\begin{footnote}{Matrix $ {\bf \mathcal{F}}$ shall not be confused with the deformation gradient ${\bf F}$.}\end{footnote} 
%
from  a $n_{tp}$-dimensional vector of ones $\bm{1}_{n_{tp}}$ and a $(n_{y} \times n_{y})$ unit matrix ${\bf{I}}_{n_{y}}$ through the Kronecker operator $\otimes$ yielding ${\bf{\mathcal{F}}} \in \mathbb{R}^{n_{y} n_{tp} \times n_{y}}$.

The estimated prediction uncertainty matrix of dimension $(n_{y} \times n_{y})$ at a point $\bf{c}_{\star}$ is given by:
\begin{equation}\label{eq:var}
\begin{aligned}
{\bf{\Upsilon}}(\bf{c}_{\star}) =& {\bf \Sigma}_{0}({\bf c}_{\star}) - \boldsymbol{{\Pi}}({\bf c}_{\star}) \boldsymbol{\Sigma}^{-1} \boldsymbol{{\Pi}}({\bf c}_{\star})^{T}  + \\
& + {\bf B}({\bf c}_{\star})\times ({\bf \mathcal{F}}^{T} \boldsymbol{\Sigma}^{-1} {\bf \mathcal{F}})^{-1} ({\bf B}({\bf c}_{\star}))^{T},
\end{aligned}
\end{equation}
with:
\begin{equation}
{\bf B}({\bf c}_{\star}) = {\bf I}_{n_{y}} - \boldsymbol{{\Pi}}({\bf c}_{\star}) \boldsymbol{\Sigma}^{-1} {\bf \mathcal{F}}\, .
\end{equation}

Accordingly, the predicted value of function ${\bf c} \mapsto {\bf y}_{rem}$ at any point ${\bf c}  \in \mathbb{X}$ lies in the $95\%$ confidence interval of the Gaussian random multivariate metamodel $\tilde{\bf y}_{rem}({\bf c})$, that is \citep{kleijnen2014multivariate}:
\begin{equation} \label{eq:y_meta}
{\bf y}_{rem}({\bf c}) \approx \tilde{\bf y}_{rem}({\bf c}) \in \hat{\bf y}_{rem}({\bf c}) \pm 1.96 \, \sqrt{\text{diag}\left( \bf{\Upsilon}(\bf{c}_{\star}) \right)}\, .
\end{equation}
%
%If not specifically indicated, the metamodel will be built in what follows based on the mean value, that is $a({\bf c}) \approx \tilde{a}({\bf c}) = \hat{a}({\bf c})$. 
%

It is noteworthy that, from Eqs. \eqref{eq:mean} and  \eqref{eq:var}, the training dataset points are exactly interpolated by the mean value of the predictor (i.e., $\hat{{\bf y}}_{rem}({\bf c}_{i}) = {\bf y}_{rem, i}$) and zero variance is obtained at the training points (i.e., ${\bf{\Upsilon}}({\bf c}_{i}) = \bm{0}$). Nevertheless, the metamodel in Eq. \eqref{eq:y_meta} depends on two set of unknown parameters ${\bf A}$ and ${\bm{\theta}}$, the determination of which represents the building of the machine learning component. The value of ${\bf A}$ and ${\bm{\theta}}$ can be estimated using a restricted maximum likelihood approach, which defines the optimization problem:
\begin{equation} \label{eq:optim_ML}
\begin{aligned}
\{\hat{{\bf A}}, \hat{{\bm{\theta}}}\} = \argmax_{{\bf A}^{\star}, {\bm{\theta}}^{\star}} &\left[-\frac{1}{4} \log(|\boldsymbol{\Sigma}|) \log({\bf \mathcal{F}}^{T} \boldsymbol{\Sigma}^{-1} {\bf \mathcal{F}}) + \right. \\
&\left.-\frac{1}{2} ({\bf y}_{rem}^{tp} - {\bf \mathcal{F}} \hat{\bm{\mu}})^{T} \boldsymbol{\Sigma}^{-1} ({\bf y}_{rem}^{tp} - {\bf \mathcal{F}} \hat{\bm{\mu}}) \right]\, .
\end{aligned}
\end{equation}
This optimization problem is solved numerically in MATLAB by employing a hybridized particle swarm optimization approach adopted from \cite{toal2011development}.

As a side note, Ordinary Kriging is chosen to map the strain input independently to the respective stress and material tangent outputs. Efforts to include the material tangent information as direct derivative information utilizing for example Gradient-Enhanced Kriging (GEK) \citep{bernardo1992some,solak2002derivative}, as e.g. employed in \cite{rocha2020fly} have been delibarately avoided. This is due to the fact that the chosen approach should scale well with an increasing number of sample points which is a major concern in large strain mechanical problems since the strain input range for which a model has to be trained may theoretically be significant, see Section \ref{sec::datasetDiscussion} for a discussion. Kriging surrogate models with gradient information suffer from dimensionality issues when the training data set gets too large since the size of the correlation matrix for single output GEK, is increased from $(n_{tp} \times n_{tp})$ to $(n_{tp} (n_{x}+1) \times n_{tp} (n_{x}+1))$ compared to Ordinary Kriging where $n_{x}$ is the input dimension. Even though efforts to make GEK utilizeable for large datasets have recently been intensified \cite{bouhlel2019gradient,chen2019screening}, Ordinary Krgiging remains one of the most well known (most fundamental) versions of Gaussian process regression for which a multitude of open-source frameworks are available, see e.g. the DACE toolbox \cite{lophaven2002dace} or PyKrige \cite{pykrige2021}, making the presented approach easily accesible.

\subsection{Training dataset: numerical experiments}\label{sec::datasetDiscussion}
 
The first fundamental ingredient of any data-driven approach is the definition of the training region $\mathcal{R}_{tr}$, that is the smallest convex envelope in the space of right Cauchy-Green deformation tensors ${\bf C}$ such that each sample point ${\bf c}_i$ lies within $\mathcal{R}_{tr}$ (with $i=1,\ldots, n_{tp}$). The training region $\mathcal{R}_{tr}$ is here defined as a hypercube around the reference (unstressed) configuration at ${\bf F} = {\bf I}$. Input training points ${\bf c}_{inp} = \text{vec}({\bf C}_{inp})$ are created from the definition of input deformation gradient tensors ${\bf F}_{inp}$, thus ensuring the positive definiteness of the introduced ${\bf C}_{inp}$. 
%
%. This choice is motivated by the fact that the definition of the components of ${\bf C}$ from ${\bf F}$ ensures the positive definiteness of the resulting tensor. 
%
In passing, it is also noted that the physical meaning of the components of ${\bf F}$ is easier to interpret than for ${\bf C}$, and thereby the training region of the metamodel as well.
%%
%\begin{enumerate}    
% %       \item  the choice ${\bf F}={\bf F}^T={\bf U}$ corresponds to assume null rigid rotations and, due to objectivity requirements, it does not affect the constitutive relationship and the training procedure (i.e., ${\bf C}_{inp} = {\bf F}_{inp}^T{\bf F}_{inp}={\bf U}_{inp}^T{\bf U}_{inp}$);
%        
%            \item the definition of the components of ${\bf C}$ from ${\bf F}$ ensures the positive definiteness of the resulting tensor;
%        
%    \item the physical meaning of the components of ${\bf F}$ is easier to interpret than for ${\bf C}$, and thereby the training region of the  metamodel as well.
%\end{enumerate}
%

A specific training region is characterized by the maximum deviation $\Delta_{tr}$ of any component of each ${\bf F}_{inp}$ from the identity tensor ${\bf I}$. Therefore, $\Delta_{tr}$ represents the stretch-half-width of the training region. 
%
%Denoting by ${\bf f}$ the vector collecting the independent components within ${\bf F}$, a specific training region is defined by the maximum deviation $\Delta_{tr}$ of each component of each ${\bf f}_{inp}$ from ${\bf f}_0=\text{vec}({\bf I})$. Therefore, $\Delta_{tr}$ represents the stretch-half-width of the training region. 
%For example, in plane strain applications, a value $\Delta_{tr}=0.1$ corresponds to a training region with $\mathbb{X} \equiv F_{11} \times F_{22} \times F_{12} \in [0.9,1.1]\times [0.9,1.1] \times [-0.1,0.1]$. In other words, 
%
%In addition to $\Delta_{tr}$, 
Two additional choices have to be made for defining the training set: the number and position of sampling points within $\mathcal{R}_{tr}$. Data-driven approaches are highly dependent on these features, which in fact depend on the application at hand. For facing this issue, Section \ref{par:specs_hybrid} will provide a systematic procedure, which will be shown to work effectively for the applications addressed in this work.

Once input points are defined, the next ingredient of a data-driven strategy is the definition of the procedure for creating dataset couples ${\bf C} \mapsto \{{\bf S}_{rem}, \, \mathbb{D}_{rem}\}$. In particular, a direct computational homogenization approach of a microstructural representative volume element is employed in this work. This strategy allows to embed the proposed model-data-driven approach within a multiscale framework for the characterization of material responses, providing a general procedure which can be used in a wide number of application cases.

 \subsubsection{Computational homogenization procedure} \label{par:appl_comp}

The true material response is obtained from a numerical homogenization procedure based on the introduction of a representative volume element (RVE), also treated in the framework of continuum mechanics. The domain of the RVE in the reference configuration is denoted by $\tilde{\Omega}_o$ and is characterized by a characteristic dimension $\ell$ significantly lower than the one $L$ of the macroscale continuum body defined on $\Omega_o$. Hereby, an appropriate scale-separation is pre-assumed, that is the RVE-level (the microscale) is attached to length scales significantly lower than the continuum body one (the macroscale), with a RVE attached to each material point ${\bf X} \in \Omega_o$.

The Hill–Mandel condition\begin{footnote}{The local variation of the virtual work computed in a material point at the macroscale is equal to the volume average of the variation of the virtual work performed on the RVE.}\end{footnote} is enforced as bridge condition for crossing scales. Denoting with $\tilde{V}$ the total RVE volume and with $\tilde{\bf X} \in \tilde{\Omega}_o$ a material point in the RVE reference configuration, let ${\bf F}_{avg}$ and ${\bf P}_{avg}$ be respectively the average of the stretch tensor $\tilde{\bf F}(\tilde{\bf X})$ and of the first Piola-Kirchhoff (1-PK) stress tensor $\tilde{\bf P}(\tilde{\bf X})$ within the RVE at the microscale, i.e.
\begin{equation} \label{eq:averageRVE}
{\bf F}_{avg}= \frac{1}{\tilde{V}}\int_{\tilde{\Omega}_o} \tilde{\bf F}(\tilde{\bf X}) \, d\Omega \, , \qquad {\bf P}_{avg}= \frac{1}{\tilde{V}}\int_{\tilde{\Omega}_o} \tilde{\bf P}(\tilde{\bf X}) \, d\Omega\, .
\end{equation}

The RVE is numerically tested through finite-element simulations by considering a series of elastic equilibrium problems with homogeneous displacement-type boundary conditions on the entire boundary $\partial \tilde{\Omega}_o$, \cite{Geers2017}, that is
\begin{equation} \label{eq:ubc}
\tilde{\bf u}_{bc} = ({\bf F}_{app}-{\bf I})\tilde{\bf X} \qquad  \forall \; \tilde{\bf X} \in \partial \tilde{\Omega}_o\, ,
\end{equation}
with ${\bf F}_{app}$ being the applied deformation.
%
%${\bf U}_{app} = {\bf U}_{app}(F_{11},F_{22},F_{12})$ defined as:
%%
%\begin{equation}
%{\bf U}_{app} = {\bf I} + (F_{11}-1) ({\bf e}_1 \otimes {\bf E}_1) + (F_{22}-1) ({\bf e}_2 \otimes {\bf E}_2)+ F_{12} ({\bf e}_1 \otimes {\bf E}_2+ {\bf e}_2 \otimes {\bf E}_1) \, ,
%\end{equation}
%%
%resulting ${\bf U}_{app}={\bf U}_{app}^T$.
%%
%such that input training points within $\mathcal{R}_{tr}$ and assigned.
%
%
With boundary conditions as in Eq. \eqref{eq:ubc}, the average of the deformation gradient ${\bf F}_{avg}$ within the RVE is equal to the applied stretch tensor ${\bf F}_{app}$, that is  ${\bf F}_{avg} = {\bf F}_{app}$,  \cite{Geers2017}. Following the Hill–Mandel condition, the micro-macro scale transition is then given by:
\begin{equation} \label{eq:micro-macro-trans}
{\bf F}={\bf F}_{avg} \quad  \text{ and } \quad {\bf P}={\bf P}_{avg}\, ,
\end{equation}
which ensures the equivalence between the average of the virtual work performed by variation of microscopic deformations on the RVE, i.e. $\delta\tilde{\bf F}$, and the virtual work performed by variation of macroscopic deformations, i.e. $\delta {\bf F}$:
\begin{equation}
\frac{1}{\tilde{V}} \int_{\tilde{\Omega}_o} \tilde{\bf P} \, : \, \delta \tilde{\bf F}  \, d\Omega =  {\bf P} \, : \, \delta {\bf F} \, .
\end{equation}
The RVE stress-strain relationship is described from Eq. \eqref{eq:averageRVE}  by defining the relationship between the macroscopic right Cauchy-Green deformation tensor (corresponding to ${\bf F}_{avg}$) and the second Piola-Kirchhoff (2-PK) stress tensor (computed from ${\bf F}_{avg}$ and ${\bf P}_{avg}$), that is:
\begin{equation} \label{eq:C-S_micro}
{\bf C}={\bf F}_{avg}^T{\bf F}_{avg} \mapsto {\bf S}_{\text{RVE}}=({\bf F}_{avg})^{-1}{\bf P}_{avg}\, .
\end{equation}

Therefore, the constitutive relationship ${\bf C}\mapsto {\bf S}$ is obtained from the solution of an equilibrium problem with boundary conditions as in Eq. \eqref{eq:ubc}. This is solved by means of finite element method (FEM). In order to provide a complete dataset $\mathcal{D}$,  the stress ${\bf S}_{\text{RVE}}$ obtained form the finite element computation is linearized with respect to the macroscopic deformation ${\bf C}$ to estimate the RVE fourth-order material tangent tensor $\mathbb{D}_{\text{RVE}}$ (cf., Eq. \eqref{eq:D}):
\begin{equation} \label{eq:Drve}
\mathbb{D}_{\text{RVE}}=2\frac{\partial {\bf S}_{\text{RVE}}}{\partial {\bf C}} \, , \quad [\mathbb{D}_{\text{RVE}}]_{AJBL} = D_{AJBL}\, .
\end{equation}
The latter represents the second elasticity tensor and it is obtained from the first elasticity tensor, namely
\begin{equation} \label{eq:Arve}
\mathbb{A}_{\text{RVE}}=\frac{\partial {\bf P}_{avg}}{\partial {\bf F}_{avg}} \, , \qquad [\mathbb{A}_{\text{RVE}}]_{iJkL} = A_{iJkL}\, ,
\end{equation}
by means of the following component-wise relationship:
\begin{equation} \label{eq:D-Arve}
D_{AJBL} = F^{-1}_{Ai} F^{-1}_{Bk} (A_{iJkL} - \delta_{ik}S_{JL})\, ,
\end{equation}
where $\delta_{ik}$ is the Kronecker delta symbol, $F^{-1}_{Ai}$ is the $Ai$ component of ${\bf F}^{-1}$, $S_{JL}$ the $JL$ component of ${\bf S}$, and the Einstein's summation rule is adopted. In turn, $\mathbb{A}_{\text{RVE}}$ is obtained by computing the partial derivative of the RVE average stress ${\bf P}_{avg}$ with respect to the independent components of the macro strain measure ${\bf F}_{avg}$. This is achieved by means of a FE-based sensitivity analysis with ${\bf P}_{avg}$ as output quantities and ${\bf F}_{avg}$ as sensitivity parameters.

%via a sensitivity analysis\begin{footnote}{Result of a sensitivity analysis is the partial derivative of an output quantity with respect to sensitivity parameters.}\end{footnote} of microstructural numerical results with respect to the independent components of the macro strain measure (namely, sensitivity parameters).

%The constitutive relationship ${\bf C}\mapsto \{{\bf S}, \mathbb{D}\}$ on the basis of the RVE characteristics requires then the solution of an equilibrium problem with boundary conditions from Eq. \eqref{eq:ubc} and the computation of post-processing quantities in Eqs. \eqref{eq:C-S_micro} and \eqref{eq:Drve}. 

In detail, the FE residual $\tilde{\bf R}_m$ of the microscopic equilibrium problem results to be a function of nodal values of the displacement field $\tilde{\bf u}_{tot}$. These are conveniently considered to be separated in effective unknowns $\tilde{\bf u}$ and values assigned by boundary conditions $\tilde{\bf u}_{bc}$. Since the microscopic nodal values of the displacement field are sensitive to the applied deformation ${\bf F}_{app}$, so is the microscopic residual. Denoting by ${\bf s} =\mathrm{vec}( {\bf F}_{app})$ the vector of sensitivity parameters, it follows $\tilde{\bf R}_m = \tilde{\bf R}_m(\tilde{\bf u}({\bf s}),\tilde{\bf u}_{bc}({\bf s}))$. As outlined in \cite{vsolinc2015}, the total derivative of $\tilde{\bf R}_m$ with respect to ${\bf s}$ yields
\begin{equation} \label{eq.SensR}
\frac{D\tilde{\bf R}_m}{D{\bf s}} = \frac{\partial\tilde{\bf R}_m}{\partial\tilde{\bf u}}\frac{D\tilde{\bf u}}{D{\bf s}}+\frac{\partial\tilde{\bf R}_m}{\partial\tilde{\bf u}_{bc}}\frac{D\tilde{\bf u}_{bc}}{D{\bf s}}=\bm{0}\, .
\end{equation}
Here, derivatives $\partial\tilde{\bf R}_m/\partial\tilde{\bf u}$ and $\partial\tilde{\bf R}_m/\partial\tilde{\bf u}_{bc}$ are the microscopic tangent matrices of the RVE equilibrium problem (i.e., the primal problem), and they are thus already computed for its solution with a Newton-Raphson iterative procedure. The total derivative $D\tilde{\bf u}_{bc}/D{\bf s}$ denotes the derivative of the imposed essential boundary conditions with respect to sensitivity parameters ${\bf s}$ known from Eq. \eqref{eq:ubc}, \cite{Korelc2016}. Finally, derivative $D\tilde{\bf u}/D{\bf s}$ is the unknown sensitivity of the microscopic nodal unknowns with respect to the chosen sensitivity parameters ${\bf s}$, and it can be computed from Eq. \eqref{eq.SensR} as:
\begin{equation}  \label{eq.SensPar}
\frac{D\tilde{\bf u}}{D{\bf s}} = - \left(\frac{\partial\tilde{\bf R}_m}{\partial\tilde{\bf u}}\right)^{-1} \frac{\partial\tilde{\bf R}_m}{\partial\tilde{\bf u}_{bc}}\frac{D\tilde{\bf u}_{bc}}{D{\bf s}}\, .
\end{equation}
Hence, given that $\mathrm{vec}({\bf F}_{avg})=\mathrm{vec}( {\bf F}_{app}) = {\bf s}$ from the micro-macro scale transition relationship \eqref{eq:micro-macro-trans}, the first elasticity tensor can be obtained as:
\begin{equation} \label{eq:AvSensD}
\mathrm{vec}(\mathbb{A}_{\text{RVE}})=\frac{ \mathrm{vec}(\partial{\bf P}_{avg})}{ \mathrm{vec}(\partial{\bf F}_{avg})} = \frac{1}{\tilde{V}}\int_{\tilde{\Omega}_o}\frac{\mathrm{vec}(\partial\tilde{\bf P}(\tilde{\bf X}))}{\partial {\bf s}}d\Omega = \frac{1}{\tilde{V}}\int_{\tilde{\Omega}_o}  \frac{\mathrm{vec}(\partial\tilde{\bf P}(\tilde{\bf X}))}{\partial\tilde{\bf u}} \frac{D\tilde{\bf u}}{D{\bf s}}d\Omega\, ,
\end{equation}
where $D\tilde{\bf u}/D{\bf s}$ has been computed in Eq. \eqref{eq.SensPar}, and $\partial\tilde{\bf P}/\partial\tilde{\bf u}$ is directly computable since $\tilde{\bf P}$ in each point $\tilde{\bf X}$ is an explicit function of the deformation gradient and hence of displacement unknowns $\tilde{\bf u}$.

For a given modelling choice (from which $\Psi_{mod}$, and then ${\bf C} \mapsto \{ {\bf S}_{mod}, \mathbb{D}_{mod}\}$, are known), the training dataset $\mathcal{D}$ in Eq. \eqref{eq:Dset} is built by applying a sequence of different ${\bf F}_{app}={\bf F}_{inp}$ to span all the sampling points defined within $\mathcal{R}_t$. For each of these assignments, the  equilibrium and sensitivity problems are solved through FEM, such to get the training datapoints ${\bf c} \mapsto {\bf y}_{rem}$ from:
\begin{equation} \label{eq:dataset_creation}
{\bf C} \mapsto {\bf S}_{rem} = {\bf S}_{\text{RVE}} - {\bf S}_{mod}\, , \qquad {\bf C} \mapsto \mathbb{D}_{rem} =  \mathbb{D}_{\text{RVE}} - \mathbb{D}_{mod}\, ,
\end{equation}
respectively from Eq. \eqref{eq:C-S_micro} and Eqs. \eqref{eq:D-Arve} and \eqref{eq:AvSensD}.

\section{Numerical applications} \label{sec:Applications}

Applications are all developed under plane-strain conditions, reducing \emph{ab initio} the number of inputs in ${\bf C}$ from 6 to 3 (i.e., ${\bf c} \in \mathbb{R}^3$). For the reference configuration, a Cartesian coordinate system with base vectors $({\bf E}_1, {\bf E}_2)$  is introduced and parametrized in (material) coordinates ${\bf X}$. For the current configuration, base vectors $({\bf e}_1, {\bf e}_2)$ are introduced. 

Standard displacement FE formulations are employed, discretizing the domain (both at the macro- and micro-scale) with non-overlapping finite elements formulated by following a Galerkin approach. All quantities (i.e., lengths, forces and stiffnesses) are expressed in appropriate and coherent SI-units. Geometrical dimensions, material constants and loading conditions of each numerical model are specified in Table \ref{tab:parameters}. 

All derivatives have been computed using the Mathematica package \textsc{AceGen}, a combined numeric-symbolic tool that allows for Automated Computational Modelling, \cite{Korelc2016}. FE simulations are conducted with the accompanying package \textsc{AceFEM}, \cite{acegen-acefem}. The nonlinear equilibrium problem is solved iteratively by means of a Newton-Raphson algorithm exploiting built-in options in \textsc{AceFEM}, based  either on a fixed load step strategy or on an adaptive path-following algorithm\begin{footnote}{In the implemented adaptive strategy, the load step size is firstly designed on the basis of a desired minimum number of steps, and possibly decreased up to one tenth.}\end{footnote}.

\begin{figure}[tb]
\centering
\includegraphics[width=\textwidth]{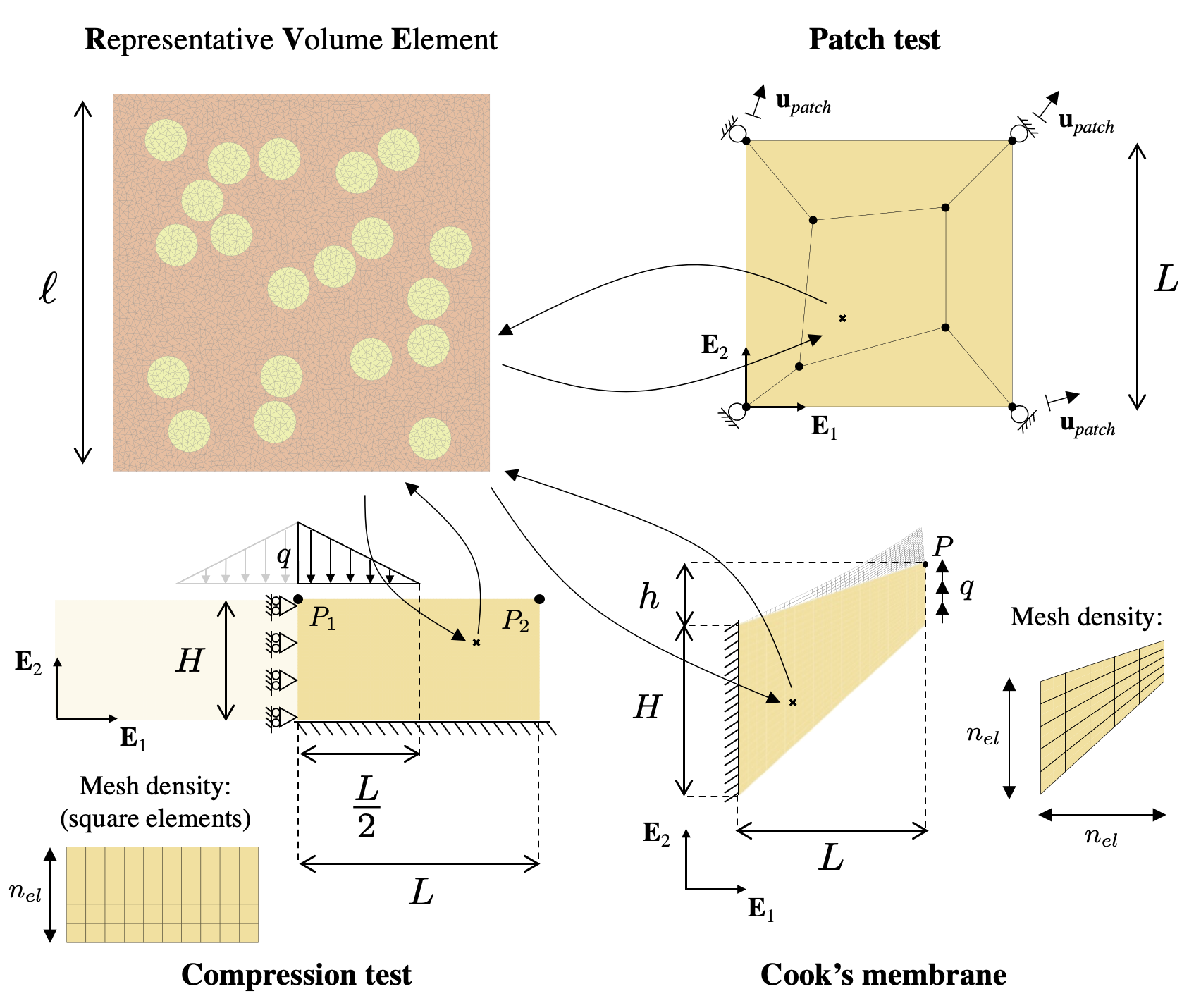}
\caption{Definition of finite-element based numerical simulations. Geometry and mesh details of the microstructural numerical model (i.e., representative volume element) and of macroscale test problems (i.e., patch test, compression test and Cook's membrane). Dimensions and loading conditions of each numerical model are specified in Table \ref{tab:parameters}. 
%Dimensions are given in cm.
}\label{fig:mesh}
\end{figure}

\def\arraystretch{1.3}
\begin{table}
\caption{Geometrical properties of macroscale numerical problems and microscale representative volume element (RVE). For the compression test and the Cook's membrane problem, reference values of mesh density $\bar{n}_{el}$ and of load $\bar{q}$ are also listed. For the RVE, values of material constants in Eq. \eqref{eq:Psi_micro} employed in numerical simulations are indicated. All quantities (i.e. lengths, forces and stiffnesses) are expressed in appropriate and coherent SI-units. }
\centering
\begin{tabular}{|l||c||c|c|c|c||c|c|c|c|c|}
\hline
\multirow{3}{*}{\makecell{Macroscale \\ problems}}& Patch test & \multicolumn{4}{|c||}{Compression test} & \multicolumn{5}{c|}{Cook's membrane} \\
\cline{2-11}
 & $L$ & $L$ & $H$ & $\bar{q}$ & $\bar{n}_{el}$ & $L$ & $H$ & $h$ & $\bar{q}$ & $\bar{n}_{el}$\\
 \cline{2-11}
 & 100  & 100 & 50 & 4 & 30 & 480 & 440 & 16 & 4 & 30 \\
										 \hline
\end{tabular} 
\vspace{1em}

\begin{tabular}{|l||c|c|c||c|c|c|c||c|c|c|c|c|}
\hline
\multirow{3}{*}{\makecell{Microscale \\ (RVE)}} &  \multicolumn{3}{c||}{Geometry} & \multicolumn{8}{c|}{Material} \\ 
\cline{2-12} 
& $\ell$ & $n_{in}$ & $v_{f}$ & $c_1^m$ & $c_2^m$ & $c_3^m$ & $c_4^m$ & $c_1^{in}$ & $c_2^{in}$ & $c_3^{in}$ & $c_4^{in}$   \\
\cline{2-12}
 & 1 & 20 & 20\% & \multicolumn{4}{c||}{1}  & \multicolumn{4}{c|}{1000} \\ 
										 \hline
\end{tabular}\label{tab:parameters}
\end{table}
\def\arraystretch{1}

\subsection{Macroscale numerical models}

At the macroscale, different structural problems are addressed for investigating the performance of the proposed model-data-driven strategy. In terms of (macroscale) FE discretization, the displacement field is interpolated with piecewise bi-linear Lagrange-type shape functions built on four-noded quadrilateral elements. It is noteworthy that, in the macroscale element formulation, the model-based constitutive response is enriched by the data-driven component via a direct implementation of the discretized version of the weak form \eqref{eq:weak} and its linearization \eqref{eq:linear}. To this goal, the metamodel \eqref{eq:y_meta} with optimized hyperparameters is implemented within the \textsc{AceFEM} environment and automatic differentiation exceptions are prescribed in the definition of the stress response ${\bf S}={\bf S}_{mod}+{\bf S}_{rem}$ to obtain the consistent material tangent $\mathbb{D}=\mathbb{D}_{mod}+\mathbb{D}_{rem}$, \cite{Korelc2016}.

In a first campaign of simulations, a patch test is reproduced to verify if the local correction (at Gauss-point level) of the constitutive behaviour through the data-driven component is correctly implemented. A square domain of length $L$ (see Table \ref{tab:parameters}) is discretized with 5 non-rectangular elements, as shown in Fig. \ref{fig:mesh}. The element patch is subjected to a series of homogeneous-type displacement-controlled boundary conditions which correspond to uniform biaxial states of strain. Hence, the displacement field at boundary nodes is set equal to:
\begin{equation}\label{eq:upatch}
{\bf u}_{patch} = ({\bf F}_{patch}-{\bf I}) {\bf X}\, .
\end{equation}
where ${\bf F}_{patch}$ is assigned case-by-case. The constitutive behaviour is analyzed by computing the data-driven component on the basis of the mean value of the metamodel (i.e., Eq. \eqref{eq:mean}) and its $\pm 95\%$ confidence interval derived from the variance output of the metamodel, Eq. \eqref{eq:var}. 

In the second and third applications, a compression test and the Cook's membrane problem are reproduced, such to verify the performance (i.e., accuracy, convergence properties and robustness) of the model-data-driven approach in structural applications. In detail, for these applications, the data-driven component is computed on the basis of the mean value of the metamodel. Applied boundary and loading conditions reproduce classical case studies and are specified in Fig. \ref{fig:mesh}.  The reason for introducing these case studies is that the deformation state attained during the compression test is mainly characterized by nominal components along the coordinate axes, while the Cook's membrane combines bending and shearing \cite{Schroder2020}. For both tests, mesh density is denoted by $n_{el}$. Reference values of mesh density and of applied load, respectively $\bar{n}_{el}$  and $\bar{q}$, are given in Table \ref{tab:parameters} together with geometrical specifications. 

\subsection{Microscale numerical model}\label{sec:appl_RVE_def}

As depicted in Fig. \ref{fig:mesh}, the RVE is chosen to be a square region of dimension $\ell$ containing $n_{in}$ circular stiff inclusions of identical dimensions and randomly distributed in a soft matrix for a total volume fraction $v_f$. Inclusions and matrix are assumed to be perfectly-bonded and hyperelastic, described respectively by the following strain-energy density functions $\Psi_{in}$ (inclusions) and $\Psi_m$ (matrix):
\begin{equation} \label{eq:Psi_micro}
\Psi_{\ast} = c_1^{\ast} I_1 +c_2^{\ast} I_2 + c_3^{\ast} I_3 + \frac{c_4^{\ast}}{2} (I_1 - 3)^2 - 2 c_5^{\ast} \text{Log}(\sqrt{I_3}) \, , \quad \ast=in,\, m\, ,
%
%\frac{\mu_{\ast}}{2} (I_1 - 3) + \frac{\lambda_{\ast}}{4} (I_3 - 1) - \left(\frac{\lambda_{\ast}}{2} + \mu_{\ast}\right)\text{Log}\left(\sqrt{I_3}\right) \, ,
% 
\end{equation}
with $(c_1^i,\ldots, c_5^i)$ and $(c_1^m,\ldots, c_5^m)$ being material parameters of the inclusions and of the matrix, respectively. In particular, $c_5^{\ast} = 2c_1^{\ast}+4c_2^{\ast} + 2c_3^{\ast}$ ensures a stress-free reference configuration, while other parameters' values are given in Table \ref{tab:parameters}, together with geometric details.  

In terms of (microscale) FE discretization, the displacement field is interpolated by means of piecewise linear Lagrange-type shape functions defined on three-noded triangular elements which allow flexibility in mesh generation. Microscale simulations will consider the applied boundary condition in Eq. \eqref{eq:ubc}, by defining the applied deformation gradient ${\bf F}_{app}$ as symmetric\begin{footnote}{For the range of deformations considered in numerical applications, the choice ${\bf F}={\bf F}^T$ corresponds to apply a RVE deformation in the form of a stretch tensor ${\bf U}$, since ${\bf F}$ also results positive definite. Given that the quantity of interest is the right Cauchy-Green deformation tensor ${\bf C}$ and it results ${\bf C} = {\bf F}^T{\bf F}={\bf U}^T{\bf U}$, this choice does not affect the constitutive relationship and the following training procedure.}\end{footnote}. Therefore, ${\bf F}_{app}$ depends on 3 independent components, reading ${\bf F}_{app} = {\bf F}_{app}(F_{11},F_{22},F_{12})$ with
\begin{equation} \label{eq:F_app_micro}
{\bf F}_{app} = {\bf I} + (F_{11}-1) ({\bf e}_1 \otimes {\bf E}_1) + (F_{22}-1) ({\bf e}_2 \otimes {\bf E}_2)+ F_{12} ({\bf e}_1 \otimes {\bf E}_2+ {\bf e}_2 \otimes {\bf E}_1) \, .
\end{equation}
Mesh size in the RVE is fixed on the basis of the outcomes of a convergence analysis (results not shown), leading to around 7800 elements (see Fig. \ref{fig:mesh}).

\subsection{Model-data-driven settings} \label{par:specs_hybrid}

The specific settings of the hybrid model-data-driven strategy employed in numerical applications are here defined.

\subsubsection{Model-driven component specification}

In order to highlight the versatility of the hybrid approach, a simple and general expression for the modelling component $\Psi_{mod}$ will be employed, that is:
\begin{equation} \label{eq:macro_model}
\Psi_{mod}({\bf C}) = C_1 \left[ I_1 - 2 \text{Log}(\sqrt{I_3})\right]\, ,
\end{equation}
where $I_1 = \text{Tr}({\bf C})$, $I_3 = \text{Det}({\bf C})$ and $C_1$ is a model parameter. Numerical results will investigate the importance of the modelling component, both in terms of accuracy and  computational cost of the machine learning training. In order to perform a consistent evaluation, an optimization procedure is conceived for obtaining the best-fitting value of parameter $C_1$, denoted as $C_1^{fit}$. The reader is referred to \ref{sec:app_optimal} for more details. For the RVE introduced in Section \ref{sec:appl_RVE_def}, it result $C_1^{fit}=3.704$. If not differently specified, $C_1 = C_1^{fit}$ is employed in all numerical simulations. 

\subsubsection{Data-driven component specification} \label{sec:data_specs}

\begin{figure}[tb]
\centering
\includegraphics[width=1.0\textwidth]{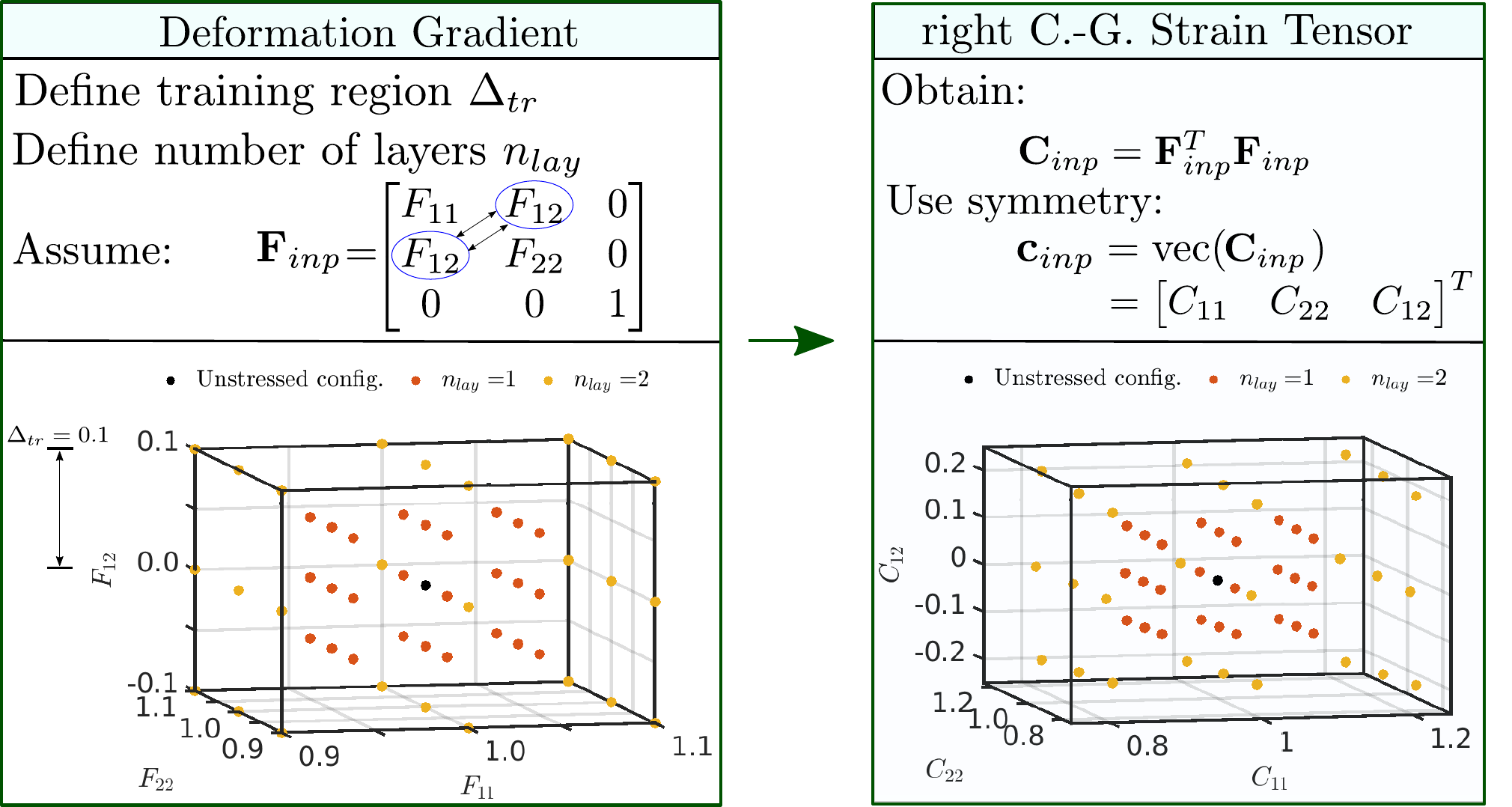}
\caption{Input data generation. Definition of training data region and sampling of $26 n_{lay}+1$ points in the space of the deformation gradient (with $F_{21}=F_{12}$) and translation into the space of right Cauchy-Green (C.-G.) strain tensor. }\label{fig:ML_summary}
\end{figure}

Starting from the definitions in Section \ref{sec::datasetDiscussion}, the data-driven component $\Psi_{rem}$ is fully defined by the number and position of sampling points within the training region $\mathcal{R}_{tr}$. After preliminary analyses, whose results are presented in \ref{sec:appendix_tr}, sample points are positioned in a hypercube grid fashion\begin{footnote}{The authors acknowledge that the performance of Kriging is reliant on the relative positions of the input samples and that specifically grid-like or nearly grid-like structures as presented here could potentially lead to problems, i.e. high-conditioned correlation matrices. The interested reader may refer to \cite{ababou1994condition} for more information. However, no issues have been faced in all the applications treated in this work.}\end{footnote}. Input training points ${\bf F}_{inp}$ are defined in agreement with Eq. \eqref{eq:F_app_micro}, and hence $\mathcal{R}_{tr}$ defines a cube region in the three-dimensional space $(F_{11}, F_{22}, F_{12})$. Sample points are generated by combining $n_{lay}$ layers of 26 points. Each of the layers defines a cubic region of stretch-half-width $\Delta_{l}$ where 8 sample points are positioned at the vertices, 12 points at the middle of each edge, and 6 points at the middle of each face of this cube. The width of each layer $l=1,\ldots, n_{lay}$ is defined as $\Delta_{l}=\Delta_{l-1} + \delta_{tr}$ with $\Delta_{0}=0$ and $\delta_{tr}=\Delta_{tr}/n_{lay}$ a measure of the density of sampling points in the training region. In order to ensure an accurate training in the proximity of the reference configuration, one sample point is added at $F_{11}=1$, $F_{22}=1$ and $F_{12}=0$ for a total of $n_{tp}=26 n_{lay} +1$ training points. The strategy of the input data generation is summarized in Figure \ref{fig:ML_summary}.

Different values of stretch-half-widths $\Delta_{tr}$ and numbers of layers $n_{lay}$ (i.e., the densities of the sampling points $\delta_{tr}$) for $\mathcal{R}_{tr}$ will be considered. In particular, $\Delta_{tr}$ will be increased up to $\Delta_{tr}^{max}=0.25$, that is a stretch range of $50\%$. In order to perform consistent comparisons, two strategies will be adopted:
\begin{enumerate}

\item the \emph{constant width case}: 7 training regions with increasing number of layers at  constant width $\Delta_{tr}=25\%$. This case employs $n_{lay} = 2$, 4, 6, 8, 10, 15, 20, such that the density of each $\mathcal{R}_{tr}$ results approximately equal to $\delta_{tr}=12.5\%$, 6.25\%, 4.1\%, 3.1\%, 2.5\%, 1.6\%, 1.25\%. The number of training points results then $n_{tp}=53$, 105, 157, 209, 261, 391, 521;

\item the \emph{constant density case}: 5 training regions with increasing width at constant density. This case employs $\Delta_{tr}= 5\%$, 10\%, 15\%, 20\%, 25\%, respectively with $n_{lay} = 2$, 4, 6, 8, 10, such that the density of each $\mathcal{R}_{tr}$ is $\delta_{tr}=2.5\%$. The number of training points results then $n_{tp}=53$, 105, 157, 209, 261.

\end{enumerate}

\subsection{Error measures} \label{par:error}

%The evaluation is separated between stress and tangent stiffness tensors, considering two additional correction snapshot matrices ${\bf Y}_{rem}^{S}$ and ${\bf Y}_{rem}^{D}$, analogous to Eq. \eqref{eq:Yrem} but utilizing only stress (${\bf s}_{rem}$) and tangent stiffness (${\bf d}_{rem}$) components respectively.

In order to evaluate the performance of the hybrid model-data-driven approach, normalized root mean-squared errors with respect to a reference solution are introduced. A control region $\mathcal{C}_{c}$ within the deformation gradient space, i.e.
\begin{equation}\label{eq:control_region}
\mathcal{C}_{c}(\lambda_c)=\{{\bf F}\, \text{ s.t. } |({\bf F}-{\bf I}): ({\bf e}_i \otimes {\bf E}_j)| =|[{\bf F}-{\bf I}]_{ij}| < \lambda_{c}\, , \; \forall \, i,j \}\, ,
\end{equation}
 is introduced for the evaluation of the models. Consider a set of $n_{ref}$ reference points ${\bf C}^k=({\bf F}^k)^T{\bf F}^k$ with ${\bf F}^k \in \mathcal{C}_{c}$ within the control region.  At each reference point, the true stress ${\bf S}_{\text{RVE}}^k$ and tangent stiffness $\mathbb{D}_{\text{RVE}}^k$ values are computed from RVE-level simulations (see Section \ref{par:appl_comp}). These values are compared with the stress and stiffness values ${\bf C}^k \mapsto \{{\bf S}^k,\mathbb{D}^k\}$ computed from the mean value obtained from the metamodel in the hybrid model-data-driven approach. Introducing the Voigt representation of these quantities,
\begin{equation}
{\bf s}_{\text{RVE}}^k = \text{vec}({\bf S}_{\text{RVE}}^k)\, , \quad {\bf d}_{\text{RVE}}^k = \text{vec}(\mathbb{D}_{\text{RVE}}^k)\, , \quad  {\bf s}^k = \text{vec}({\bf S}^k)\, , \quad {\bf d}^k = \text{vec}(\mathbb{D}^k)\, ,
\end{equation}
the normalized-root-mean squared errors of the $j$-th component of stress and tangent stiffness vectors yield:
\begin{subequations}
\begin{align}
& \mathcal{E}_{S,j} = \dfrac{1}{n_{ref}} \dfrac{\sqrt{\sum_{k=1}^{n_{ref}} (s_{j}^{k} - s_{\text{RVE},j}^{k})^{2}}}{\max_{k}(s_{\text{RVE},j}^k) - \min_{k}(s_{\text{RVE},j}^k)}\, , \\
& \mathcal{E}_{D,j} =\dfrac{1}{n_{ref}} \dfrac{\sqrt{\sum_{k=1}^{n_{ref}} (d_{j}^{k} - d_{\text{RVE},j}^{k})^{2}}}{\max_{k}(d_{\text{RVE},j}^k) - \min_{k}(d_{\text{RVE},j}^k)}\, ,
\end{align}
\end{subequations}
and the following global error metrics  are defined as:
\begin{subequations} \label{eq:ErrorMetricis}
\begin{equation} \label{eq:ErrorMetricis_a}
\mathcal{E}_{S}(\lambda_{c}) = \text{mean}_{j} \left(\mathcal{E}_{S,j}\right)\, ,  \qquad   \mathcal{E}_{D}(\lambda_{c}) = \text{mean}_{j} \left(\mathcal{E}_{D,j}\right)\, ,  
\end{equation}
and associated respectively with the stress fitting quality ($\mathcal{E}_{S}$), the tangent stiffness fitting quality ($\mathcal{E}_{D}$). An overall fitting quality $\mathcal{E}_{tot}$ is then introduced as: 
\begin{equation}  \label{eq:ErrorMetricis_b}
\mathcal{E}_{tot}(\lambda_{c}) = \frac{\mathcal{E}_{S}(\lambda_{c})+\mathcal{E}_{D}(\lambda_{c})}{2} \, .
\end{equation}
\end{subequations}

\section{Results}\label{sec::resultsAndAppli}

%The following Section \ref{par:patch_res} presents the results of the patch test, hence investigating the performance at the single material point level. Thereafter, Sections \ref{par:punch_res} and  \ref{par:Cook_res} evaluate the performance of the proposed hybrid strategy at the structural level, i.e. on the punch test and the Cook's membrane problem. 

Results obtained with a model-data-driven approach are compared with a classical model-driven strategy (based on Eq. \eqref{eq:macro_model} with $C_1=C_1^{fit}$ and $\Psi_{rem}=0$), as well as a purely data-driven approach (i.e., $C_1=0$, or equivalently $\Psi_{mod}=0$).
Horizontal (along ${\bf E}_1$) and vertical (along ${\bf E}_2$) displacements are denoted respectively by $u$ and $v$. Deformation and stress tensors are represented in a Cartesian coordinate system aligned with $({\bf E}_1,{\bf E}_2)$. The (fourth-order) material tangent constitutive tensor $\mathbb{D}$ is represented as a second-order tensor according to the Voigt notation, resulting in a $3\times 3$ matrix  for the addressed two-dimensional applications. Matrix component $(i,j)$ is denoted by $[ \, \bullet \,]_{ij}$.

\subsection{Model-data-driven fitting: single material point} \label{par:patch_res}

The predictive capabilities of the proposed approach are firstly explored on a single material point by addressing the patch test under different homogeneous biaxial stretch states. In all cases, the computed stresses and strains result homogeneous within the domain, and the patch test is thereby passed. Therefore, the constitutive response at a single material point (i.e., in a Gauss point) is investigated and the accuracy of the hybrid model-data-driven strategy in reproducing the true constitutive behaviour is assessed by computing the global error measures defined in Section \ref{par:error}. In detail, different combinations of $F_{11}$, $F_{22}$ and $F_{12}$ are considered for ${\bf F}_{app}$ in Eq. \eqref{eq:F_app_micro} to compute the micromechanical response via numerical homogenization. For each combination, the RVE constitutive response is compared with the constitutive relationship obtained at the Gauss point level of the patch of elements by applying ${\bf F}_{patch}={\bf F}_{app}$.

\begin{figure}[tbp]
\centering
\subfigure[Uniaxial test with ${\bf F}_{app} = {\bf F}_{app}^{uni}$]{\includegraphics[width=\textwidth]{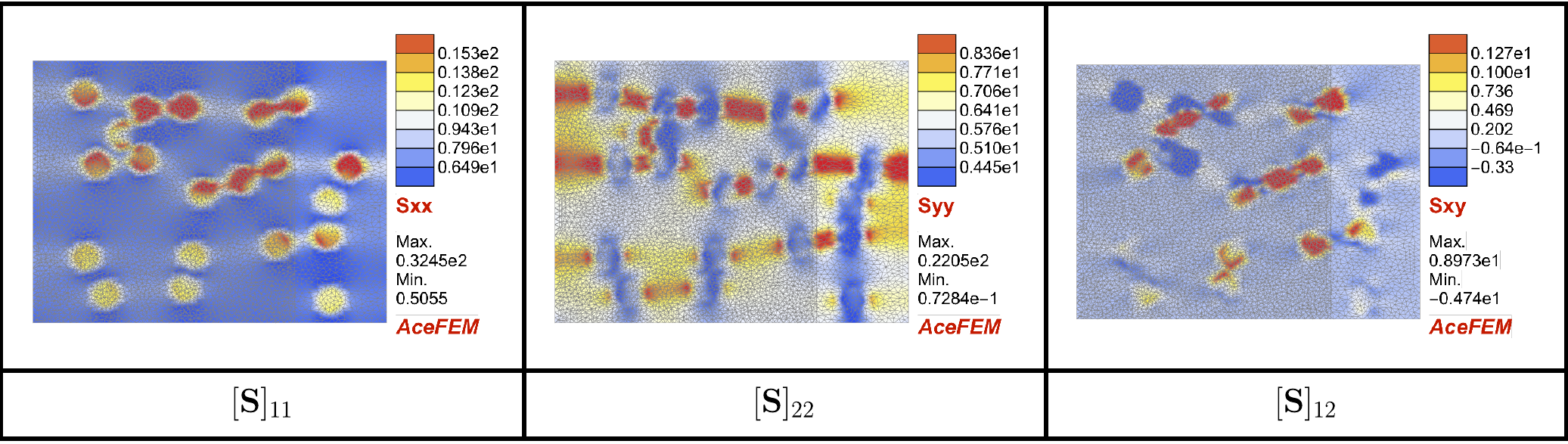} \label{fig:NH_patch_micro_a}}
\subfigure[Multi-axial test with ${\bf F}_{app} = {\bf F}_{app}^{multi} $]{\includegraphics[width=\textwidth]{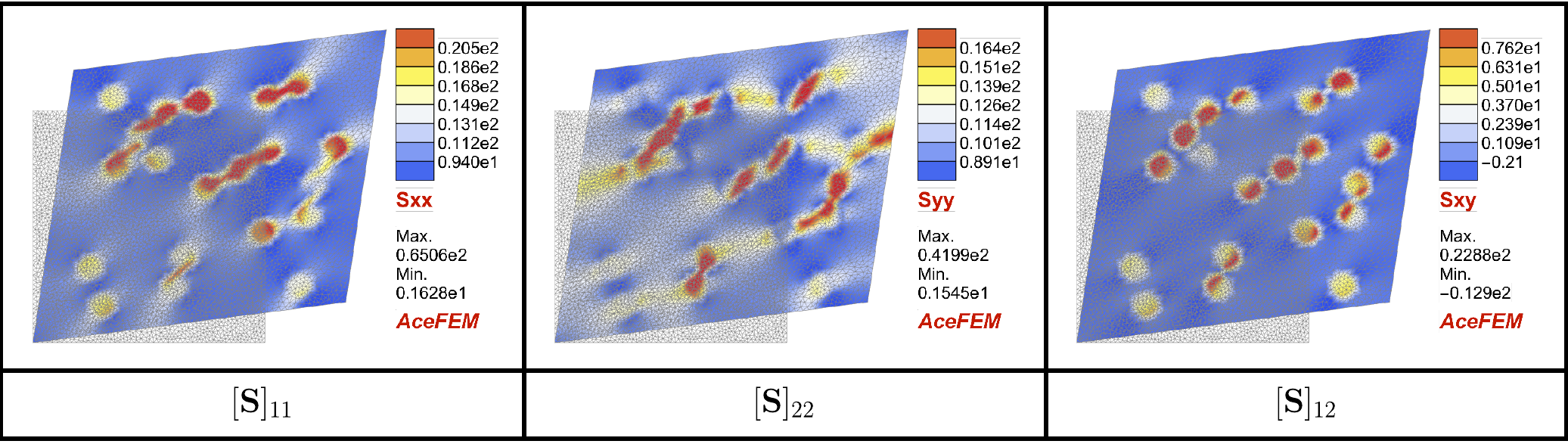} \label{fig:NH_patch_micro_b}}
\caption{Microstructural response obtained with: a) uniaxial displacement-driven test along the first coordinate axis as in Eq. \eqref{eq:patch_F11} ; b) multi-axial displacement-driven test as in Eq. \eqref{eq:patch_mix}. Components of the second Piola-Kirchhoff stress tensor ${\bf S}$ are represented within the final deformed configuration of the domain. }
\label{fig:NH_patch_micro}
\end{figure}

\begin{figure}[tbp]
\centering
\subfigure[Small training region: $\Delta_{tr}=5\%$ and $n_{lay}=2$ (i.e., $\delta_{tr}=2.5\%$)]{\includegraphics[width=0.92\textwidth]{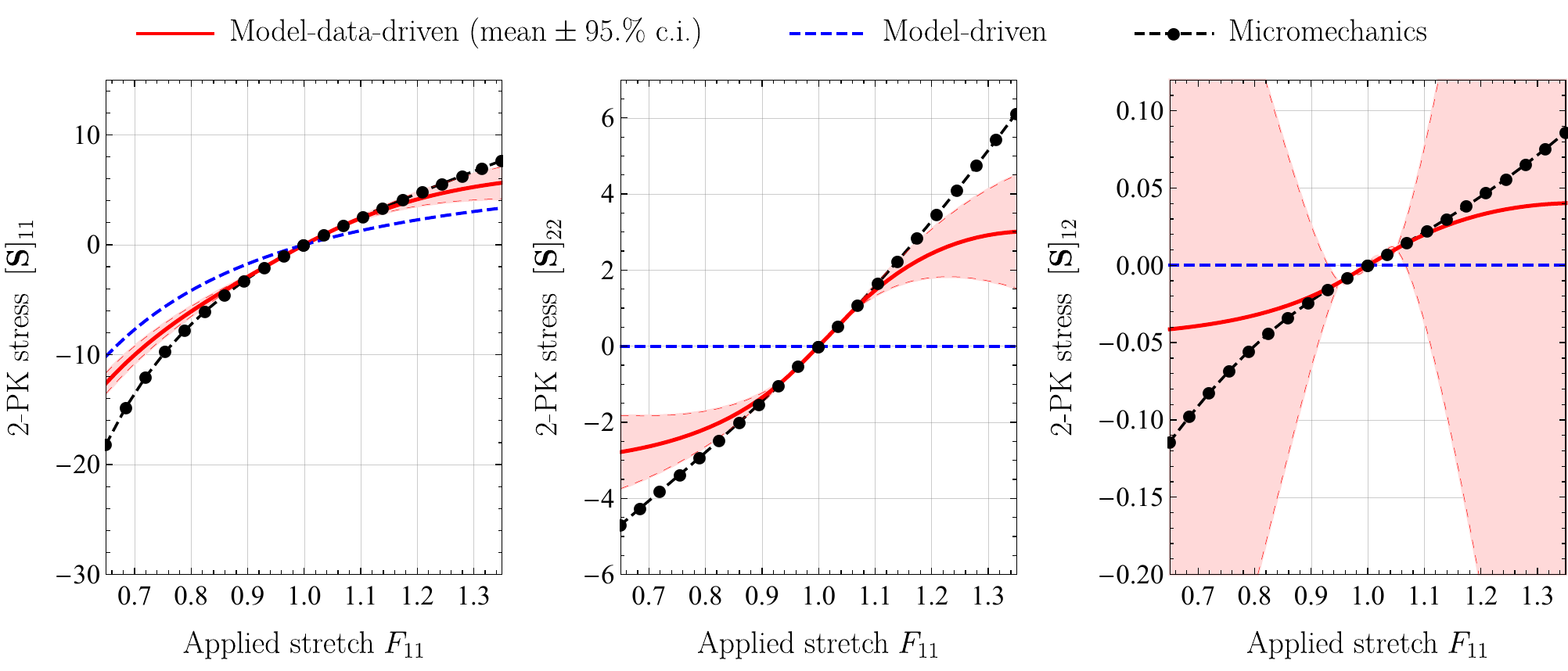} \label{fig:patch-test_res_new_a}}
\subfigure[Large training region: $\Delta_{tr}=25\%$ and $n_{lay}=10$ (i.e., $\delta_{tr}=2.5\%$)]{\includegraphics[width=0.92\textwidth]{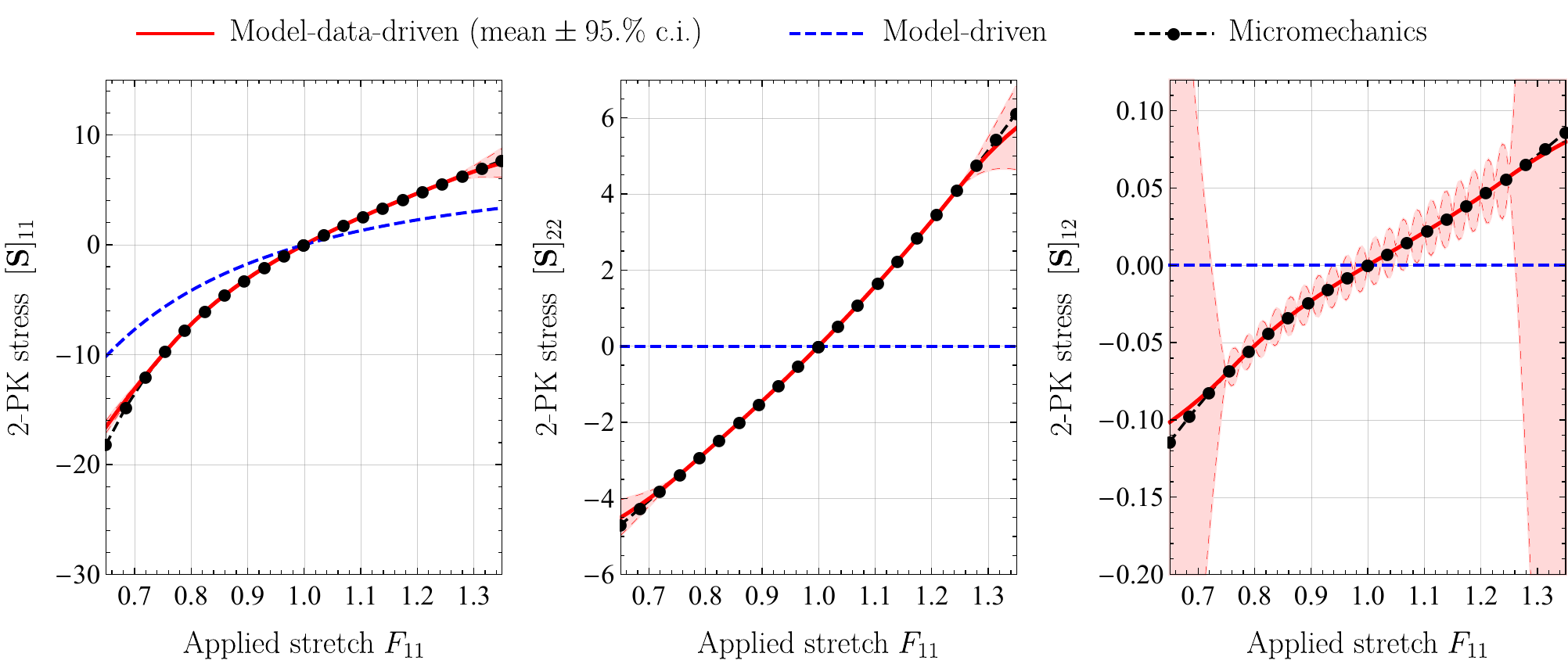} \label{fig:patch-test_res_new_b}}
\subfigure[Sparse training points: $\Delta_{tr}=25\%$ and $n_{lay}=6$ (i.e., $\delta_{tr}=4.1\%$)]{\includegraphics[width=0.92\textwidth]{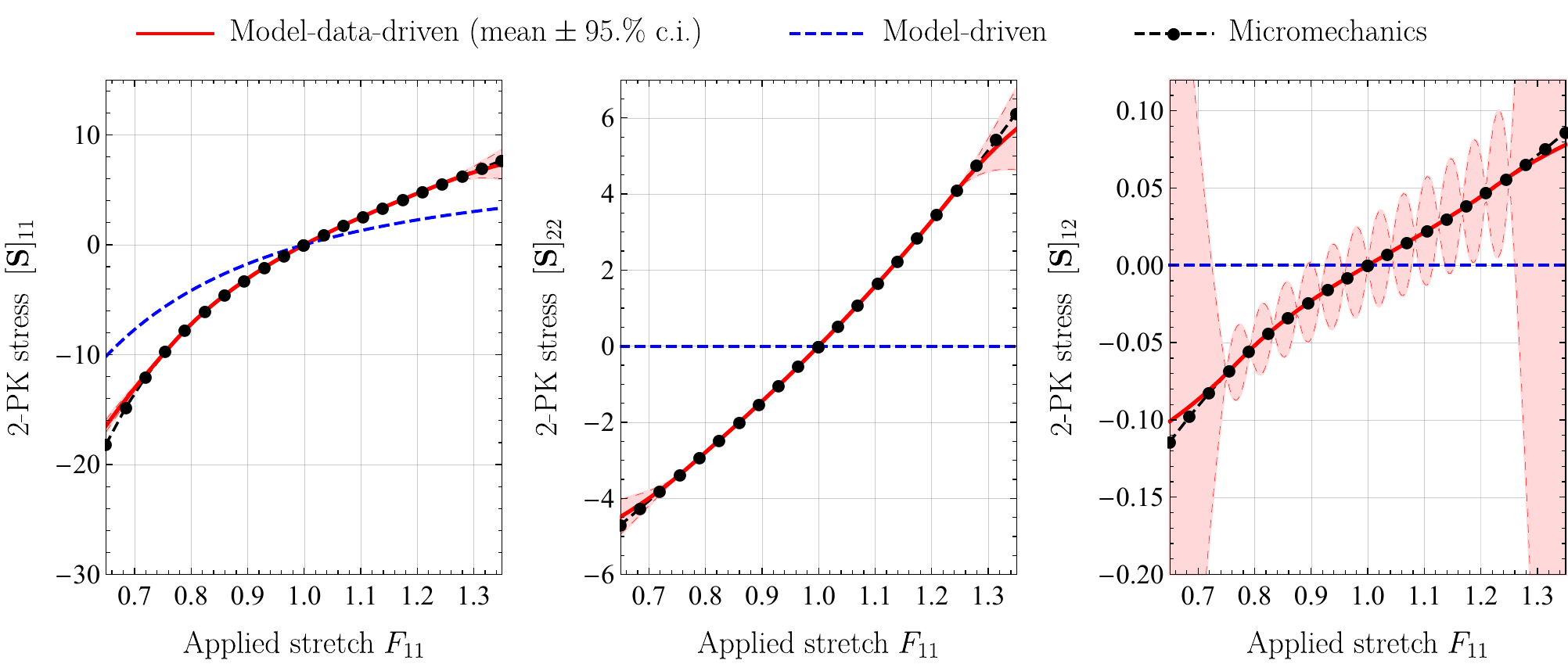} \label{fig:patch-test_res_new_c}}
\caption{Patch test: constitutive behaviour obtained with a uniaxial displacement-driven test along the first coordinate axis (Eq. \eqref{eq:patch_F11}). Results show stress components $[{\bf S}]_{11}$ (left), $[{\bf S}]_{22}$ (center) and $[{\bf S}]_{12}$ (right) versus applied stretch $F_{11}$. The actual micromechanical response is compared with the model-driven simulation and different model-data-driven simulations. The latter are based on metamodels trained with small (top), large (middle), and sparse (bottom) training regions. The output of model-data-driven simulations is represented in terms of mean (red continuous curve) $\pm 95\%$ confidence intervals (shaded areas).}
\label{fig:patch-test_res_new}
\end{figure}

Representative examples on the obtained constitutive responses are presented in Fig. \ref{fig:patch-test_res_new} by addressing a uniaxial displacement-driven traction test along the first coordinate axis, that is: 
\begin{equation} \label{eq:patch_F11}
{\bf F}_{app}^{uni} = F_{11} ({\bf e}_1 \otimes {\bf E}_1) + {\bf e}_2 \otimes {\bf E}_2\, ,
\end{equation}
corresponding to a microstructural deformation as shown in Fig. \ref{fig:NH_patch_micro_a}. Different training regions for the data-driven component are considered: two combinations with the same density $\delta_{tr}$ but different widths $\Delta_{tr}$ (cf., Fig.  \ref{fig:patch-test_res_new_a} and \ref{fig:patch-test_res_new_b}) and two with the same width $\Delta_{tr}$ and different densities $\delta_{tr}$ (cf., Fig.  \ref{fig:patch-test_res_new_b} and \ref{fig:patch-test_res_new_c}). The model-data-driven strategy always improves the fitting capabilities of the model-only approach with respect to the true micromechanical response. Remarkably, the hybrid strategy corrects the postulated analytical law in the stress components which are identically equal to zero from $\Psi_{mod}$. This adjusts the physical outcome obtained from the constitutive model. Within the training region, higher is the density of training points, lower is the uncertainty of the model (see the shear stress components in Figs. \ref{fig:patch-test_res_new_b} and \ref{fig:patch-test_res_new_c}). An excellent fit is obtained with the mean response of the data-driven component: larger the training region, wider is the region where the model-data-driven approach perfectly matches the true response. 

\begin{figure}[p]
\centering
\subfigure[Small training region: $\Delta_{tr}=5\%$ and $n_{lay}=2$ (i.e., $\delta_{tr}=2.5\%$)]{\includegraphics[width=0.92\textwidth]{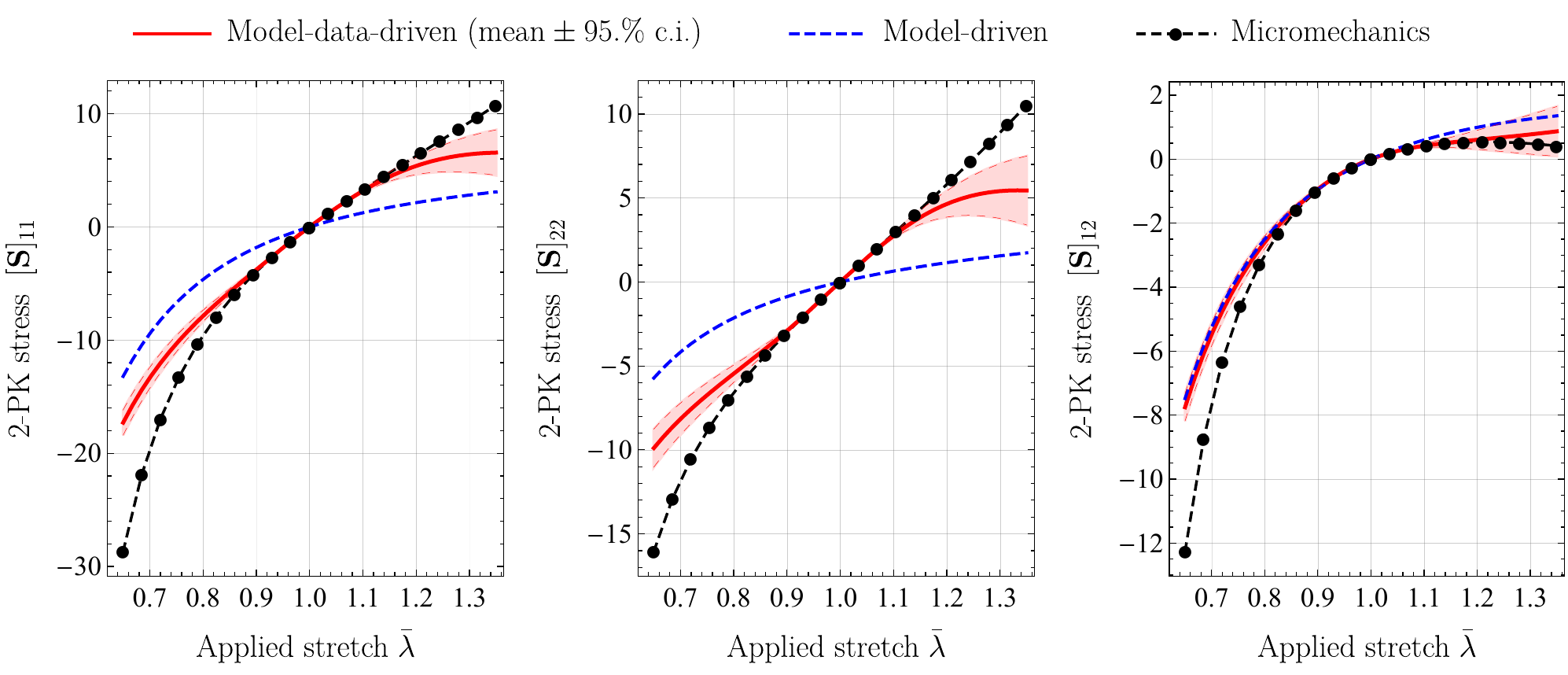} \label{fig:patch-test_res_mix_a}}
\subfigure[Large training region: $\Delta_{tr}=25\%$ and $n_{lay}=10$ (i.e., $\delta_{tr}=2.5\%$)]{\includegraphics[width=0.92\textwidth]{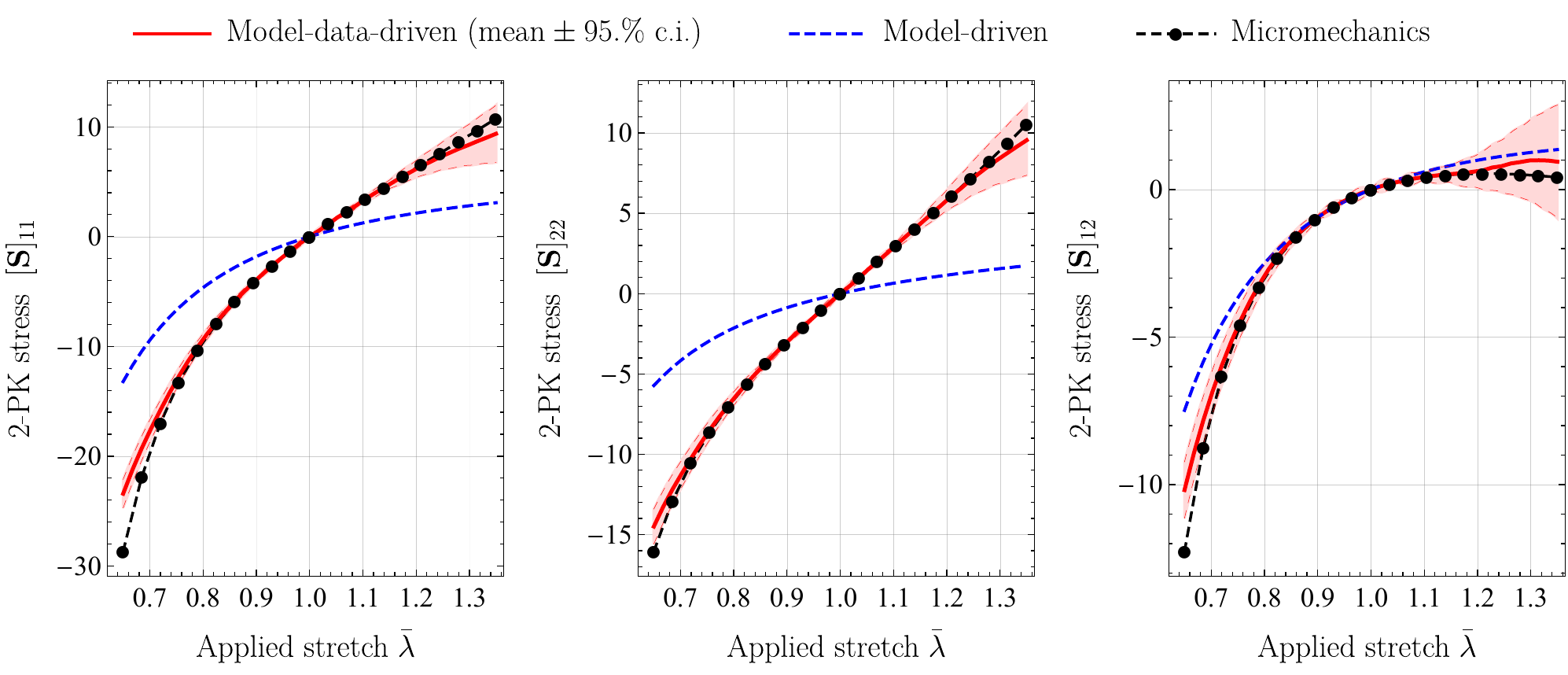} \label{fig:patch-test_res_mix_b}}
\subfigure[Sparse training points: $\Delta_{tr}=25\%$ and $n_{lay}=6$ (i.e., $\delta_{tr}=4.1\%$)]{\includegraphics[width=0.92\textwidth]{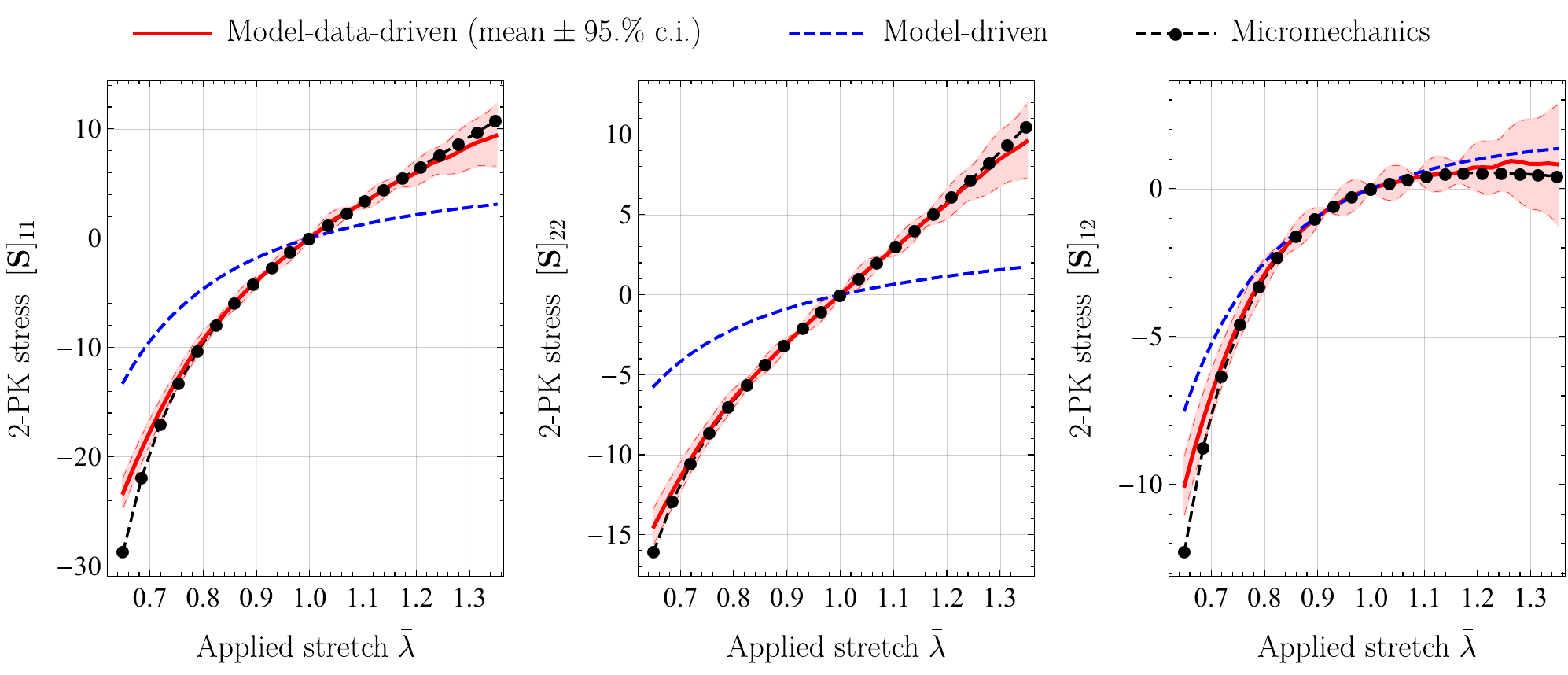} \label{fig:patch-test_res_mix_c}}
\caption{Patch test: constitutive behaviour obtained with a multi-axial displacement-driven test (Eq. \eqref{eq:patch_mix}). Results show stress components $[{\bf S}]_{11}$ (left), $[{\bf S}]_{22}$ (center) and $[{\bf S}]_{12}$ (right) versus applied stretch $\bar{\lambda}$. The true micromechanical response is compared with the model-driven simulation and different model-data-driven simulations. The latter are based on metamodels trained with small (top), large (middle), and sparse (bottom) training regions. The output of model-data-driven simulations is represented in terms of mean (red continuous curve) $\pm 95\%$ confidence intervals (shaded areas).}
\label{fig:patch-test_res_mix}
\end{figure}

The same outcomes are obtained by considering more complex loading conditions, like a multi-axial displacement-driven test associated with:  
\begin{equation} \label{eq:patch_mix}
{\bf F}_{app}^{multi} = \bar{\lambda}({\bf e}_1 \otimes {\bf E}_1) + \frac{\bar{\lambda}}{2}({\bf e}_2 \otimes {\bf E}_2) + \left(\frac{\bar{\lambda}-1}{2}\right)({\bf e}_1 \otimes {\bf E}_2+{\bf e}_2 \otimes {\bf E}_1)\, ,
\end{equation}
corresponding to a microstructural deformation as shown in Fig. \ref{fig:NH_patch_micro_b}. This case study is introduced since material points attain states of deformation which do not correspond to datapoints within the training region. Also for this application, the model-data-driven strategy accurately corrects the constitutive behaviour with respect to the model-driven one, matching the true constitutive behaviour within the training region. In this case, it can be highlighted that the modelling component plays an important role, since the model-data-driven component tends towards the model-driven response outside the training region (see for instance the shear stress in the compressive region in Fig. \ref{fig:patch-test_res_mix_a}).

Qualitative considerations on the effect of different definitions of the training region for the data-driven component are analyzed quantitatively in the following Section \ref{sec:res_material_tr}. The effect of the accuracy of the modelling component is analyzed in Section \ref{sec:res_material_mod}.

\subsubsection{Variation of the training region} \label{sec:res_material_tr}

\begin{figure}[tbp]
    \centering
\subfigure[Constant width case]{\includegraphics[width=0.45\textwidth]{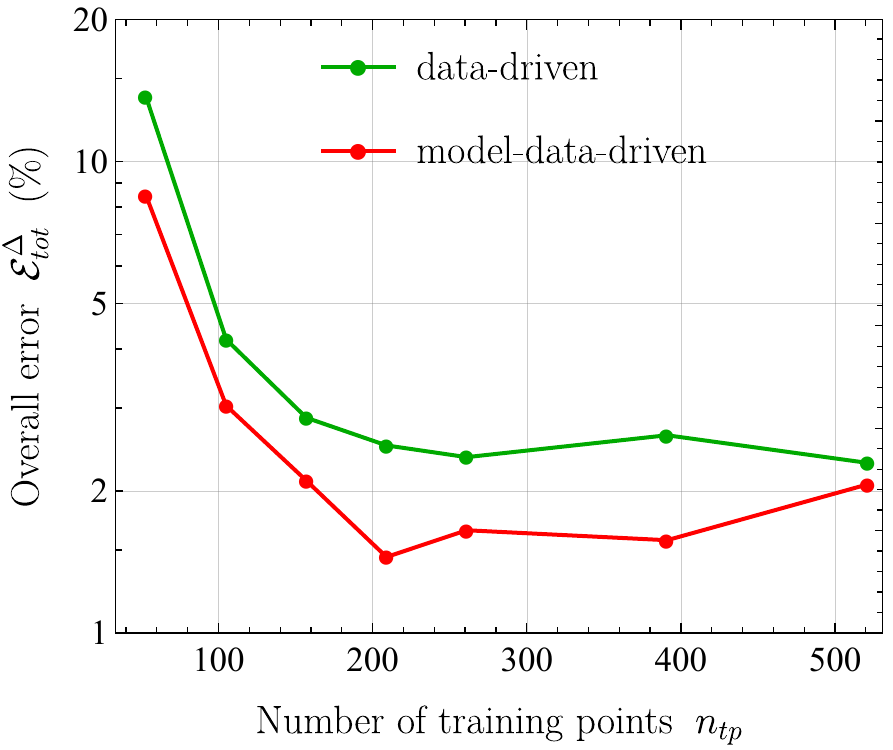} \label{fig:error_dens_results}} \hspace{1em}
\subfigure[Constant density case]{\includegraphics[width=0.45\textwidth]{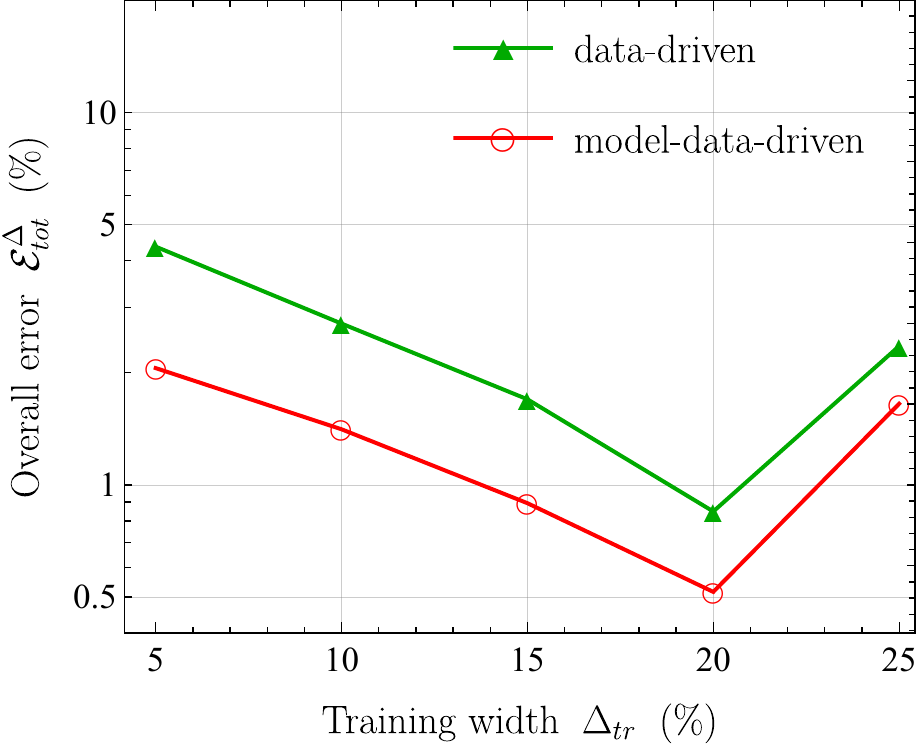} \label{fig:error_width_results}}
\caption{Patch test: overall error $\mathcal{E}_{tot}^{\Delta}$ of the predicted constitutive response for different training regions $\mathcal{R}_{tr}$ within the control region $\mathcal{C}_c^{\Delta}$ with width $\Delta_c = 0.25$: a) constant width case with $\Delta_{tr}=25\%$; b) constant density case with $\delta_{tr}=2.5\%$. Plots show outcomes from the model-data-driven approach and the data-driven approach.}
    \label{fig:error_width_dens_results}
\end{figure}

\begin{figure}[tbp]
\centering
\subfigure[Constant width case]{\includegraphics[width=\textwidth]{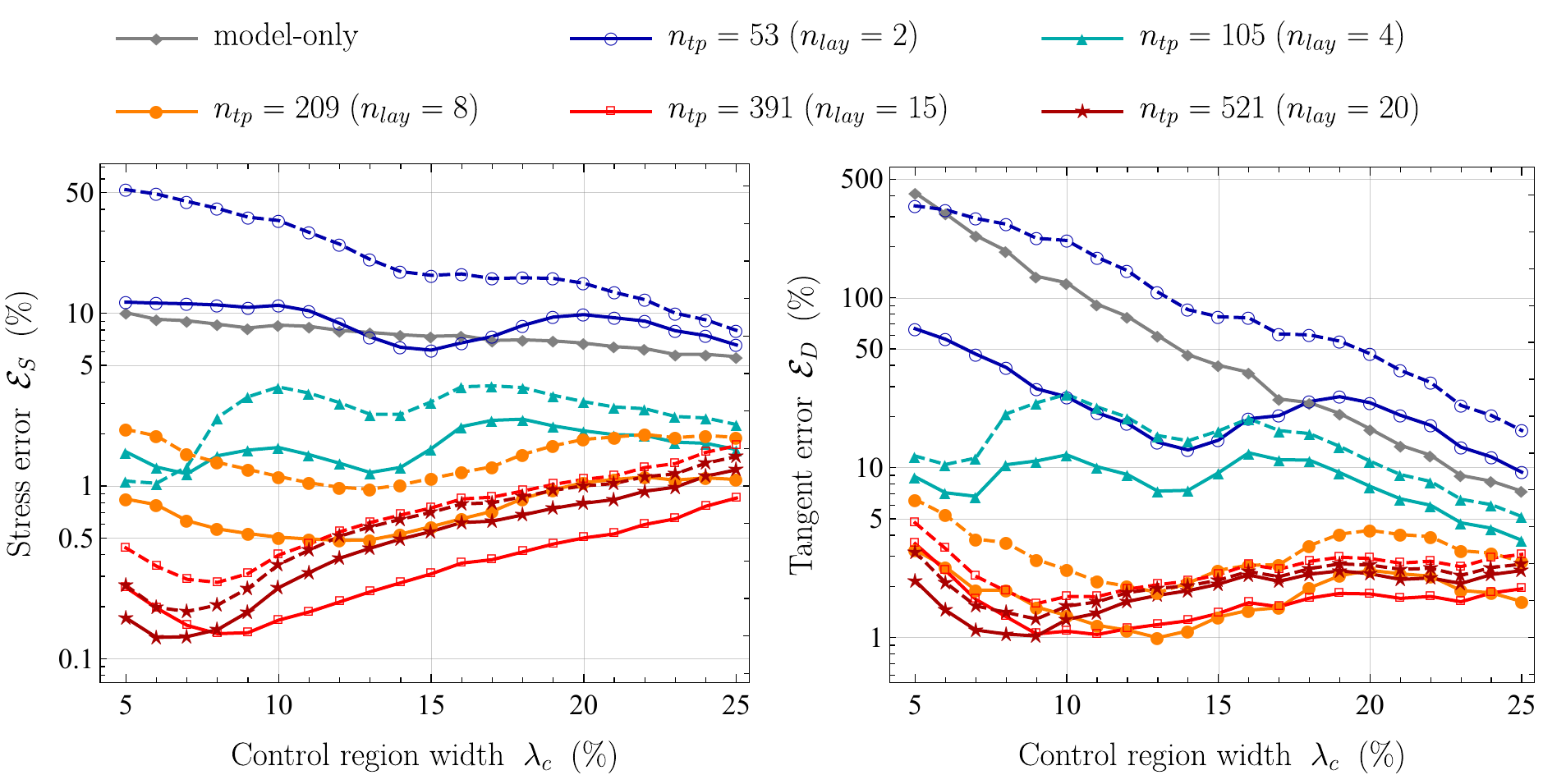} \label{fig:var_error_density_results}}	%
\subfigure[Constant density case]{\includegraphics[width=\textwidth]{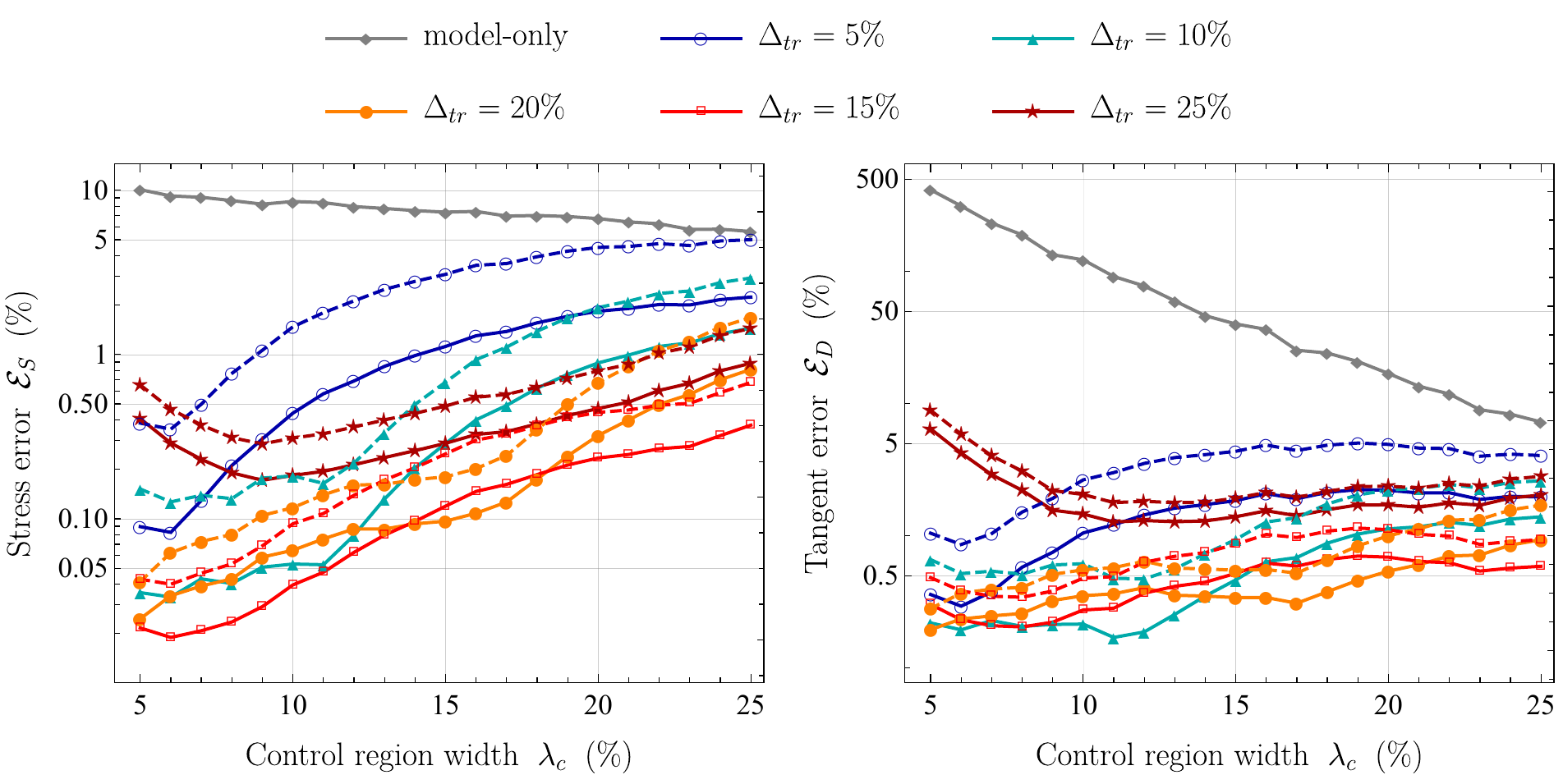} \label{fig:var_error_width_results}}
\caption{Patch test: errors of the predicted constitutive response for different training regions $\mathcal{R}_{tr}$ within the control regions $\mathcal{C}_c(\lambda_c)$ with width $\lambda_c \in (0.05,0.25)$: a) constant width case with $\Delta_{tr}=25\%$; b) constant density case with $\delta_{tr}=2.5\%$. Plots show the stress error $\mathcal{E}_{S}(\lambda_c)$ and the tangent error $\mathcal{E}_{D}(\lambda_c)$ of the hybrid approach (continuous coloured lines), the model-driven approach (continuous grey line), and the data-driven approach (dashed coloured lines).}
\label{fig:var_error_width_dens_results}
\end{figure}

This section investigates the performance of the different strategies by varying the training regions according to the \emph{constant width case} and the \emph{constant density case} (see Section \ref{par:specs_hybrid}). For each case under investigation, the global error metrics defined in Eqs. \eqref{eq:ErrorMetricis} are evaluated within the control region $\mathcal{C}_c^{\Delta}=\mathcal{C}_c(\Delta_c)$ with width $\Delta_c = 0.25$. In detail, $n_{ref} = 30.000$ reference points are sampled within $\mathcal{C}_c^{\Delta}$ with Latin Hypercube Sampling (LHD), \cite{stein1987large}. For the case study under investigation, error $\mathcal{E}_{tot}^{\Delta}=\mathcal{E}_{tot}(\Delta_c)$ is presented in Fig. \ref{fig:error_width_dens_results} both for the model-data-driven and the data-driven approaches. In addition, to evaluate how the error changes at different stretch levels, errors $\mathcal{E}_S(\lambda_c)$ and $\mathcal{E}_D(\lambda_c)$ are also plotted for control regions $\mathcal{C}_c(\lambda_c) \subseteq \mathcal{C}_c^{\Delta}$ of different widths. For the computation of these errors, reference points are selected within $\mathcal{C}_c(\lambda_c)$ as a sub-set of the 30.000 LHD points previously introduced within $\mathcal{C}_c^{\Delta}$. The obtained $\mathcal{E}_S(\lambda_c)$ and $\mathcal{E}_D(\lambda_c)$ for the addressed case studies are shown in Fig. \ref{fig:var_error_width_dens_results}. As reference, results are also compared with the ones based on a model-driven prediction.

Addressing the \emph{constant width case} (i.e., with a varying density $\delta_{tr}$ and number of training points $n_{tp}$), the overall error $\mathcal{E}_{tot}^{\Delta}$  is reported in Fig. \ref{fig:error_dens_results}. 
%
%It can be seen that the machine-learning tool yields the expected outcome, that is a decrease of the training density (aka, an increase of $n_{tp}$) for a given training region results in a decrease of the error. In detail, 
%
The error rapidly decreases by increasing the number of training points from the lowest density ($n_{tp}=53$ and $\delta_{tr}=12.5\%$) and remains low and fairly constant afterwards ($\mathcal{E}_{tot} \approx 2\%$ for $n_{tp}>157$, i.e. $\delta_{tr}<4\%$). This outcome shows that, for the application at hand, the maximum possible fitting capabilities of the machine learning component have been reached, at least with the strategy employed for the definition of the training set. The same considerations can be traced from Fig. \ref{fig:var_error_density_results} which shows stress and tangent errors $\mathcal{E}_S(\lambda_c)$ and $\mathcal{E}_D(\lambda_c)$ for control regions $\mathcal{C}_c(\lambda_c) \subseteq \mathcal{C}_c^{\Delta}$ of different widths.

Interestingly, Figs. \ref{fig:error_dens_results} and \ref{fig:var_error_density_results} reveal that the error can also slightly increase by employing a higher number of training points. In fact, the number of datapoints is directly proportional to the complexity of the minimization problem which is solved for the training of the data-driven component. For applications requiring a higher precision of the data-driven component, an improvement of the optimization algorithm for the machine-learning training or a different definition of the position of the training points might be pursued. However, the error associated with the hybrid approach is always lower than the one obtained with the data-driven strategy, showing a reduction up to 50\% relative error. This opens the door to improvements based on optimization of the modelling component, possibility which will be investigated in the following Section \ref{sec:res_material_mod}.

Addressing the \emph{constant density case} (i.e., with a varying training width $\Delta_{tr}$), the overall error $\mathcal{E}_{tot}^{\Delta}$ is presented in Fig. \ref{fig:error_width_results}. As shown before, the model-data-driven strategy always outperforms the data-driven one reducing by half the obtained error. In particular, errors are inversely proportional to the training width for $5\% \leq \Delta_{tr} \leq 20\%$, but it increases for $\Delta_{tr} = 25\%$. This outcome is due to the increasing non-linearities of the constitutive response in the stretch interval $(0.2,0.25)$, as it can be qualitatively observed in Fig. \ref{fig:patch-test_res_new} (see $\bar{\lambda} \approx 0.75$). 

To investigate thoroughly the reason of this result, the individual stress and tangent errors $\mathcal{E}_S(\lambda_c)$ and $\mathcal{E}_D(\lambda_c)$ are shown in Fig. \ref{fig:var_error_width_results}. First of all, it can be seen that,  for $\Delta_{tr} \in (0.05,0.2)$, the microstructural response is very well predicted inside each training region. The error significantly increases outside the training regions, tending towards the one of the model-driven approach, though remaining significantly lower for the range of stretches here investigated. Secondly, it can be noticed that the case with $\Delta_{tr} = 25\%$ shows a worse performance than the one with $\Delta_{tr} = 20\%$, confirming the evidence observed  from Fig. \ref{fig:error_width_results}. In particular, the error significantly increases at low stretches: the high non-linearities in the stretch interval $(0.2,0.25)$ make the machine-learning procedure more challenging and the numerical strategy for building the metamodel (i.e., for solving the optimization problem in Eq. \eqref{eq:optim_ML}) less effective. Then, the quality of the overall fitting reduces. It is worth highlighting that this outcome occurs for both the hybrid and the data-driven approach. Improvements on the data-driven component (e.g., better optimization algorithms, different definition of the datapoints) will automatically increase the efficacy of the hybrid strategy. 

\subsubsection{Effect of the accuracy of the modelling component} \label{sec:res_material_mod}

This section investigates if the performance of the model-data-driven approach also depends on the accuracy of the macromodel, that is on the resemblance between functionals $\Psi_{el}$ and $\Psi_{mod}$ in Eq. \eqref{eq:S}. Preliminary considerations can be traced, for instance, from Figs. \ref{fig:patch-test_res_mix} and \ref{fig:var_error_width_dens_results}. In fact, the model-data-driven response tends towards the model-driven one outside the training region. Therefore, an accurate modelling component would improve the performance of the hybrid strategy outside $\mathcal{R}_{tr}$.

\begin{figure}[tbp]
\centering
\includegraphics[width=0.98\textwidth]{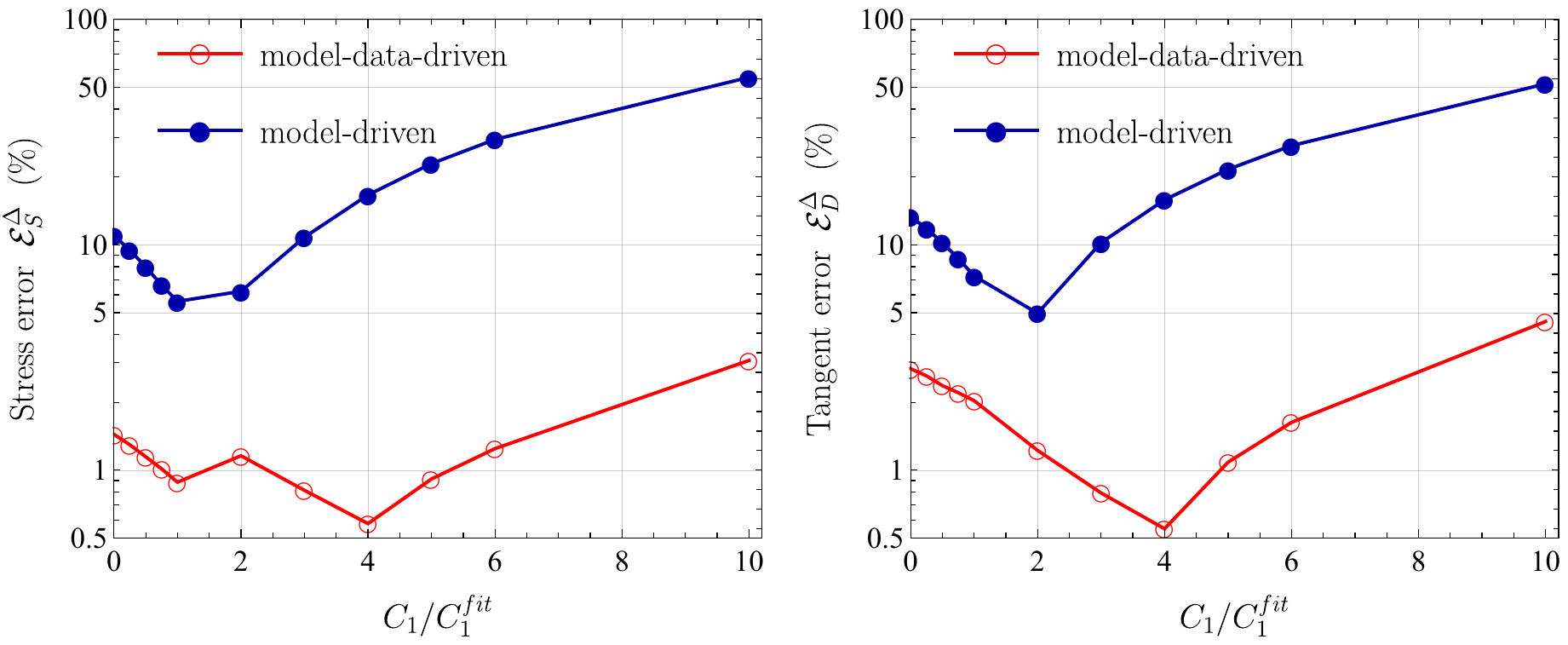}
\caption{Patch test: stress error $\mathcal{E}_{S}^{\Delta}$ (left) and tangent error $\mathcal{E}_{D}^{\Delta}$ (right) obtained with the model-driven  and the model-data-driven approach (with $\Delta_{tr} = 25\%$ and $\delta_{tr}=2.5\%$, that is $n_{tp}=261$ sample points). }\label{fig:error_C1}
\end{figure}

As regards the performance within $\mathcal{R}_{tr}$, additional analyses are conducted by training the data-driven component of the hybrid strategy for different values of the model parameter $C_1$ in $\Psi_{mod}$ of Eq. \eqref{eq:macro_model}, considering $C_1 \in (0,10) C_1^{fit}$. It is worth highlighting that $C_1=0$ corresponds to a purely data-driven strategy. In all tests, the training region is fixed with $\Delta_{tr} =25 \%$ and $n_{lay}=10$ (resulting in $\delta_{tr}=2.5\%$ and $n_{tp}=261$). Errors $\mathcal{E}_{S}^{\Delta}$ and $\mathcal{E}_{D}^{\Delta}$ are depicted in Fig. \ref{fig:error_C1} over variations of $C_{1}$ for both the model-driven approach as well as for the different generated model-data-driven strategies. As expected, the model-driven approach is most accurate in terms of stress error $\mathcal{E}_{S}^{\Delta}$ at around $C_1\approx C_1^{fit}$ and it significantly increases for wrong choices of $C_1$. On the other hand, the error obtained with the hybrid approach is significantly more robust than the model-driven one over variations of $C_1$. This property is advantageous for the preliminary calibration stage of the modelling component. However, the accuracy of the machine learnt correction is not independent from the quality of the fitting of the modelling component, the error significantly increasing when the order of magnitude of $C_1$ is wrongly chosen in the first place.

\subsection{Model-data-driven structural response: compression test} \label{par:punch_res}

\begin{figure}[tbp]
\centering
\includegraphics[width=0.98\textwidth]{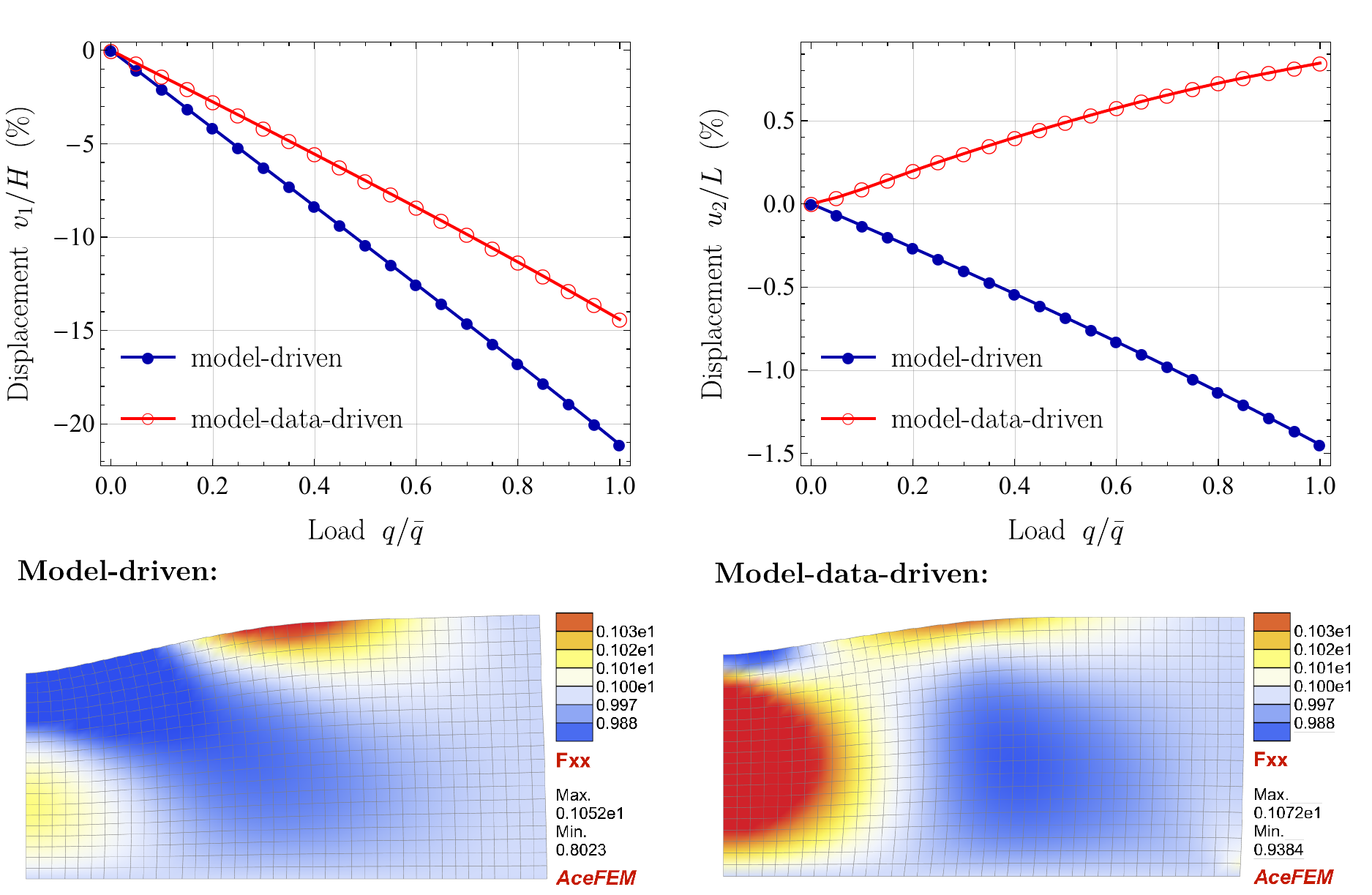}
\caption{Compression test: structural response obtained with a model-driven approach and a model-data-driven approach ($\Delta_{tr}=25\%$ with $n_{lay}=10$). Top: vertical displacement $v_1$ at point $P_1$, normalized with respect to specimen width $H$, and horizontal displacement $u_2$ at point $P_2$, normalized with respect to specimen length $L$ (see Fig. \ref{fig:mesh}). Bottom: distribution of deformation component $[{\bf F}]_{11}$ within the domain. Simulation settings: target maximum load equal to $\bar{q}$; fixed step algorithm with 20 load steps. Contour plots are shown in the final deformed configuration.}\label{fig:punch_res}
\end{figure}

The structural response obtained for the compression test with the hybrid model-data-driven approach  is reported in Fig. \ref{fig:punch_res}, and compared with the one associated with the model-driven approach. The figure shows relevant displacements and the distribution of deformation component $[{\bf F}]_{11}$. Although main features  of the constitutive approach are captured by the model component $\Psi_{mod}$ of the strain-energy function, the introduction of microscale numerical homogenization information in $\Psi_{rem}$ by means of a data-driven strategy significantly changes the outcomes of the compression test. For the latter results, the data-driven settings correspond to a training region $\Delta_{tr}=25\%$ and number of layers $n_{lay}=10$. 

%Outcomes obtained with different settings are presented next, allowing also for a consistent comparison between the proposed approach and a data-driven one.

\subsubsection{Compression test: different training settings}

\begin{figure}[tbp]
\centering
\subfigure[Displacement $v_1/H$ versus load $q/\bar{q}$]{\includegraphics[width=0.98\textwidth]{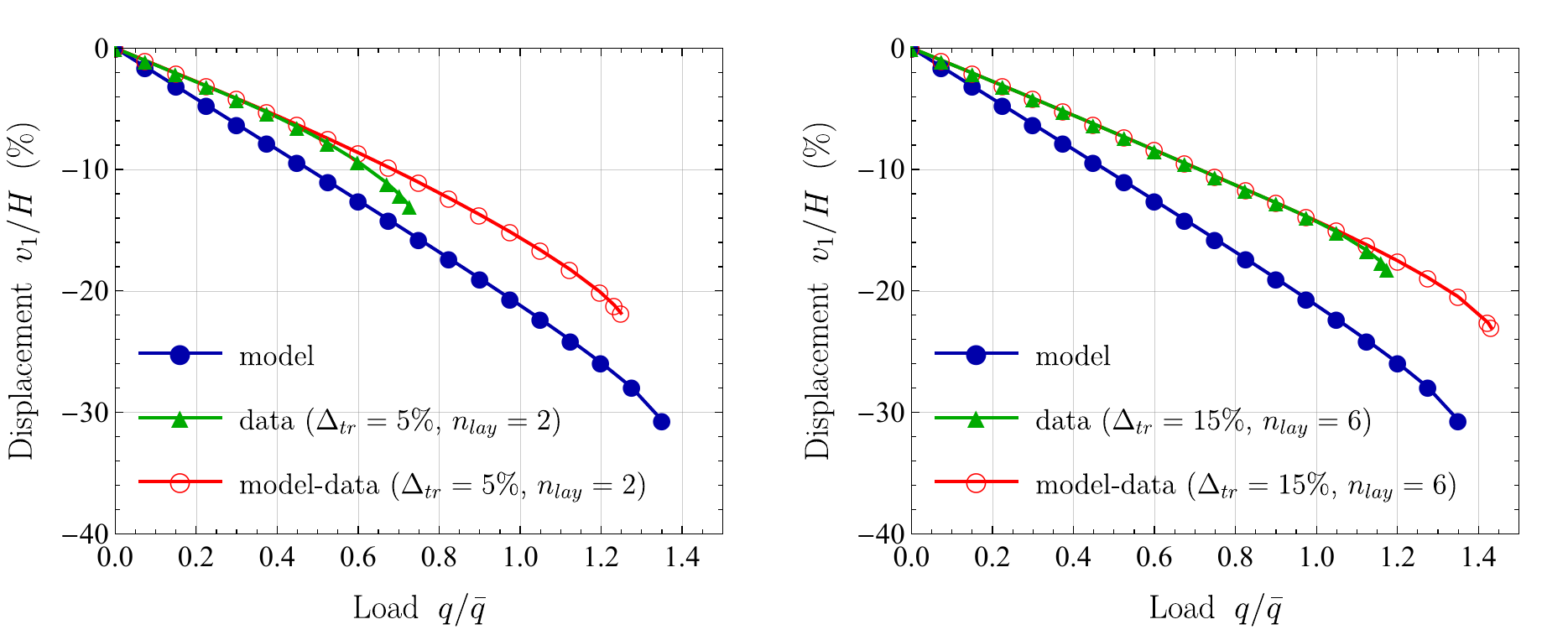} \label{fig:punch_res_comparison_a}}
\subfigure[CG deformation components in the domain versus load $q/\bar{q}$: mean values (line with symbols); standard deviations (error bars); and maximum/minimum values (dashed lines) within the domain. The training region is also reported as the gray shaded area.]{\includegraphics[width=0.98\textwidth]{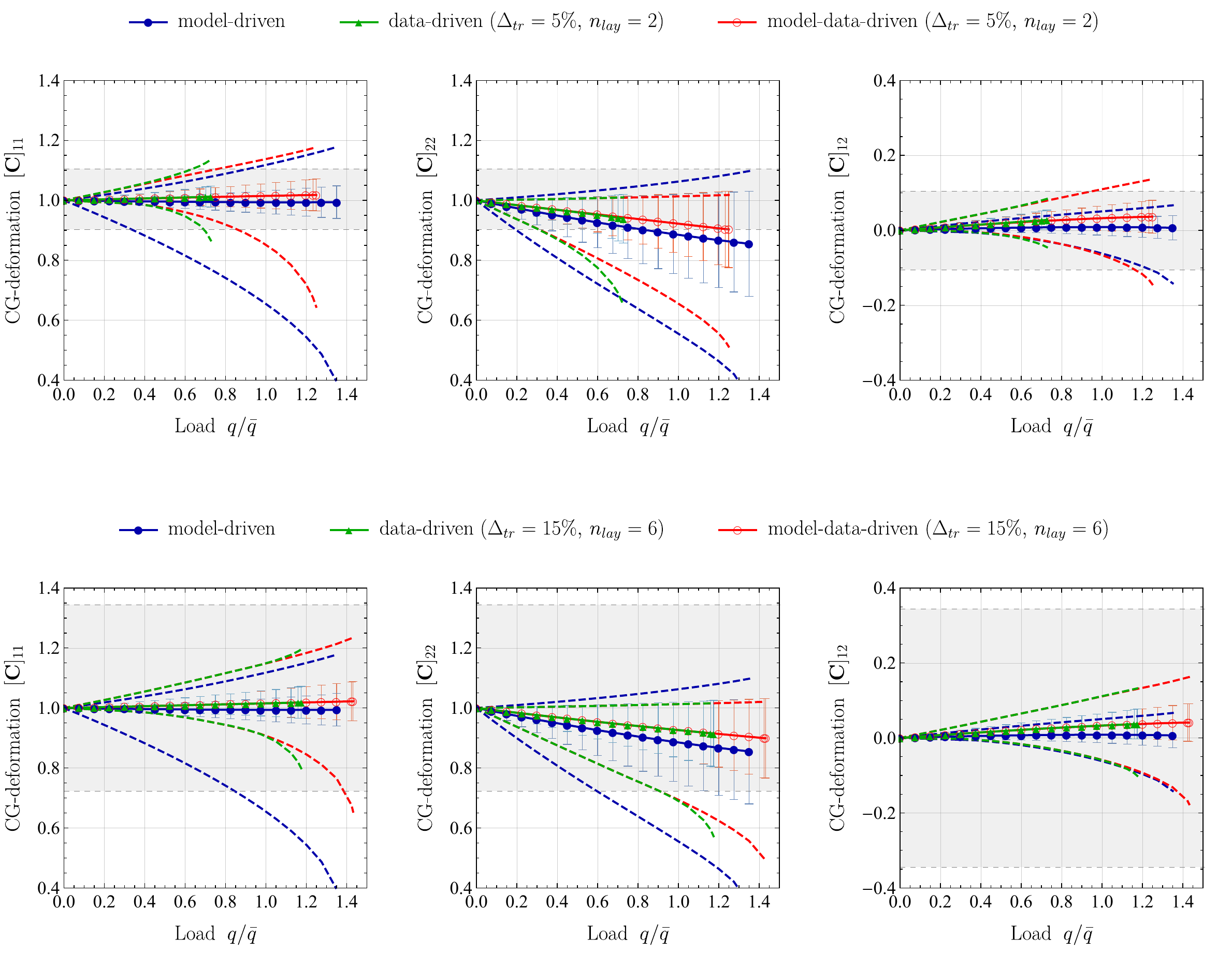}  \label{fig:punch_res_comparison_b}}
\caption{Compression test: structural response obtained with two different training regions, that is $\Delta_{tr}=5\%$ with $n_{lay}=2$ and $\Delta_{tr}=15\%$ with $n_{lay}=6$ (varying width at constant density). Results show: a) the obtained macroscopic structural behaviour; b) components of the Cauchy-Green (CG) deformation tensor ${\bf C}$ within the domain. Simulation settings: target maximum load equal to $1.5\bar{q}$; adaptive load step algorithm with minimum number of steps equal to 20.}
\label{fig:punch_res_comparison}
\end{figure}

This section presents results obtained with different settings of the training region, and compared with the response obtained by employing a purely data-driven strategy. Firstly, Fig. \ref{fig:punch_res_comparison_a} shows results obtained with two training regions of different width (i.e., $\Delta_{tr}=5\%$ and $\Delta_{tr}=15\%$), but same training points density (i.e., $n_{lay}=2$ and $n_{lay}=6$). The simulation is run by imposing a target maximum load $q=1.5 \bar{q}$, and results interrupt when divergence in the iterative solution scheme occurs. The model-data-driven simulation is significantly more robust than the data-driven one, with the maximum load reached with $\Delta_{tr}=5\%$ respectively equal to $q\approx 1.25 \bar{q}$ (model-data) and $q\approx 0.7 \bar{q}$ (data), while with $\Delta_{tr}=15\%$ to $q\approx 1.45 \bar{q}$ and $q\approx 1.2\bar{q}$.

The reason for divergence of the data-driven simulation can be investigated in Fig. \ref{fig:punch_res_comparison_b}, which shows the values of the Cauchy-Green deformation tensor's components in terms of mean values, standard deviations and maximum/minimum values in the domain at each load step. The data-driven simulation diverges very close to the load level at which the deformation in the domain take values outside of the training region. Moreover, although outside the training region the response should not be fully trusted, it is noteworthy that model-data-driven results are not highly different in the two cases even for the load range for which deformation lies outside $\Delta_{tr}=5\%$ but inside $\Delta_{tr}=15\%$. 

\begin{figure}[tbp]
\centering
\subfigure[Last converged load (left) and numerical performance as in Eq. \eqref{eq:IPI} (right)]{\includegraphics[width=0.98\textwidth]{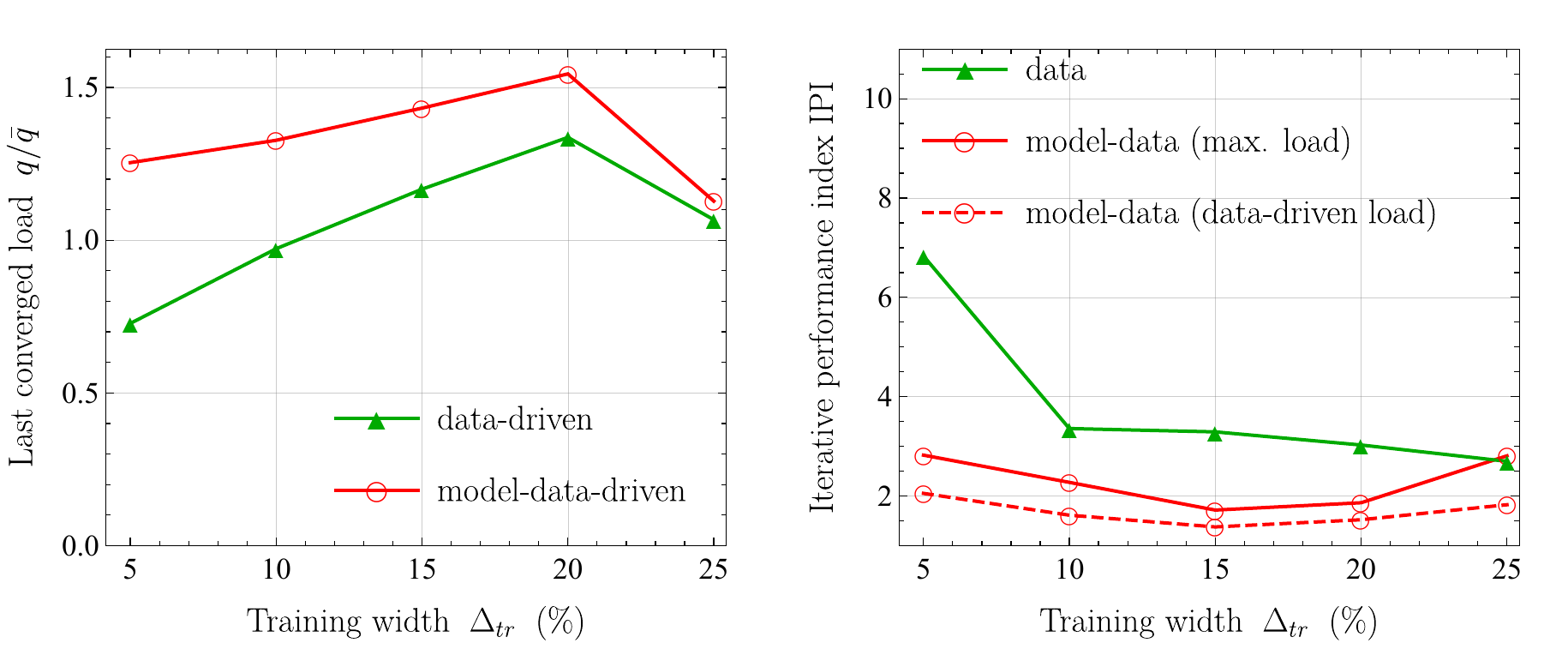} \label{fig:punch_res_width_a}}
\subfigure[Variation of displacement $v_1$ with respect to the reference case $\Delta_{tr}=20\%$]{\includegraphics[width=0.98\textwidth]{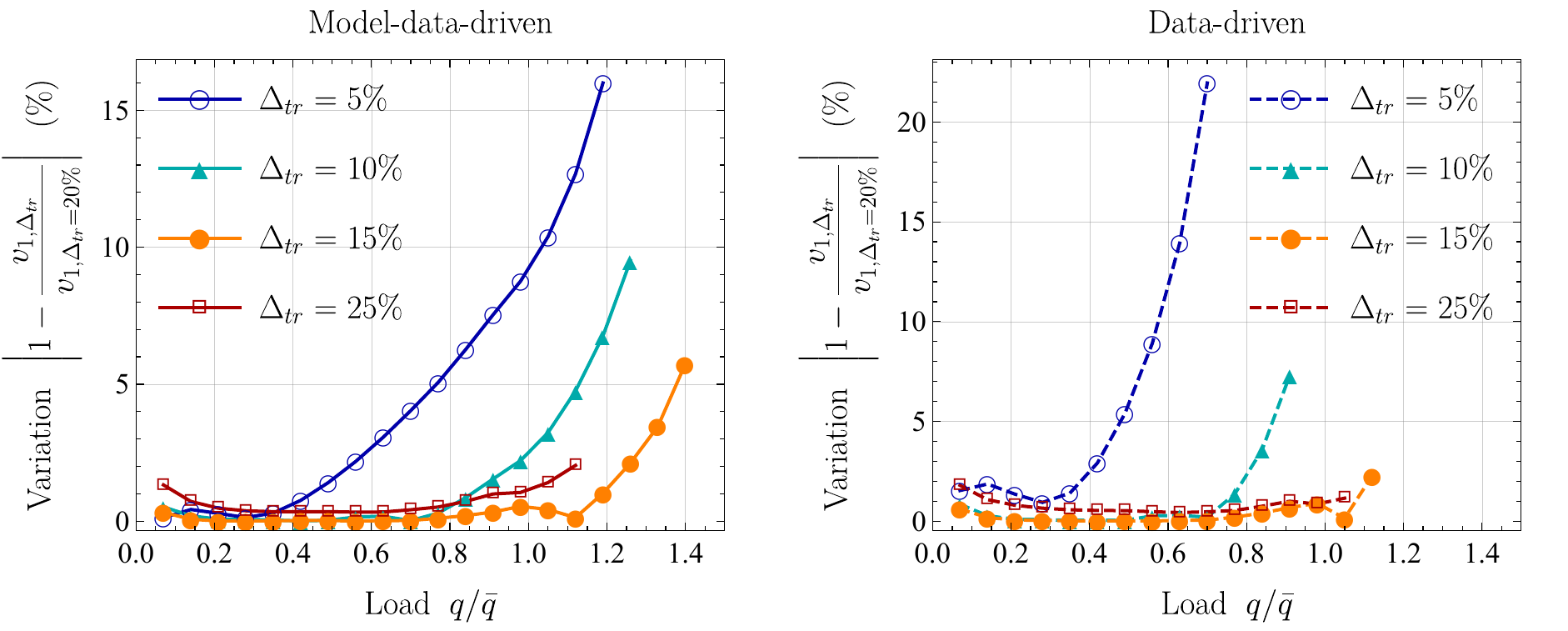} \label{fig:punch_res_width_b}}
\caption{Compression test: training region with constant density and varying width (\emph{constant density case} in Section \ref{par:specs_hybrid}). Results investigate on: a) numerical robustness and performance; b) sensitivity with respect to training region definition, from which accuracy can be evaluated. Simulation settings: target maximum load equal to $1.75\bar{q}$; adaptive load step algorithm with minimum number of steps equal to 25.}\label{fig:punch_res_width}
\end{figure}

This outcome is confirmed in Fig. \ref{fig:punch_res_width}, which shows results obtained at constant density and varying width (\emph{constant density case} in Section \ref{par:specs_hybrid}). Firstly, it is highlighted that the last converged load is always higher with the model-data-driven strategy than with the data-driven one (see Fig. \ref{fig:punch_res_width_a}). As previously noted, data-driven simulations diverge as soon as exceeding the training width, while model-data-driven ones not. Remarkably, the trend of the last converged load mimics the one of the constitutive error (cf., Fig. \ref{fig:error_width_results}). Moreover, Fig. \ref{fig:punch_res_width_b} shows the variation of the obtained displacement with respect to the one corresponding to a reference simulation setting. In this case, the reference setting is chosen equal to $\Delta_{tr}=20\%$, which corresponds to the most accurate constitutive fitting (cf., Fig. \ref{fig:error_width_results}). To be consistent, the reference of model-data-driven results is obtained with a model-data-driven approach, while the one of data-driven results with a data-driven strategy. Small variations denote that results are minimimally sensitive with respect to the definition of the training dataset, which in turn is an indication of accuracy for the training scheme. In fact, variations of data-driven results steeply increase at a certain load level (associated with the exceeding of the training region) until divergence. Also variations (i.e., error) of model-data-driven results increase outside of the training region, but now in a gradual and controllable manner.

Furthermore, the performance of the iterative solution schemes with the two approaches is investigated by introducing the iterative performance index IPI, defined as:
\begin{equation} \label{eq:IPI}
\text{IPI} = \frac{N_{it}}{N_{it}^{mod}}
\end{equation}
where $N_{it}$ and $N_{it}^{mod}$ are the average number of Newton-Raphson iterations per load step, respectively for the simulation at hand and for a model-driven simulation at the same load level (for which quadratic convergence is achieved). First of all, values of the IPI index in Fig. \ref{fig:punch_res_width_a} are analyzed by comparing data-driven simulations with the ones from a model-data-driven simulation where the applied load corresponds to the final one obtained in the data-driven case. In this case, the required number of iterations in the model-data-driven case is significantly lower than in the data-driven case. Remarkably, model-data-driven values are very close to one, revealing that the number of iterations is practically the same as in the model-driven case, although the constitutive behaviour is generally more complex. The model-data correction on the constitutive response (both in terms of stresses and tangent matrix) allows to obtain a behaviour of the Newton-Raphson iterative scheme close to quadratic convergence, revealing that a consistent tangent is practically employed (as from analytical derivations possible with a model-driven strategy).

%as if a purely model-driven strategy was adopted.

Moreover, Fig. \ref{fig:punch_res_width_a} reports the IPI index also for the model-data-driven case at the maximum load level. Increasing the load above the maximum one reached with data-driven simulations (that is, outside the training region), the needed numbers of iterations increases, around by a factor of 2. This is expected and justified by the fact that, outside the training region, the tangent matrix correction takes values significantly different from the consistent tangent of the stress function. This mismatch is the reason for the final divergence.

\begin{figure}[tbp]
\centering
\subfigure[Last converged load (left) and numerical performance as in Eq. \eqref{eq:IPI} (right)]{\includegraphics[width=0.98\textwidth]{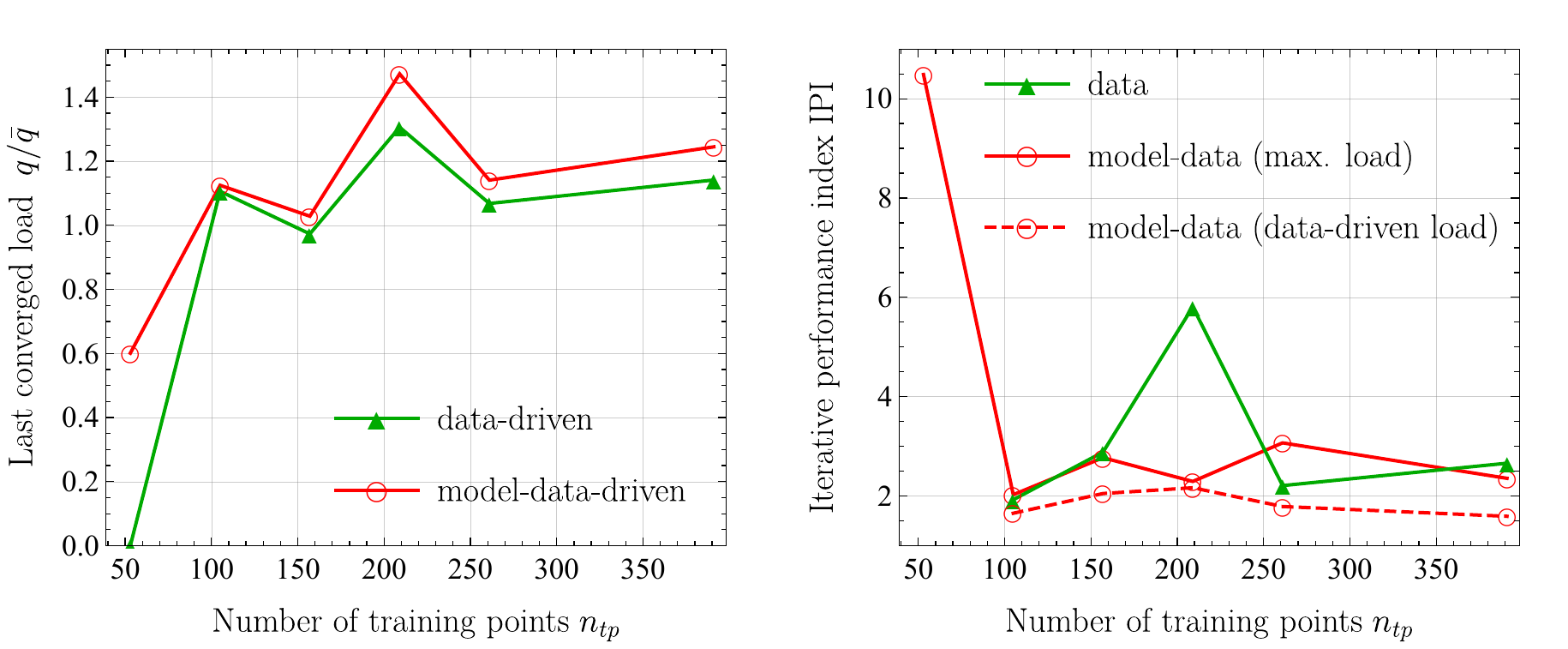}  \label{fig:punch_res_dens_a}}
\subfigure[Variation of displacement $v_1$ with respect to the reference case $n_{tp}=209$ (i.e., $n_{lay}=8$)]{\includegraphics[width=0.98\textwidth]{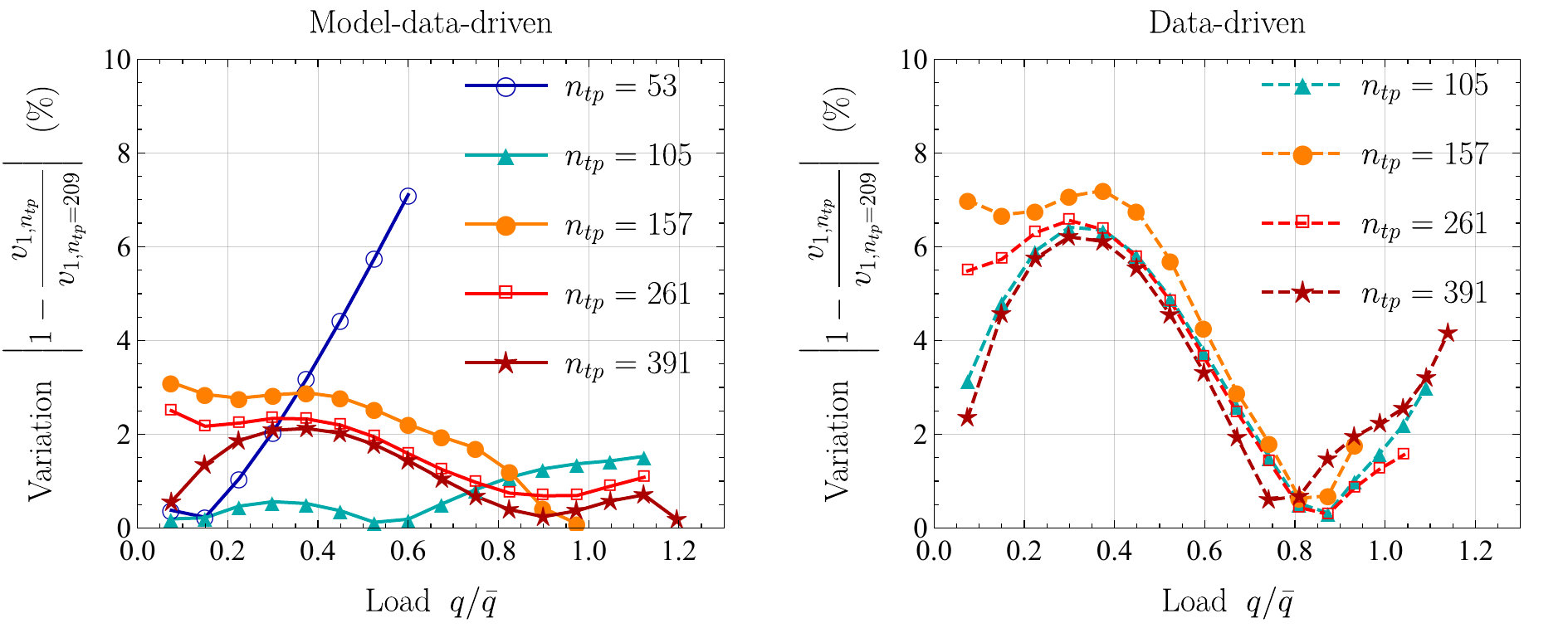}  \label{fig:punch_res_dens_b}}
\caption{Compression test: training region with constant width and varying density (\emph{constant width case} in Section \ref{par:specs_hybrid}). Results investigate on: a) numerical robustness and performance; b) sensitivity with respect to training region definition, from which accuracy can be evaluated. Simulation settings: mesh density $n_{el}=30$; target maximum load equal to $1.5\bar{q}$; adaptive load step algorithm with minimum number of steps equal to 20.}\label{fig:punch_res_dens}
\end{figure}

An additional case study is considered in Fig. \ref{fig:punch_res_dens} which presents results obtained at constant width and varying density (\emph{constant width case} in Section \ref{par:specs_hybrid}). Also in this case, a model-data-driven strategy outperforms a data-driven one. In truth, above a minimum threshold value for $n_{tp}$, the two approaches show similar values for the last converged load, as well as the iterative performance index (see Fig. \ref{fig:punch_res_dens_a}). However, as shown in Fig. \ref{fig:punch_res_dens_b}, the variation of results with respect to the reference case (with $n_{tp}=209$, that is $n_{lay}=8$) are significantly lower for the model-data-driven approach than for the data-driven one in the entire range of loads. This proves an higher robustness of numerical results with respect to training dataset settings, and hence higher accuracy.

In conclusion, the obtained results clearly prove that numerical performance and accuracy of model-data-driven simulations are significantly less sensitive to training width size than a purely data-driven approach. 

\subsubsection{Compression test: spatial discretization convergence}

\begin{figure}[tbp]
\centering
\includegraphics[width=0.98\textwidth]{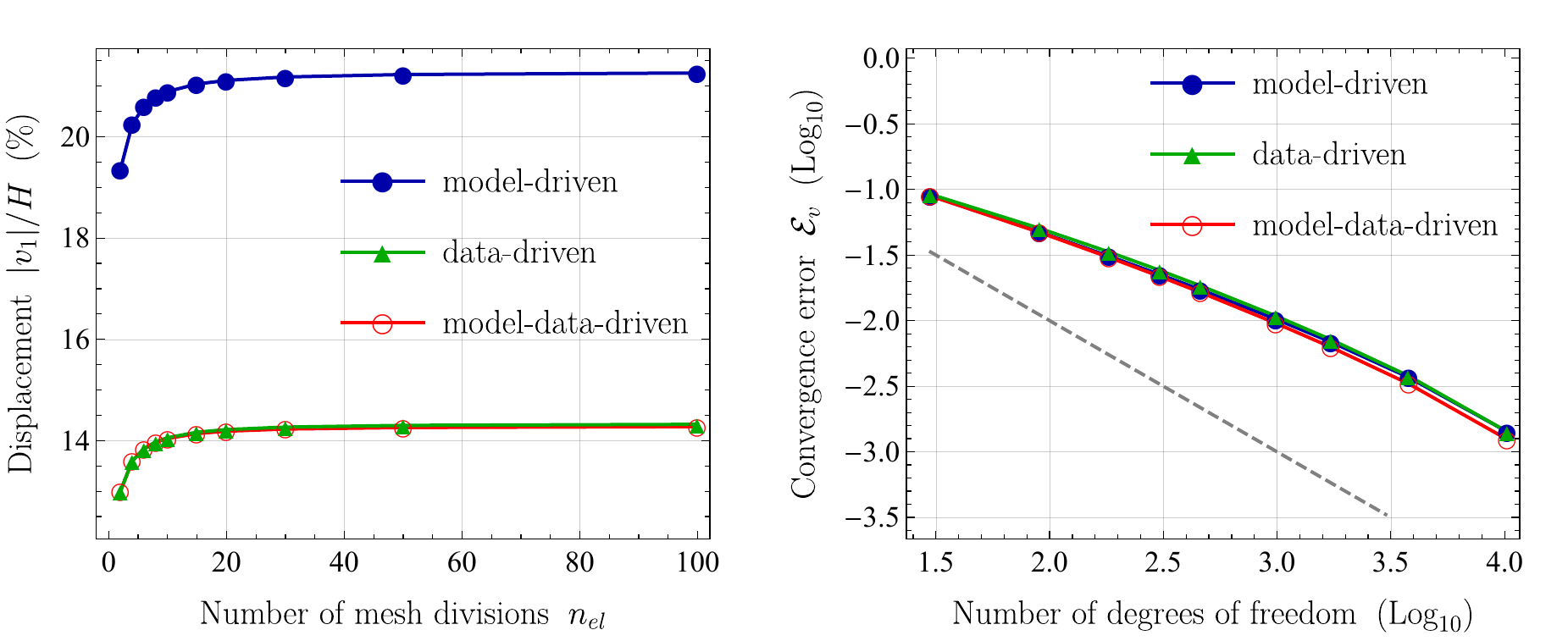}
\caption{Compression test: convergence behavior for increasing mesh refinement. Left: displacement $v_1$ normalized with respect to specimen width $H$ versus the number of mesh divisions $n_{el}$. Right: displacement convergence error $\mathcal{E}_v$ (right) versus the total number of degrees of freedom. Simulation settings: target maximum load equal to $\bar{q}$; fixed step algorithm with 10 load steps.}\label{fig:punch_res_convergence}
\end{figure}

Next, numerical performance is  investigated in terms of the convergence behaviour at different spatial discretizations. The question is if the adopted model-data-driven strategy (or data-driven one) affects the rate of convergence related to mesh refinements. This is analyzed in Fig. \ref{fig:punch_res_convergence}, which shows displacement $v_1$ and the convergence error $\mathcal{E}_v$ for different mesh densities. Error $\mathcal{E}_v=|v_1/v_{1,\infty}-1|$ compares the vertical displacement $v$ at point $P_1$ with respect to value $v_{1,\infty}$ obtained from overkill solutions with mesh density equal to $n_{el}=100$ and the same constitutive modelling approach. 

The results demonstrate that the rate of convergence of the hybrid strategy is identical to the one obtained with the model-driven approach, both being close to the expected theoretical one\begin{footnote}{The rate of convergence is not identical to the theoretical value of one since the plotted error is computed at a single point, and it is not the $L_2$-norm.}\end{footnote}. A purely data-driven strategy would also give the same rate of convergence.

\subsection{Model-data-driven structural response: Cook's membrane} \label{par:Cook_res}

\begin{sidewaysfigure}[tbp]
\centering
\includegraphics[width=0.85\textwidth]{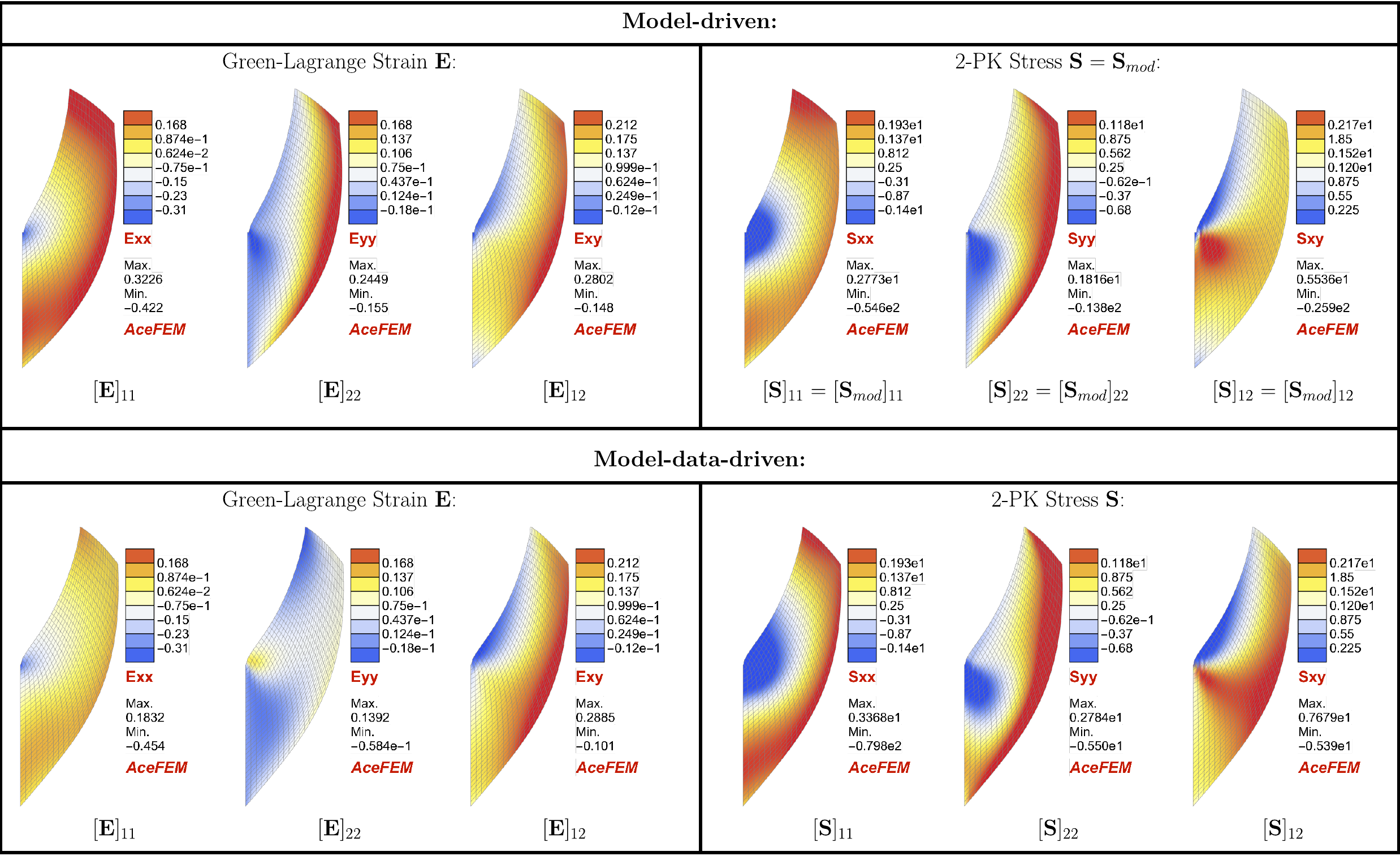}
\caption{Cook's membrane: structural response obtained with a model-driven approach and a model-data-driven approach ($\Delta_{tr}=15\%$ with $n_{lay}=6$). Distribution of Green-Lagrange strain ${\bf E}=({\bf C}-{\bf I})/2$ and Second Piola-Kirchhoff stress ${\bf S}$ tensors within the domain. Simulation settings: target maximum load equal to $\bar{q}$; fixed step algorithm with 20 load steps. Results are shown in the final deformed configuration.}
\label{fig:cook_res_maps}
\end{sidewaysfigure}

\begin{figure}[tbp]
\centering
\includegraphics[width=0.95\textwidth]{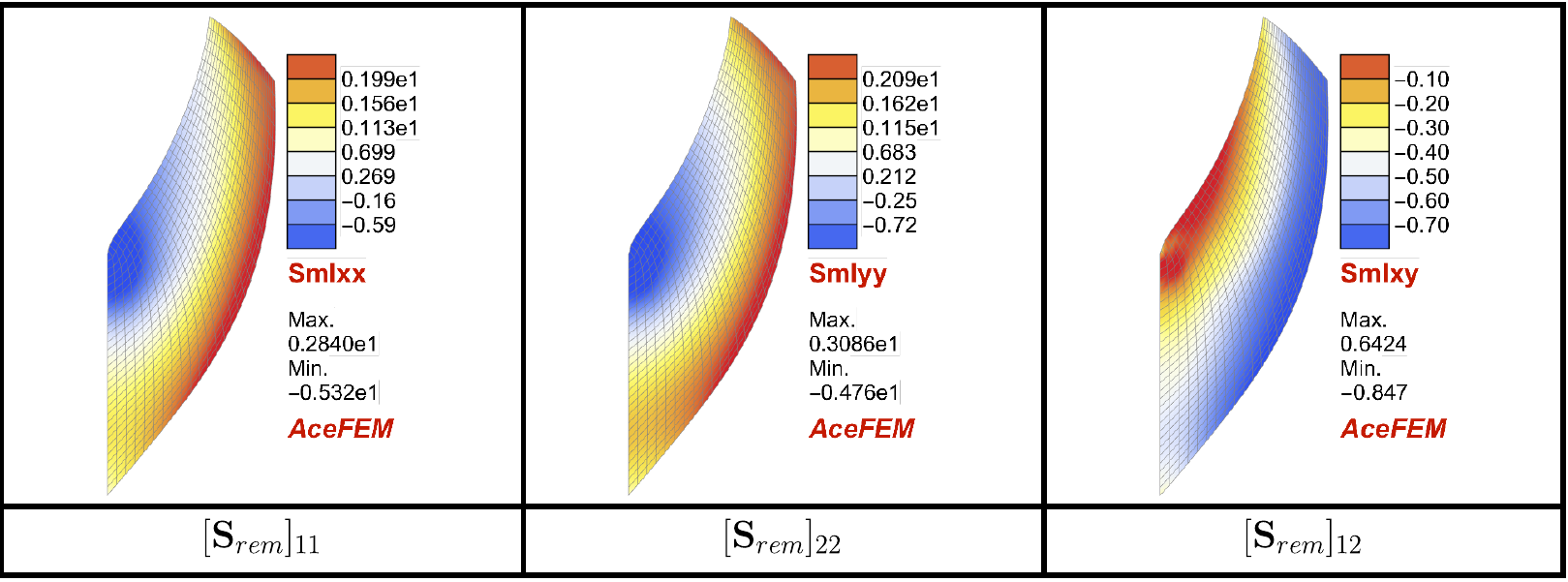}
\caption{Cook's membrane: distribution of the remainder component of the Second Piola-Kirchhoff stress ${\bf S}_{rem}$ introduced by the data-driven metamodel within the domain. Simulation settings: training dataset characterized by $\Delta_{tr}=15\%$ with $n_{lay}=6$; target maximum load equal to $\bar{q}$; fixed step algorithm with 20 load steps. Results are shown in the final deformed configuration.}
\label{fig:cook_res_maps_rem}
\end{figure}

The structural response obtained for  the Cook's membrane is reported in Fig. \ref{fig:cook_res_maps} which shows that, also in this case, the model-driven response is significantly different from the model-data-driven one. The stress remainder component introduced by the metamodel is reported in Fig. \ref{fig:cook_res_maps_rem}, showing that the data-driven correction is significant within the whole domain.

The considerations traced for the compression test with different training settings are generally confirmed for this case study, although some differences arise due to differences in the internal deformation. These will be commented next. 

\subsubsection{Cook's membrane: different training settings at constant density}

\begin{figure}[tbp]
\centering
\subfigure[Displacement $v_1/H$ versus load $q/\bar{q}$]{\includegraphics[width=0.98\textwidth]{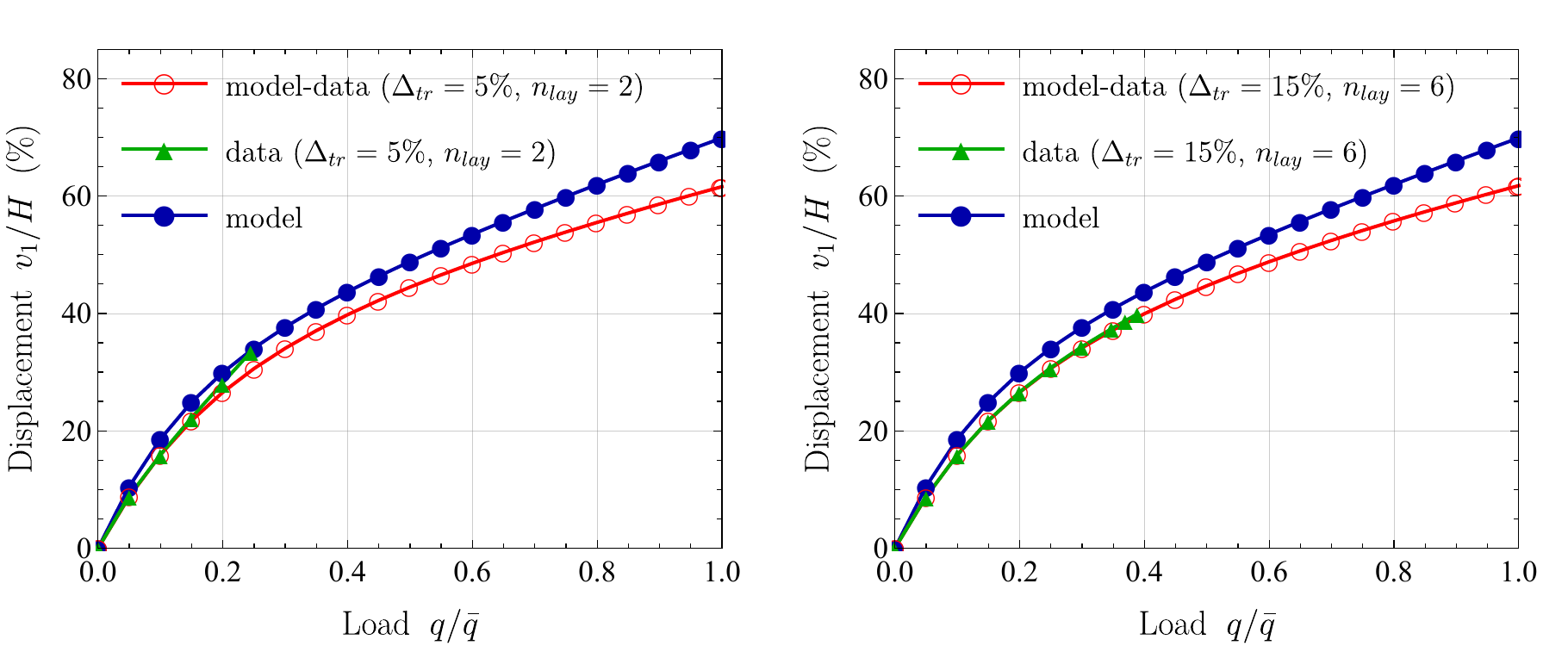} \label{fig:cook_res_comparison_a}}
\subfigure[CG deformation components in the domain versus load $q/\bar{q}$: mean values (line with symbols); standard deviations (error bars); and maximum/minimum values (dashed lines) within the domain. The training region is also reported as the gray shaded area.]{\includegraphics[width=0.98\textwidth]{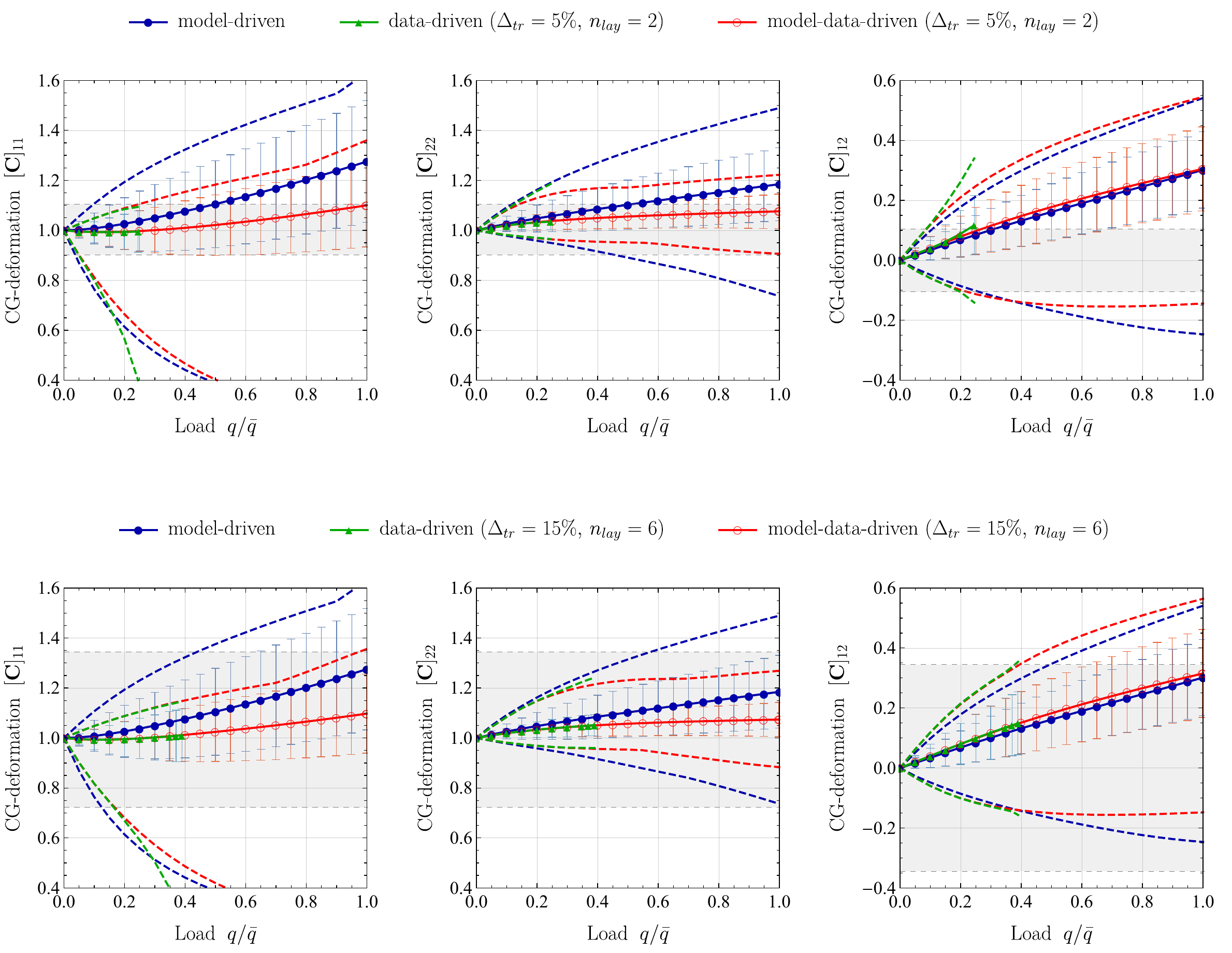}  \label{fig:cook_res_comparison_b}}
\caption{Cook's membrane: structural response obtained with two different training regions, that is $\Delta_{tr}=5\%$ with $n_{lay}=2$ and $\Delta_{tr}=15\%$ with $n_{lay}=6$ (varying width at constant density). Results show: a) the obtained macroscopic structural behaviour; b) components of the Cauchy-Green (CG) deformation tensor ${\bf C}$ within the domain. Simulation settings: target maximum load equal to $\bar{q}$; adaptive load step algorithm with minimum number of steps equal to 30.}
\label{fig:cook_res_comparison}
\end{figure}

\begin{figure}[tbp]
\centering
\subfigure[Last converged load (left) and numerical performance as in Eq. \eqref{eq:IPI} (right)]{\includegraphics[width=0.98\textwidth]{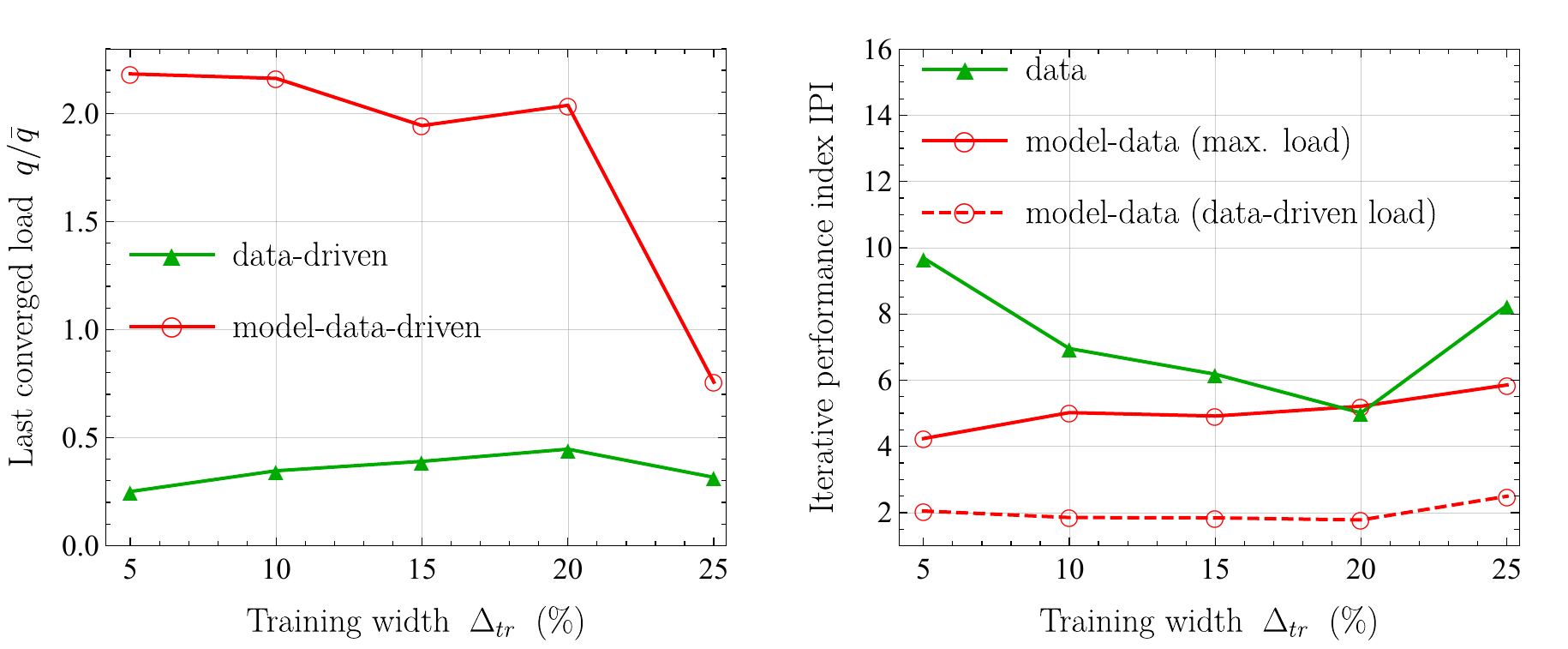} \label{fig:cook_res_width_a}}
\subfigure[Variation of displacement $v_1$ with respect to the reference case $\Delta_{tr}=20\%$]{\includegraphics[width=0.98\textwidth]{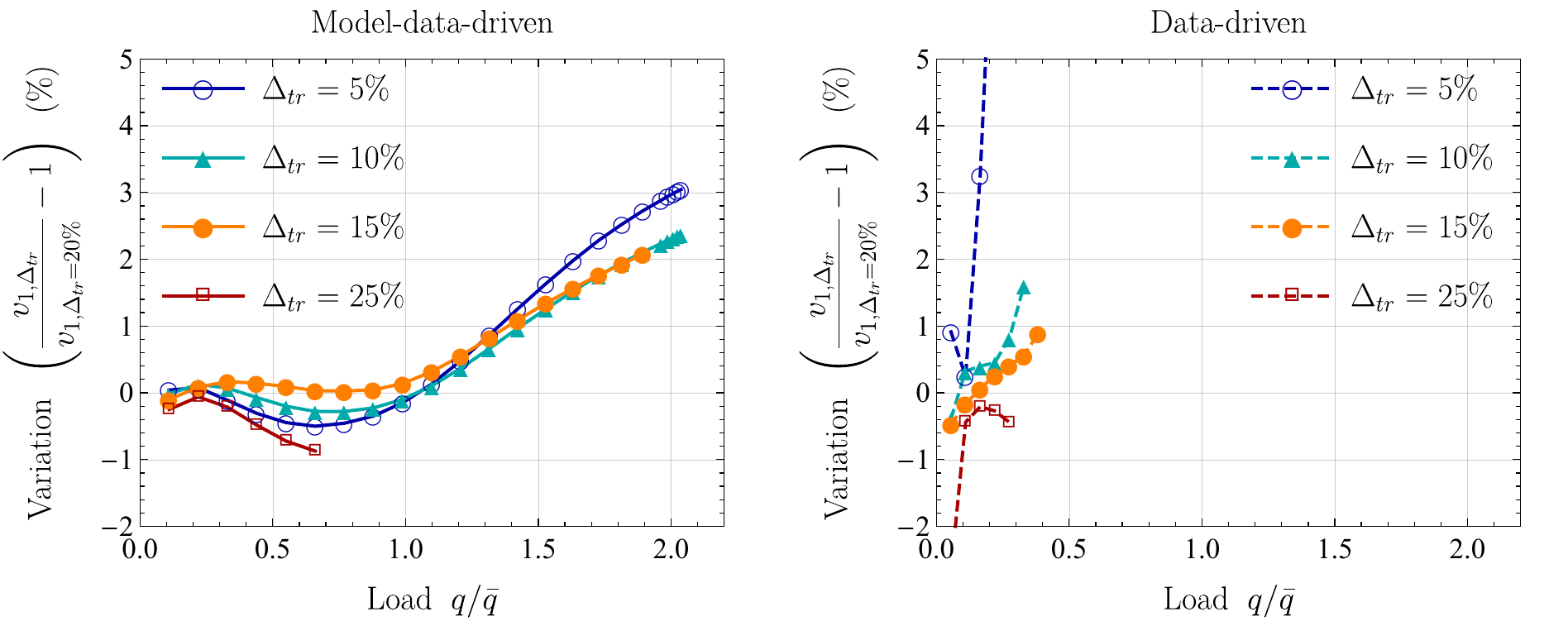} \label{fig:cook_res_width_b}}
\caption{Cook's membrane: training region with constant density and varying width (\emph{constant density case} in Section \ref{par:specs_hybrid}). Results investigate on: a) numerical robustness and performance; b) sensitivity with respect to training region definition, from which accuracy can be evaluated. Simulation settings: target maximum load equal to $2.2\bar{q}$; adaptive load step algorithm with minimum number of steps equal to 20}
\label{fig:cook_res_width}
\end{figure}

The Cook's membrane behaviour obtained with different training regions is shown in Fig. \ref{fig:cook_res_comparison} and compared with the solution obtained with data-driven strategies. The figure reports the relationship between the applied load $q$ and the vertical displacement $v_1$ in point $P_1$, as well as the distribution of internal deformations. In addition, Fig. \ref{fig:cook_res_width} presents the quantitative assessment of robustness and accuracy in the \emph{constant density case} (see Section \ref{par:specs_hybrid}). In this case, the outperformance of the model-data-driven strategy on the data-driven one is more significant than in the previously-analyzed compression test, both in terms of maximum applied load and numerical performance (i.e., IPI index).  Remarkably, the model-data-driven strategy converges for high loads also for small training widths and results do not vary significantly, at least for loads $q<1.2\bar{q}$. Interestingly, the metamodel associated with the widest training width $\Delta_{tr}=25\%$ diverges at load levels significantly lower than smaller training widths. As previously noticed from the error measurements on the constitutive behaviour (see Fig. \ref{fig:var_error_width_dens_results}) and confirmed by results' variation in Fig. \ref{fig:cook_res_comparison_b}, this is due to an increase of inaccuracies in the small strain region.

\subsubsection{Cook's membrane: different training settings at constant width}

\begin{figure}[tbp]
\centering
\subfigure[Last converged load (left) and numerical performance as in Eq. \eqref{eq:IPI} (right)]{\includegraphics[width=0.98\textwidth]{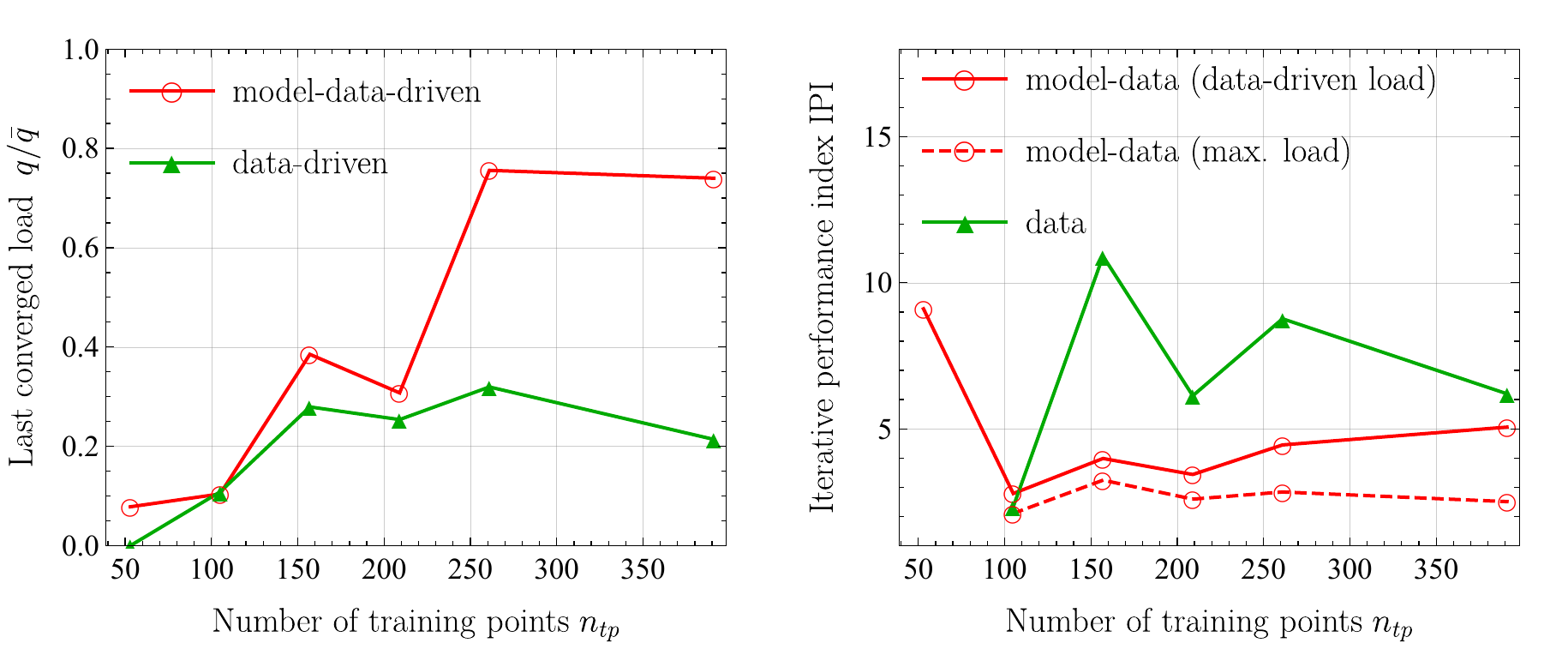}  \label{fig:cook_res_dens_a}}
\subfigure[Variation of displacement $v_1$ with respect to the reference case $n_{tp}=261$ (i.e., $n_{lay}=10$)]{\includegraphics[width=0.98\textwidth]{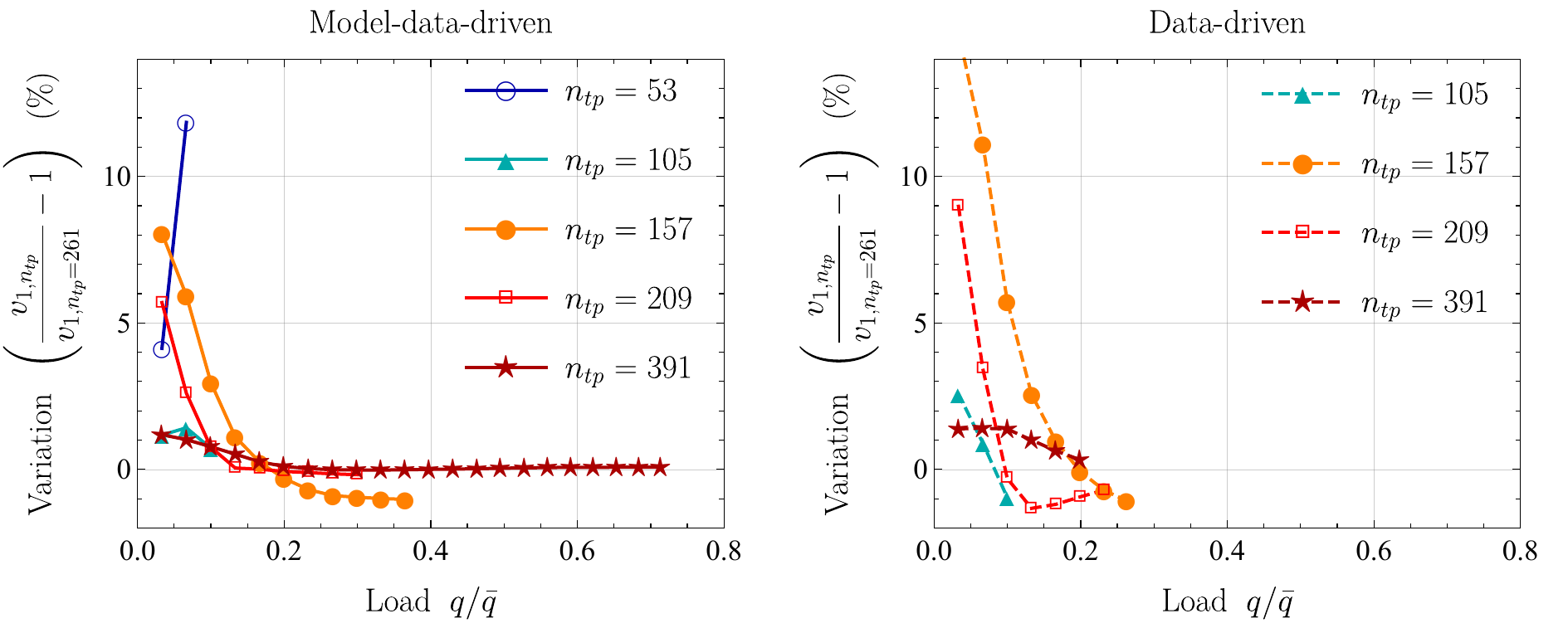}  \label{fig:cook_res_dens_b}}
\caption{Cook's membrane: training region with constant width and varying density (\emph{constant width case} in Section \ref{par:specs_hybrid}). Results investigate on: a) numerical robustness and performance; b) sensitivity with respect to training region definition, from which accuracy can be evaluated. Simulation settings: target maximum load equal to $\bar{q}$; adaptive load step algorithm with minimum number of steps equal to 30.}
\label{fig:cook_res_dens}
\end{figure}

\begin{figure}[tbp]
\centering
\includegraphics[width=\textwidth]{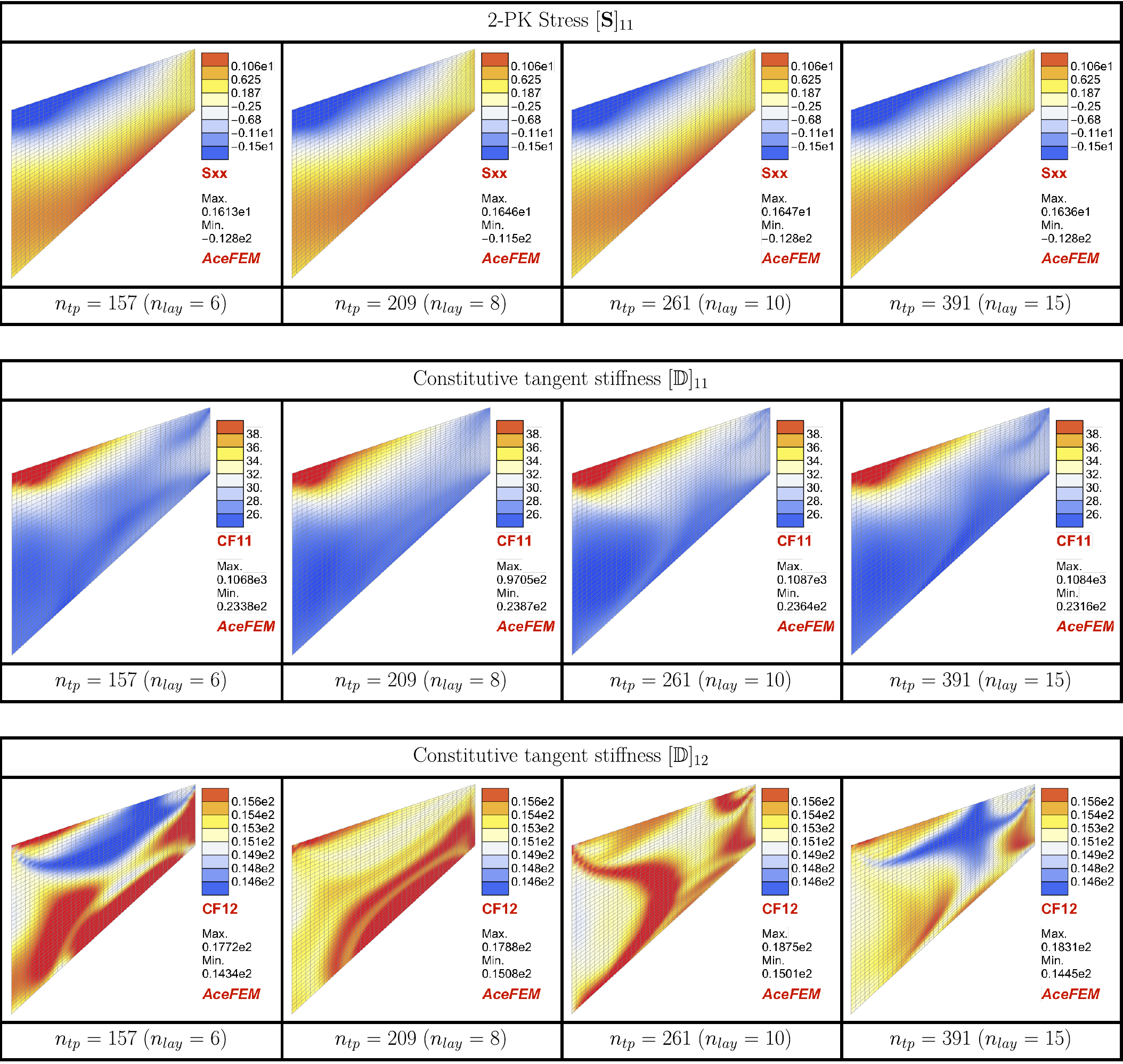}
\caption{Cook's membrane: distribution of representative components of the Second Piola-Kirchhoff stress ${\bf S}$ and the constitutive tangent stiffness $\mathbb{D}$ tensors within the domain obtained with model-data-driven strategies characterized by a training width of $\Delta_{tr}=25\%$ and different densities of sampling points (i.e., different number of layers $n_{lay}$ and training points $n_{tp}$). Simulation settings: mesh density $n_{el}=40$; target maximum load equal to $0.25\bar{q}$; fixed step algorithm with 10 load steps. Results are shown in the initial undeformed configuration. Full set of results for this case study are shown in \ref{sec:app_results}.}
\label{fig:cook_res_maps_comparison}
\end{figure}

Robustness and accuracy obtained in the \emph{constant width case} (see Section \ref{par:specs_hybrid}) for the Cook's membrane problem are shown in Fig. \ref{fig:cook_res_dens}. Generally, an increase of sampling density corresponds to an increase of the maximum reached load. An exception is represented by the settings with $n_{lay}=8$, that is $n_{tp}=209$. The source of divergence, explaining differences in the behaviour of the different settings, has been found in variations of metamodel correction in some components of the tangent constitutive matrix $\mathbb{D}$. In fact, distribution and values of second Piola-Kirchhoff stress tensor components ${\bf S}$ obtained with different settings are mainly coincident, and the same holds true for $\mathbb{D}$ apart especially component $[\mathbb{D}]_{12}$ which highly varies case-by-case. This is shown in Fig. \ref{fig:cook_res_maps_comparison}, which presents representative outcomes on ${\bf S}$ and $\mathbb{D}$ for different training settings at constant width. For the sake of completeness, the full set of results is reported in \ref{sec:app_results} where the same outcomes from the compression test are reported. From these results, it can be noted that $[\mathbb{D}]_{12}$ varies also for the compression test, but with variations of  minor extent. This is the reason why the solution of the compression test was able to reach high loads also with $n_{lay}=8$ (i.e., $n_{tp}=209$). In general, these considerations indicate that the performance of metamodels built from data-driven strategies are highly case dependent. However, as a final remark, it is noteworthy that the model-data-driven strategy confirms also here to be more robust and accurate than a purely data-driven one.

\section{Conclusions} \label{sec:conclusions}

This paper presents a model-data-driven computational multiscale approach that leverages data from the micro scale to enhance the accuracy of classical material models. Training points for the correction of the model-driven response through data-driven strategies are generated by means of a nonlinear computational homogenization procedure. With regard to the data-driven component, a machine learning technique based on Ordinary Kriging is chosen for upscaling microscale information at the macroscale. Ordinary Kriging is employed due to its proficient performance for relatively small data sets, probabilistic convergences and exact interpolation capabilities at input sample points. A coherent method to generate the dataset in a given strain range is also proposed. 

The obtained model-data-driven constitutive relationship is highly accurate, linked to microstructural features, and numerically efficient. As a matter of fact, the presented model always yields physical responses, accurately correcting the constitutive response within the training region, also far away from existing samples. Furthermore, the strategy is implemented in a finite element framework for the analysis of the nonlinear structural response in large deformations. Results demonstrate that the presented method is highly versatile, accurate and robust. Model-data-driven simulations inherit the level of information of FE$^2$-approaches, at the computational cost of purely model-driven simulations. Accordingly, this strategy potentially makes computational multiscale approaches more widely applicable.

The proposed approach is different from data-driven strategies due to the fundamental role of the modelling part. As a matter of fact, the model-data-driven strategy outperforms data-driven approaches and allows to benefit from decades of scientific discovery in the field of model-driven constitutive descriptions. It is shown that, with a suitable choice of the constitutive model, the size of the training dataset can be significantly reduced, while obtaining an excellent fit. This is relevant since massive amount of information might not be available in the context of material modelling. For instance, data needs to be generated from the output of numerical simulations which is computationally demanding. Even though the number of possible simulation evaluations is technically not limited, a lower number of needed samples for the metamodel is not only highly beneficial in terms of productivity but also in terms of making the data set \lq\lq{}realistically\rq\rq{} obtainable. In contrast to big data, this approach is called smart data, see e.g. \cite{chinesta2018virtual}. In addition, thanks to the modelling component, model-data-driven simulations are robust even when the deformation state lies outside the considered training region. Indeed, while data-driven strategies should not be trusted outside the training region, the error of a model-data-driven strategy remains low for a wide range of deformation exceeding the training region limits. Finally, the width of the training region of the machine learning component can be properly designed on the basis of preliminary simulations relying only on the modelling component, given that the latter provides an acceptable approximation.

%The proposed methodology finds common motivations with the work by Chinesta and co-workers \cite{Gonzales2019a,Gonzalez2019b,Ibanez2019}, but significantly differs in methods and technicalities. It also has similarities with 

In future works, an adaptive enlargement of the training space could be considered, see e.g. \cite{fuhg2020state}. In this regard, since the strategy developed in this work for the definition of the training dataset is based on an incremental approach, it is suitable to be employed for incorporating a strain input outside the current training domain. The framework could be also extended to incorporate multi-fidelity kriging, where different simulation tools (converged and non-converged results, or different level of simulations) are combined in a unique kriging metamodel, see for instance \cite{Nachar2019}. These improvements will allow to extend the proposed implementation to three-dimensional problems, to extend the proposed approach to time-dependent processes, and to consider more than two length-scales.

\bibliography{bibMLpaper}

\clearpage
%%%%%%%%%%%

\newpage

\appendix

\setcounter{figure}{0}
\renewcommand{\thefigure}{A\arabic{figure}}

\section{Parameter identification for the macroscale model} \label{sec:app_optimal}

In order to consistently investigate the importance of the modelling component, the best-fitting value $C_1^{fit}$ of parameter $C_1$ in Eq. \eqref{eq:macro_model} is looked for on the basis of RVE microstructural responses. Firstly, reference RVE homogenized constitutive responses are obtained by addressing a series of three numerical tests with boundary conditions defined as (see Eqs. \eqref{eq:ubc} and \eqref{eq:F_app_micro}):
\begin{subequations} \label{eq:NH_res_Ftests}
\begin{align}
& 1) \;  {\bf F}_{11}(F_{11}) = {\bf F}_{app}(F_{11},1,0) \, , \quad F_{11} \in [0.75,1.25]\, ,\\
& 2) \;{\bf F}_{22}(F_{22})  = {\bf F}_{app}(1,F_{22},0) \, , \quad F_{22} \in [0.75,1.25]\, , \\
& 3) \; {\bf F}_{12}(F_{12}) =  {\bf F}_{app}(1,1,F_{12}) \, , \quad F_{22} \in [-0.25,0.25]\, .
\end{align}
\end{subequations}

The macroscale model parameter $C_1^{fit}$  is estimated as the optimal value minimizing the difference between ${\bf S}_{\text{RVE}}$ and ${\bf S}_{mod}$, where the latter is function of $C_1$, namely  ${\bf S}_{mod} = {\bf S}_{mod}({\bf C}; C_1)$. The following objective function $f_{obj}(C_1)$ is introduced and minimized through Wolfram Mathematica:
\begin{equation*}
f_{obj}(C_1) = \sum_{ij}  \sum_{k}  \|{\bf S}_{\text{RVE}}({\bf C}_{ij,k}) - {\bf S}_{mod}({\bf C}_{ij,k}; C_1)\|\, ,
%f_{obj}(C_1) = \sum_{ij}  \sum_{k}  \|({\bf S}_{nh}({\bf C}_{ij,k}) - {\bf S}_{mod}({\bf C}_{ij,k})|_{C_1}):({\bf e}_i \otimes {\bf e}_j)\|\, ,
\end{equation*}
where ${\bf C}_{ij,k} = ({\bf F}_{ij}(\lambda_{k}))^T {\bf F}_{ij}(\lambda_{k})$ is the right Cauchy-Green deformation tensor associated with the stretch tensor ${\bf F}_{ij}$ ($ij = 11$, 22, 12), computed at 10 equispaced stretch values $\lambda_{k}$ in the ranges introduced in Eqs. \eqref{eq:NH_res_Ftests} with $k=1,\ldots,20$. 

\clearpage
\newpage
\setcounter{figure}{0}
\renewcommand{\thefigure}{B\arabic{figure}}

\section{Auxiliary results on the definition of the training set}\label{sec:appendix_tr}

As introduced in Section \ref{sec:data_specs}, the single input ${\bf c}_i$ is defined from a value for the deformation gradient ${\bf F}$ of the form of a (symmetric) right stretch tensor. However, even considering a plane strain condition, possible combinations of the 3 input values of the deformation gradient ($F_{11}$, $F_{22}$ and $F_{12}$=$F_{21}$) are defined over the whole real space and are therefore unbounded, that is $\mathbb{X}\equiv \mathbb{R}^3$ in general. Of course, in practice, they vary by a certain amount around the unstressed configuration. However, this consciously undefined \lq\lq{}amount\rq\rq{} depends on the application of interest and is generally only known in tendencies and not in effective strain values. Hence, estimating the input range for which a given machine learning model yields valid results is crucial. An indicator for this range is given by the value domain covered by the training data of the model. Outside of this training space, the tool has no information about the mapping between input and output data, can only extrapolate, and should generally not be trusted. Apart from the width of the training region, other factors that might affect the accuracy of the hybrid model-data driven approach are the position of the sampling points within the training region both in relative terms (i.e., their distance) and in absolute terms (i.e., their configuration).

A short study is provided that assesses the fitting capabilities of models generated with different choices on the position of the sampling points. In detail, LHD and grid configurations are compared in \ref{sec:app_LHD_grid} and training regions with different number of sample points are employed in \ref{sec:app_enlarge}. Finally, \ref{sec:app_incr} presents an incremental approach that provides a systematic way of sampling new points to enlarge the training region. 

The study, conducted at the single material point level, introduces an analytical expression for the strain-energy density $\Psi_{el}({\bf C})$ defining the actual material micromechanical response of the RVE:
\begin{equation} \label{eq:analytical_law}
\Psi_{el}({\bf C}) = c_1 I_1 +c_2 I_2 + c_3 I_3 + \frac{c_4}{2} (I_1 - 3)^2 - 2 c_5 \text{Log}(\sqrt{I_3})\, ,
\end{equation}
where $I_2 = \text{Cof}({\bf C})$ with $\text{Cof}(\cdot)$ denoting the cofactor operator. Parameters values are chosen equal to $c_1 = 1000$ and $c_2 =  c_3 = c_4 = 100$. In order to ensure a stress-free condition in the reference configuration, i.e. ${\bf S}({\bf I}) = {\bf 0}$, it results $c_5 = 2 c_1 + 4 c_2 + 2 c_3 = 2600$. The model component of the hybrid model-data driven approach is chosen equal to the one in Eq. \eqref{eq:macro_model} with $C_1=1000$.

\subsection{LHD versus Grids}\label{sec:app_LHD_grid}

\begin{figure}%[tb]
\centering
\subfigure[Scenario 1 \label{fig:3Diff_sample_position_a}]{\includegraphics[width=0.32\textwidth]{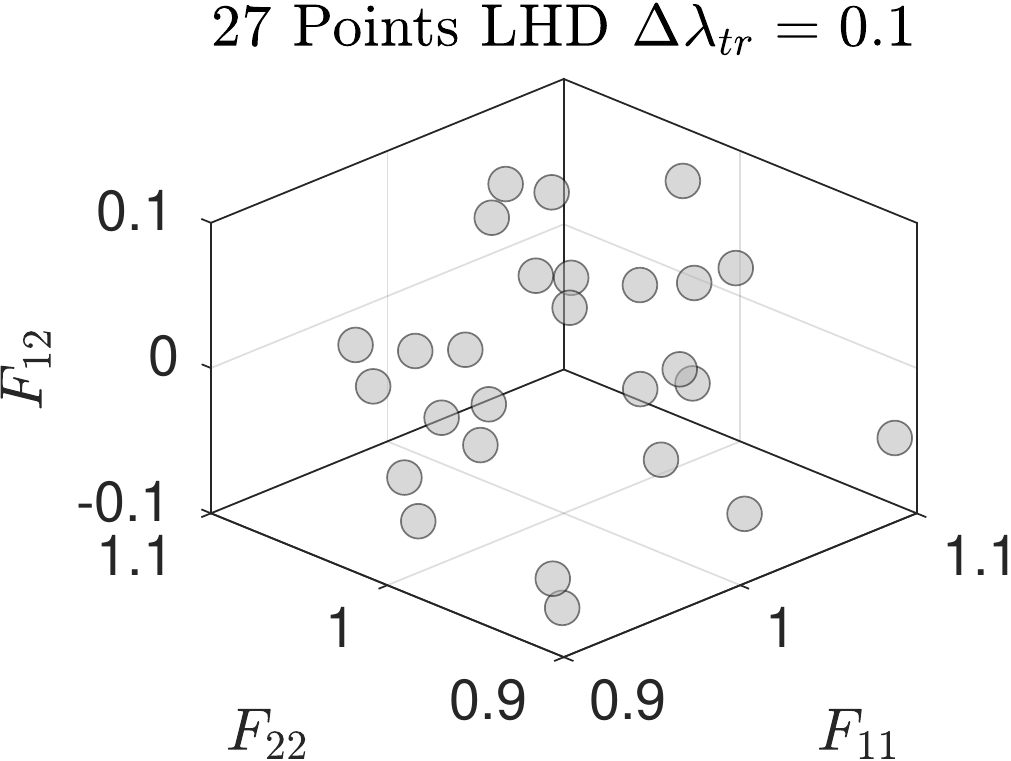}}%
\subfigure[Scenario 2 \label{fig:3Diff_sample_position_b}]{\includegraphics[width=0.32\textwidth]{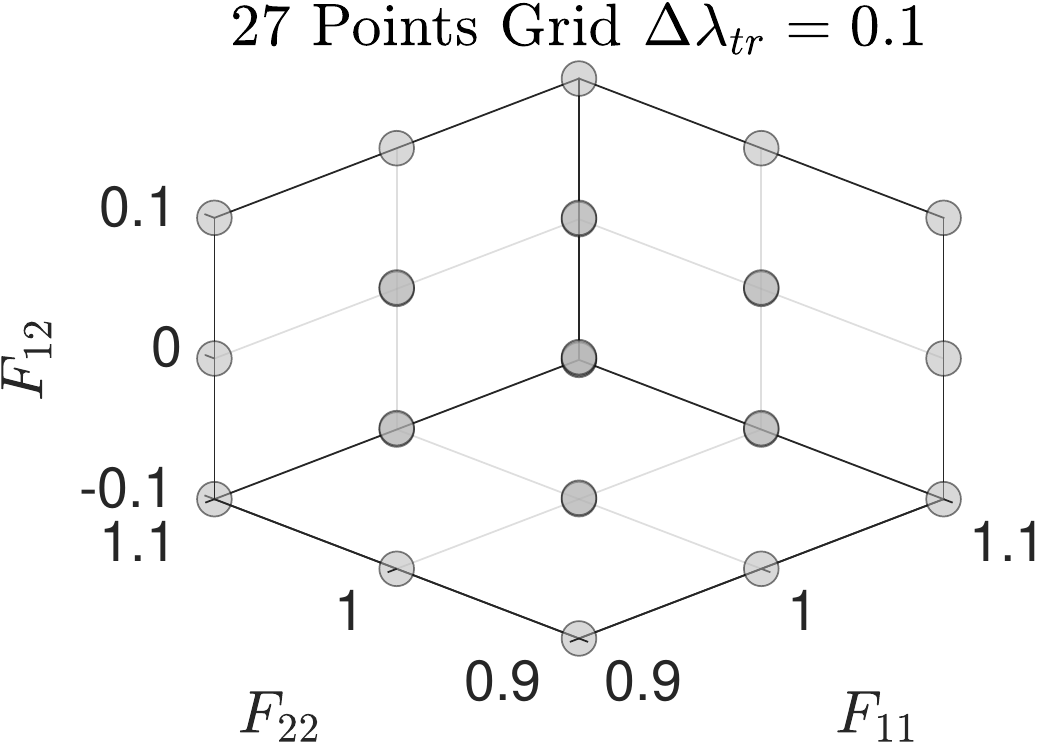}}
\subfigure[Scenario 3 \label{fig:3Diff_sample_position_c}]{\includegraphics[width=0.32\textwidth]{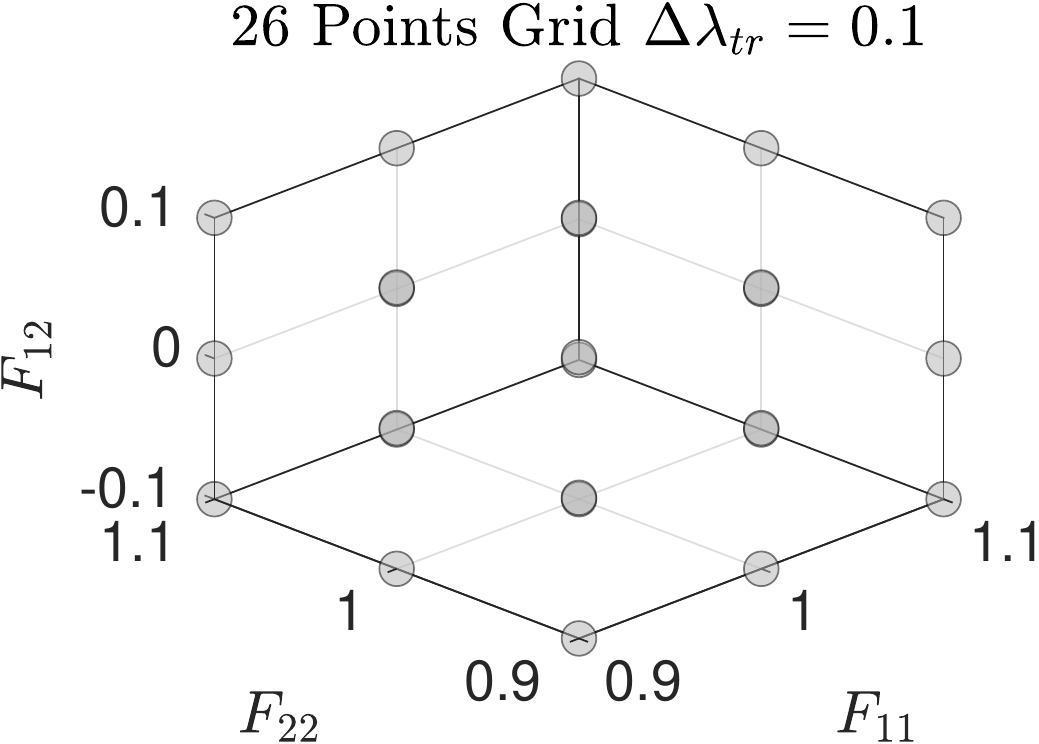}}
\subfigure[Scenario 4 \label{fig:2Diff_sample_positions_a}]{\includegraphics[width=0.3\textwidth]{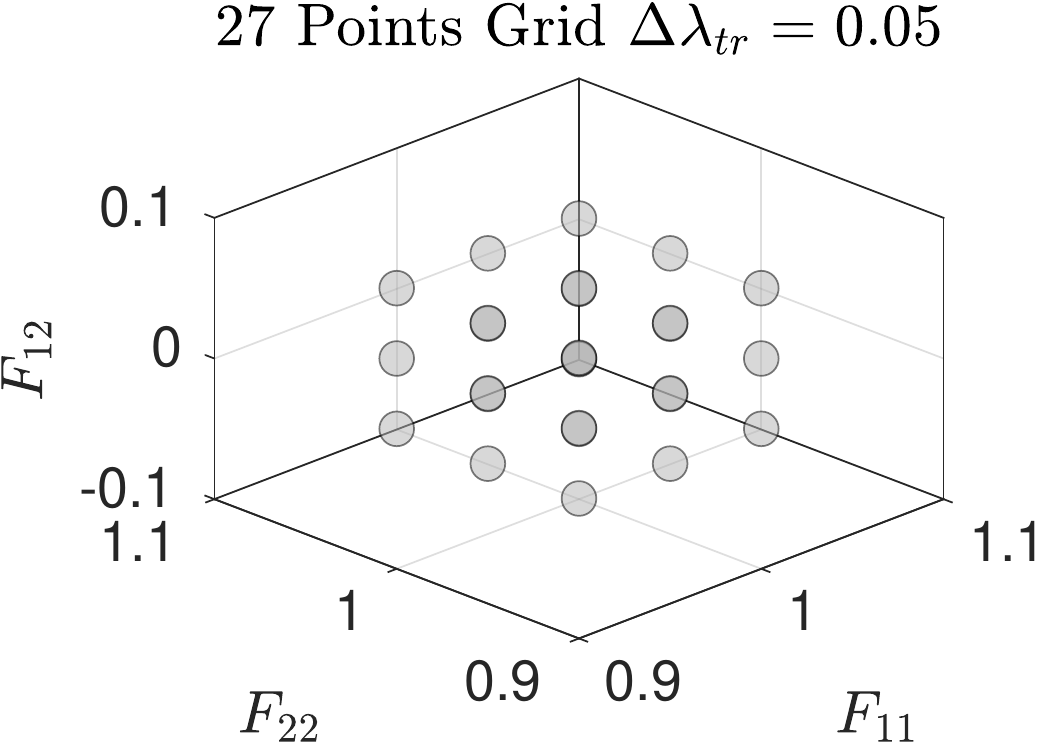}} 
\subfigure[Scenario 5 \label{fig:2Diff_sample_positions_b}]{\includegraphics[width=0.3\textwidth]{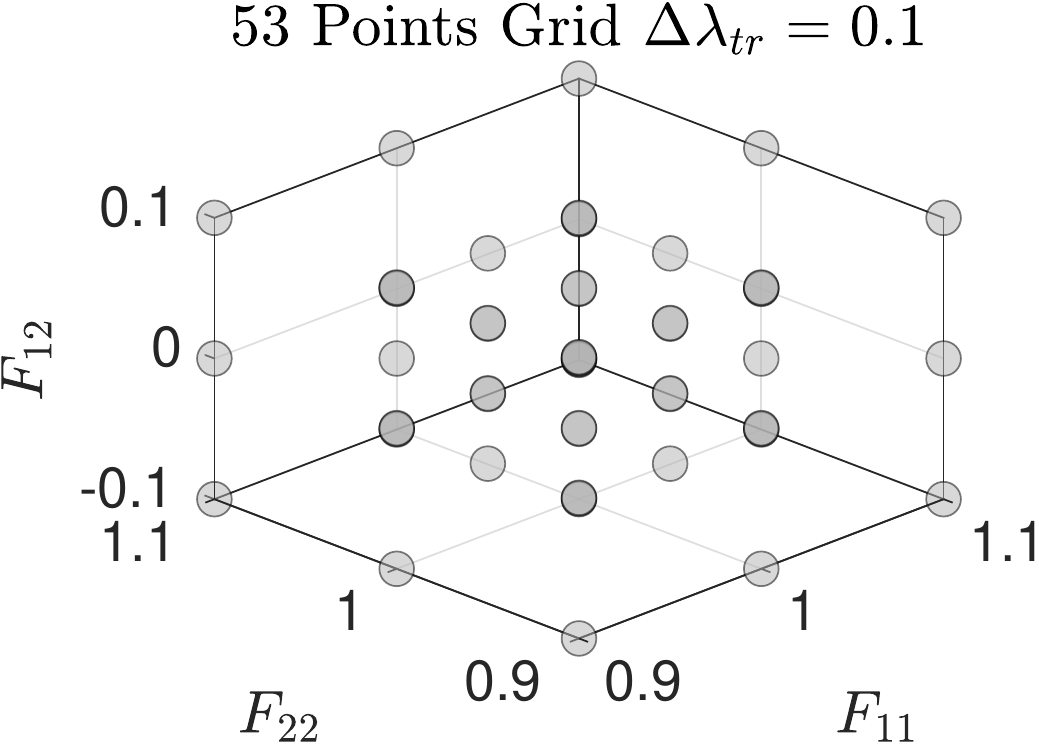}}
\caption{Five different sample scenarios. From (a) to (c): sample points in the training region $\Delta_{tr}=0.1$. In detail, LHD with $n_{tp}=27$ points (scenario 1), grid with $n_{tp}=27$ points and reference configuration (scenario 2), and grid with $n_{tp}=26$ points and no reference configuration (scenario 3). Sub-plots (d) and (e): training regions of half-width $\Delta_{tr}=0.05$ with $n_{tp}=27$ points (scenario 4) and $\Delta_{tr}=0.1$ with $n_{tp}=53$ points (scenario 5).}\label{fig:3Diff_sample_positions}
\end{figure}

Consider the three sample position scenarios in the training region $\Delta_{tr}$ as given in Figure \ref{fig:3Diff_sample_positions}(a) to \ref{fig:3Diff_sample_positions}(c). The first scenario (Figure \ref{fig:3Diff_sample_positions}(a)) is based on $n_{tp}=27$ samples generated using LHD, which (by chance) does not create a sample that represents the unstressed configuration. Figure \ref{fig:3Diff_sample_positions}(b) depicts the scenario which constructs a cubic grid with 26 vertices and a central sample corresponding exactly to the unstressed configuration, thus yielding $n_{tp}=27$ also in this case. The third scenario shown in Figure \ref{fig:3Diff_sample_positions}(c) has an equivalent cubic-grid based on $n_{tp}=26$ points but omits the central point. 

\begin{figure}[tb]
\centering
\subfigure[Stress-norm error]{\includegraphics[width=0.46\textwidth]{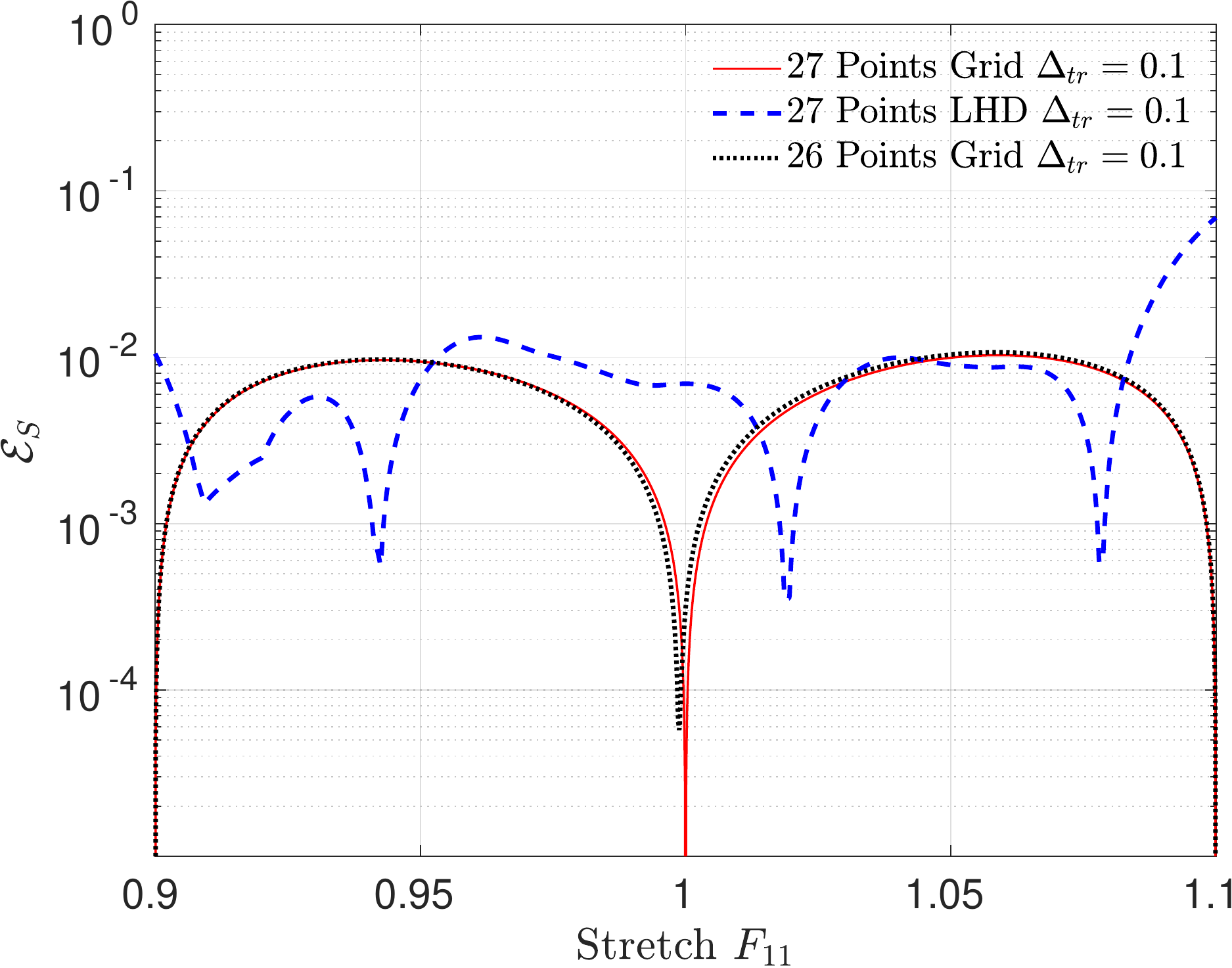} \label{fig:3Diff_sample_positions_errornorm_a}}%
\subfigure[Tangent-norm error]{\includegraphics[width=0.46\textwidth]{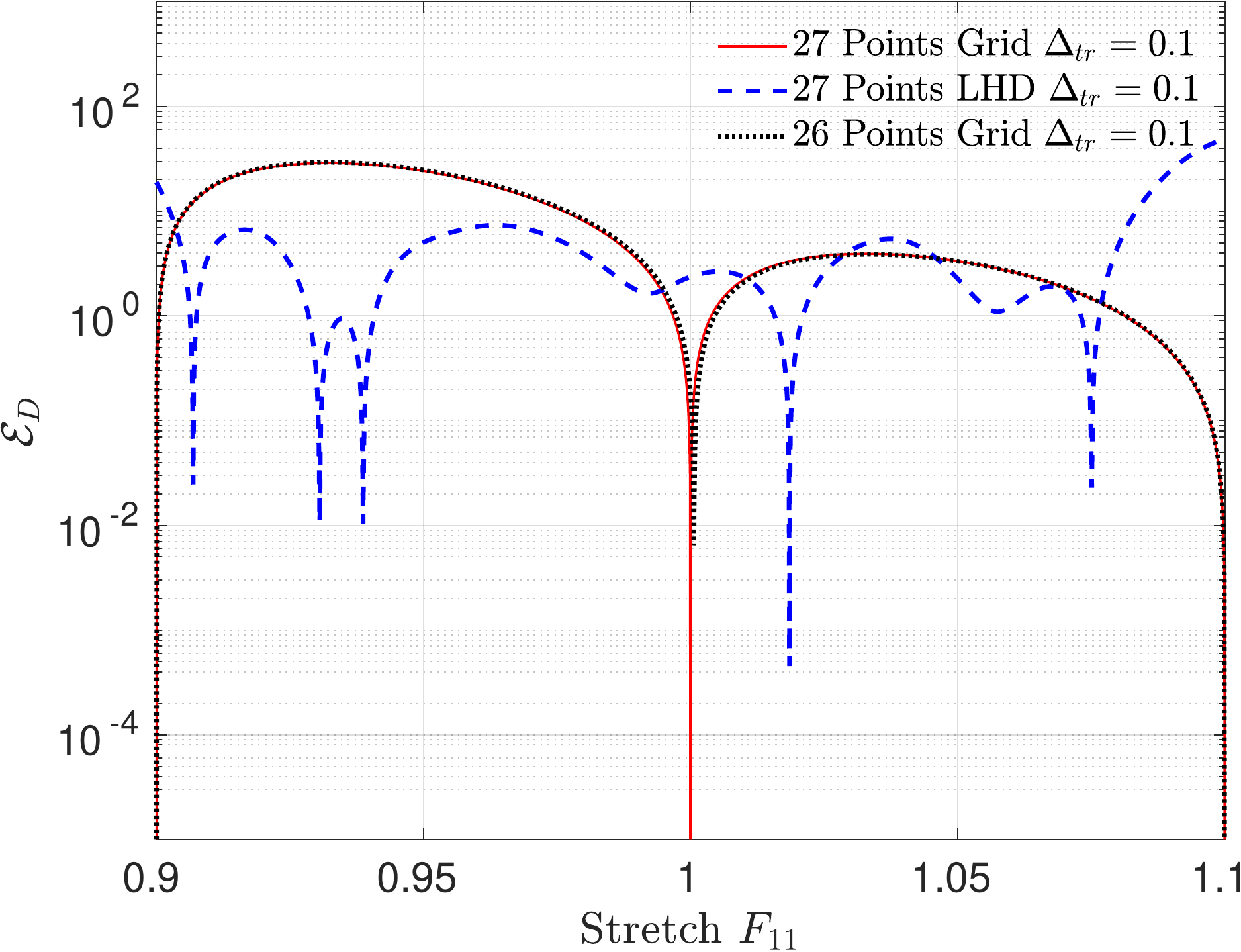} \label{fig:3Diff_sample_positions_errornorm_b}}
\caption{Errors around the unstressed configuration with $F_{22} = 1$ and $F_{12}=0$ for scenarios 1 to 3 in Figs. \ref{fig:3Diff_sample_positions}(a) to \ref{fig:3Diff_sample_positions}(c). The error measures in Eq. \eqref{eq:ErrorMetricis_a}, originally introduced for the multiaxial case, are here used in a slightly altered form where the control region is adapted to fit the uniaxial case, that is considering $\abs{F_{11} - 1} < \lambda_{c}$ in Eq. \eqref{eq:control_region}.}\label{fig:3Diff_sample_positions_errornorm}
\end{figure}

Figure \ref{fig:3Diff_sample_positions_errornorm} depicts the errrors between the true stress (Fig. \ref{fig:3Diff_sample_positions_errornorm}(a)) and stiffness tangent tensor(Fig. \ref{fig:3Diff_sample_positions_errornorm}(b)) and their respective predictions over the stretch $F_{11}$ around the unstressed configuration with $F_{22} = 1$ and $F_{12} = 0$. Three results can be noticed. First of all, the quality of the resulting metamodel is highly dependent on the choice of the dataset. Secondly, the two grid-like structures produce a more predictable, i.e. stable, error structure than the metamodel created from the LHD samples. Lastly, it can be observed that when a sample point is added at the unstressed configuration the physical quantities at this point are exactly fit. When this is not the case, like in scenario 3, the metamodel output does not yield an unstressed state at ${\bf F}  = {\bf I}$.   

\subsection{Enlargement of the training region} \label{sec:app_enlarge}

Consider now the two additional sampling scenarios given in Figure \ref{fig:3Diff_sample_positions}(d) and \ref{fig:3Diff_sample_positions}(e). For these two scenario, the grid-like structure with a central point is preserved. However, scenario 4 is generated in the training space of half-width $\Delta_{tr}=0.05$ with $n_{tp}=27$ points, whereas scenario 5 adds an additional layer of 26 samples around the points of scenario 4 to fill out the final training space of half-width $\Delta_{tr} = 0.1$, resulting in $n_{tp}=53$ points. 

%\begin{figure}[tb]
%\centering
%\subfigure[Scenario 4 \label{fig:2Diff_sample_positions_a}]{\includegraphics[width=0.3\textwidth]{figures/Grid_27_005.pdf}} 
%\subfigure[Scenario 5 \label{fig:2Diff_sample_positions_b}]{\includegraphics[width=0.3\textwidth]{figures/Grid_53_01.pdf}}
%%
%\caption{Two additional sample scenarios in the training regions of half-width $\Delta_{tr}=0.05$ (Scenario 4) and  $\Delta_{tr}=0.1$ (Scenario 5). \comm{Change labels: bigger and $\Delta_{tr}$ instead of $\Delta \lambda_{ML}$}}\label{fig:2Diff_sample_positions}
%\end{figure}

\begin{figure}[tbp]
\centering
\subfigure[Stress-norm error]{\includegraphics[width=0.46\textwidth]{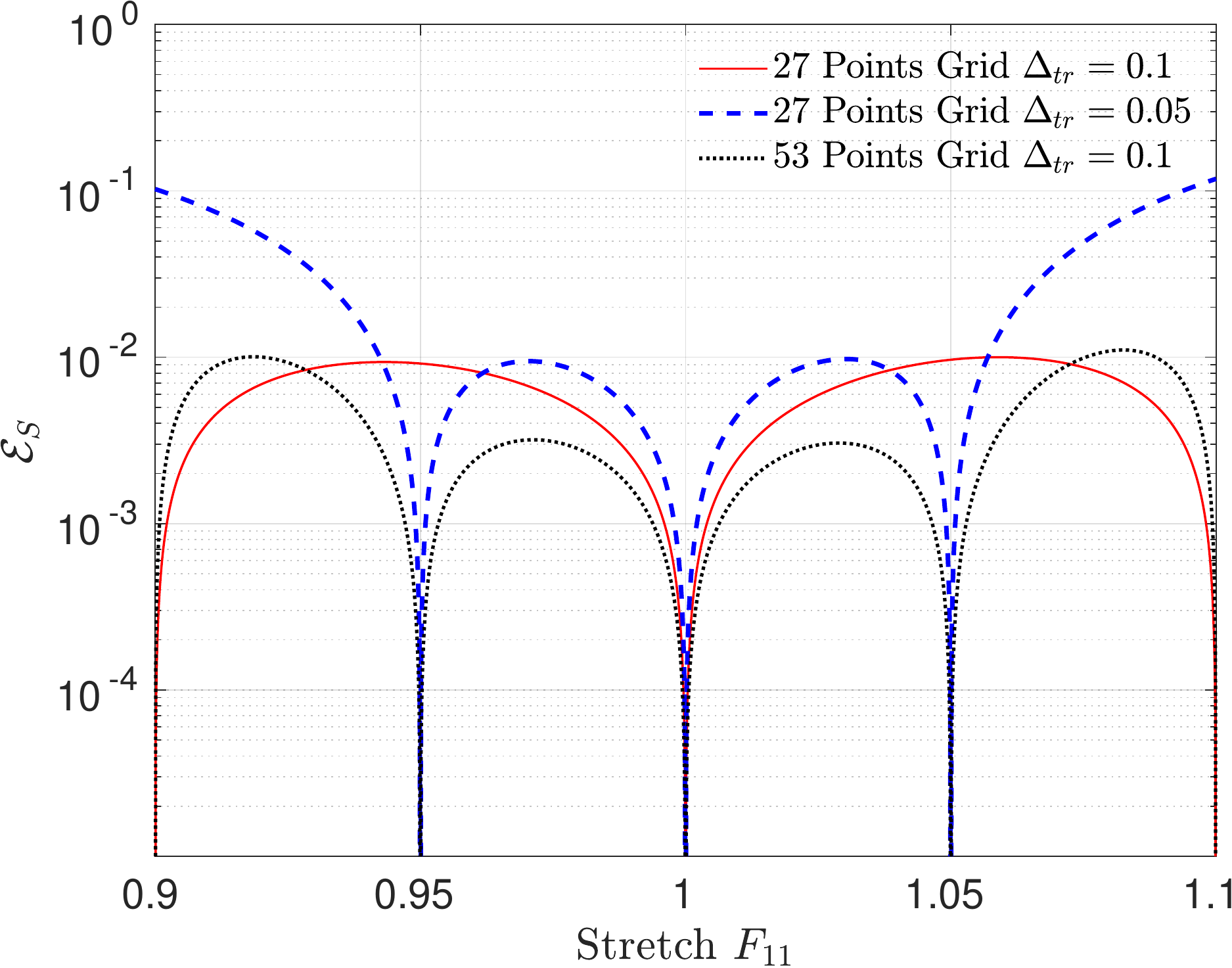} \label{fig:2Diff_sample_positions_errornorm_a}}%
\subfigure[Tangent-norm error]{\includegraphics[width=0.46\textwidth]{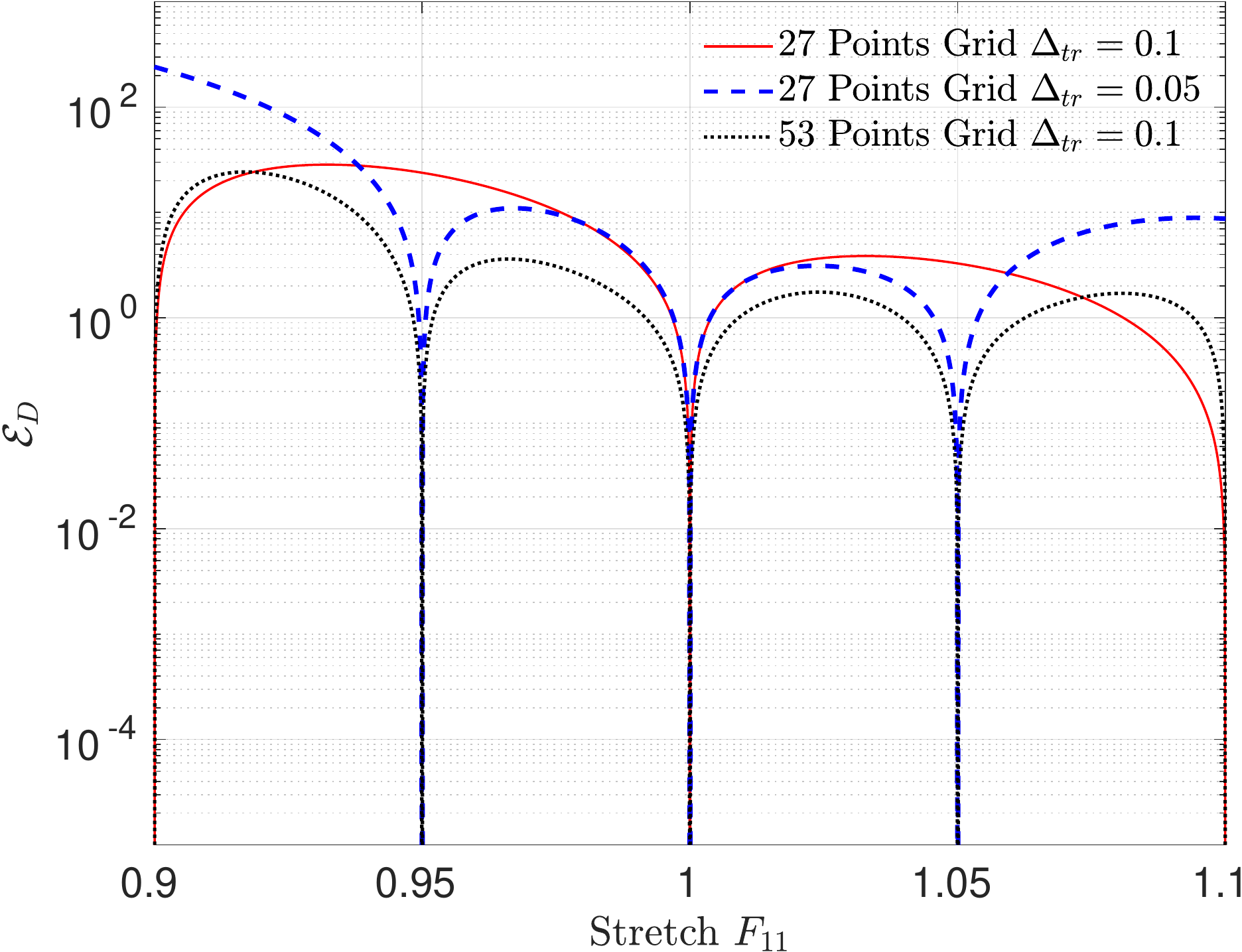} \label{fig:2Diff_sample_positions_errornorm_b}}
\caption{Errors around the unstressed configuration (varying $F_{11}$ with $F_{22} = 1$ and $F_{12}=0$) for sampling scenario 2 (Fig. \ref{fig:3Diff_sample_positions}(b)), scenario 4 (Fig. \ref{fig:3Diff_sample_positions}(d)), and scenario 5 (Fig. \ref{fig:3Diff_sample_positions}(e)). The error measures in Eq. \eqref{eq:ErrorMetricis_a}, originally introduced for the multiaxial case, are here used in a slightly altered form where the control region is adapted to fit the uniaxial case, that is considering $\abs{F_{11} - 1} < \lambda_{c}$ in Eq. \eqref{eq:control_region}.}\label{fig:2Diff_sample_positions_errornorm}
\end{figure}

In Fig. \ref{fig:2Diff_sample_positions_errornorm}, the errors between the true and predicted output of the stress (Fig. \ref{fig:2Diff_sample_positions_errornorm}(a)) and the stiffness tensor (Fig. \ref{fig:2Diff_sample_positions_errornorm}(b)) of these two sampling scenarios are compared to the sampling approach of scenario 2 (see Fig. \ref{fig:3Diff_sample_positions}(b)). It is noticeable that inside the training space of the respective metamodels the error behaviour is again nearly symmetric and rather predictable. However, it can be observed that the error increases considerably or looses its symmetric behaviour outside of the training space $\Delta_{tr}=0.05$ for scenario 4. As already mentioned this is an unavoidable but expected phenomenon.
Furthermore it can be seen that the error reduces when the cubic-grid structure consisting of 27 points is sampled in a smaller training space (c.f., scenarios 2 and 4). The addition of another 26 samples to the cubic-grid samples of scenario 4, as given by scenario 5, is able to preserve the nearly symmetric behaviour and also obtains a lower error than scenario 2 in all of the considered stretch domain. 

\subsection{Systematic generation of training regions: incremental strategy} \label{sec:app_incr}

\begin{figure}[tbp]
    \centering
\includegraphics[width=0.75\textwidth]{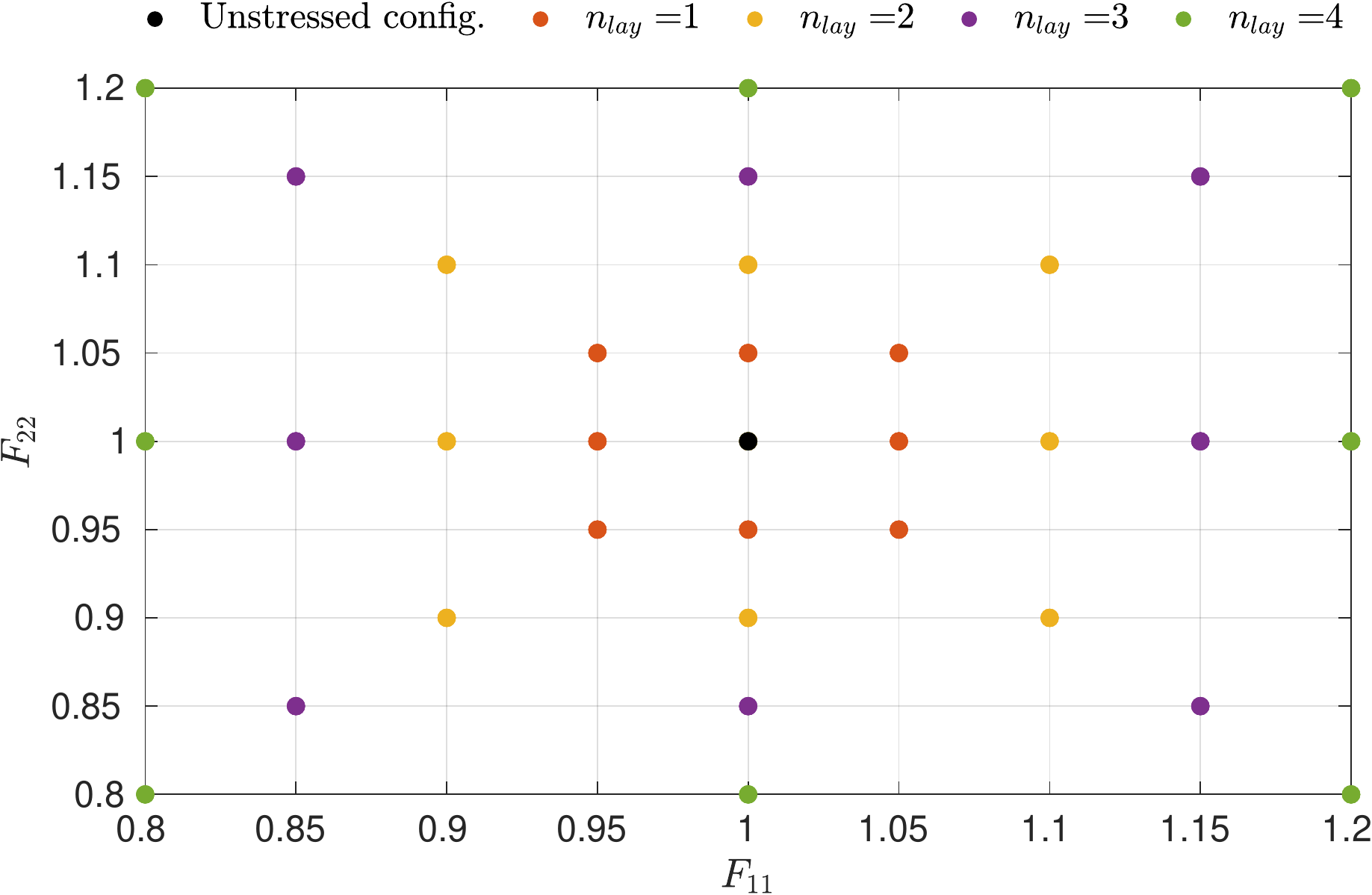}
\caption{Incremental generation of the training region $\mathcal{R}_{tr}$ and corresponding sample points projected into the $F_{11}-F_{22}$ plane with half-width $\Delta_{tr}=0.2$  and $n_{lay}=4$ layers (density $\delta_{tr}=\Delta_{tr}/n_{lay} = 0.05$). Different colors denote the 26 samples added in each layer and in the reference configuration. }
    \label{fig:Cubic26_Process}
\end{figure}

The afore-traced observations allow to develop a methodology to systematically generate sample points, analogously to the process from Scenario 4 to Scenario 5 in Fig. \ref{fig:2Diff_sample_positions_errornorm}. The training points are generated by combining $n_{lay}$ layers of 26 points. Each of the layers defines a cubic region of stretch-half-width $\Delta_{l}$ centered in the point defining the reference configuration (i.e., $F_{11}=1$, $F_{22}=1$ and $F_{12}=0$). In each cubic layer, 8 sample points  are positioned at the vertices, 12 points at the middle of each edge, and 6 points at the middle of each face of this cube. The width of each layer $l=1,\ldots, n_{lay}$ is defined as $\Delta_{l}=\Delta_{l-1} + \delta_{tr}$ with $\Delta_{0}=0$ and $\delta_{tr}=\Delta_{tr}/n_{lay}$ a measure of the density of sampling points in the training region. In order to ensure an accurate training in the proximity of the reference configuration, one sample point is added at $F_{11}=1$, $F_{22}=1$ and $F_{12}=0$, for a total of $n_{tp}=26 n_{lay} +1$ training points. As a notation rule, the training region is then characterized as $\mathcal{R}_{tr}=\mathcal{R}_{tr}(\Delta_{tr},n_{lay})$. The resultant sample point positions of this process for $\mathcal{R}_{tr}(0.2,4)$ can be seen in Figure \ref{fig:Cubic26_Process}.

%The values of $\Delta \lambda_{ML}^0$ and $D\lambda$ required to obtain effective metamodels are dependent on the complexity of the investigated material model and the structural application at hand. By means of preliminary finite element simulations (reproducing the patch test presented in the following Section \ref{par:patch_res}), it was found that, for the present applications, an initial $\Delta \lambda_{ML}^0=0.025$ (to obtain good convergence behavior around the unstressed configuration) and consecutive increases of $\Delta \lambda_{ML}$ with $D\lambda=0.05$ yield proficient metamodels. 

%The proposed strategy is then fully characterized by specifying: the initial training region $\Delta \lambda_{ML}^0$, the incremental enlargement width $D\lambda$, and the final width $\Delta \lambda_{ML}$, denoting it as $\text{US}(\Delta \lambda_{ML}^0,D\lambda)$. For instance, $\text{US}(0.025,0.05)$ would generate $n_{tp}=79$ samples in $\Delta \lambda_{ML} = 0.125$. 

\clearpage
\newpage

\setcounter{figure}{0}
\renewcommand{\thefigure}{C\arabic{figure}}

\section{Additional results with different training settings at constant width} \label{sec:app_results}

\begin{figure}[p]
\centering
\subfigure[$n_{tp}=53$  ($n_{lay}=6$)]{\includegraphics[width=0.98\textwidth]{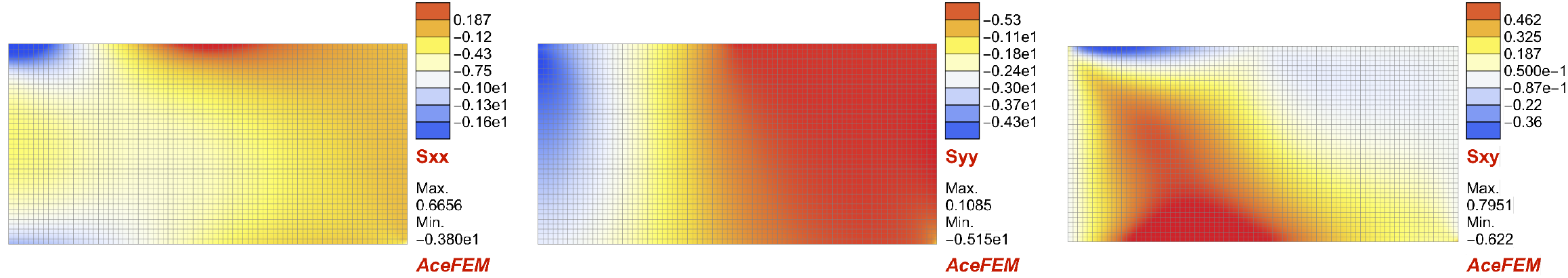}}
\subfigure[$n_{tp}=209$  ($n_{lay}=8$)]{\includegraphics[width=0.98\textwidth]{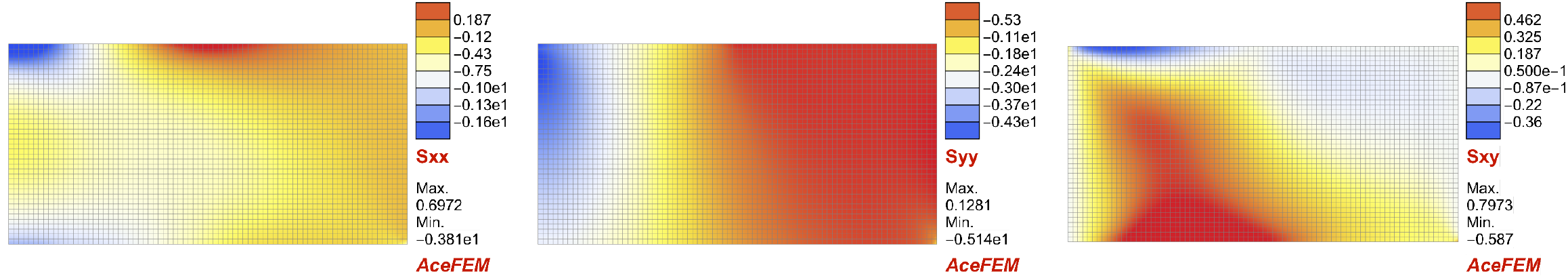}}
\subfigure[$n_{tp}=261$  ($n_{lay}=10$)]{\includegraphics[width=0.98\textwidth]{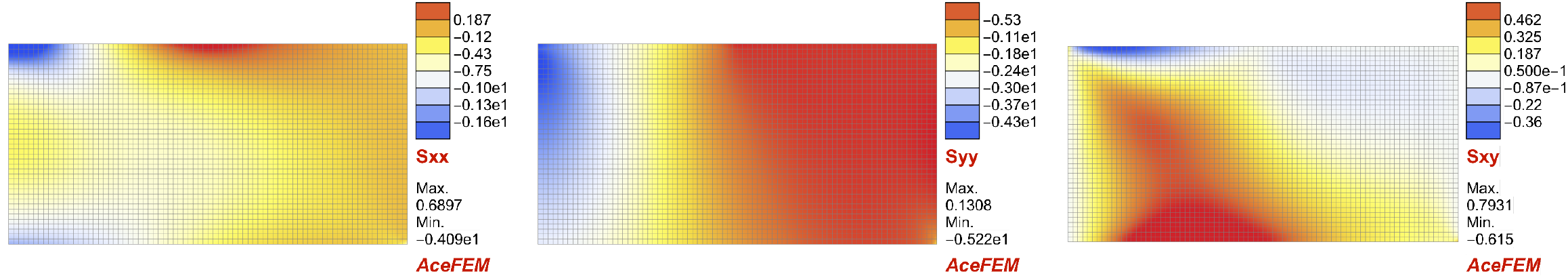}}
\subfigure[$n_{tp}=391$  ($n_{lay}=15$)]{\includegraphics[width=0.98\textwidth]{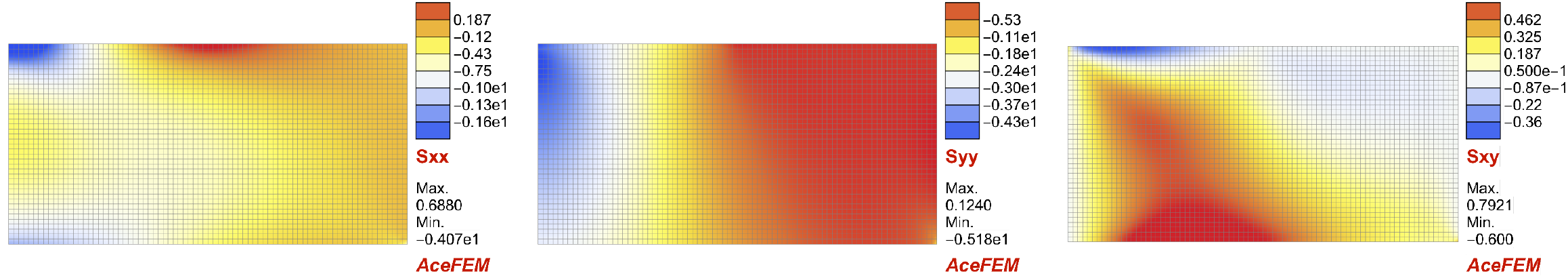}}
\caption{Compression test: distribution of components of the Second Piola-Kirchhoff stress tensor $[{\bf S}]_{11}$ (left), $[{\bf S}]_{22}$ (center) and $[{\bf S}]_{12}$ (right) within the domain obtained with model-data-driven strategies characterized by a training width of $\Delta_{tr}=25\%$ and different densities of sampling points (i.e., different number of layers $n_{lay}$ and training points $n_{tp}$).}
\label{fig:punch_S_comparison_all}
\end{figure}

\begin{sidewaysfigure}[p]
\centering
\includegraphics[width=\textwidth]{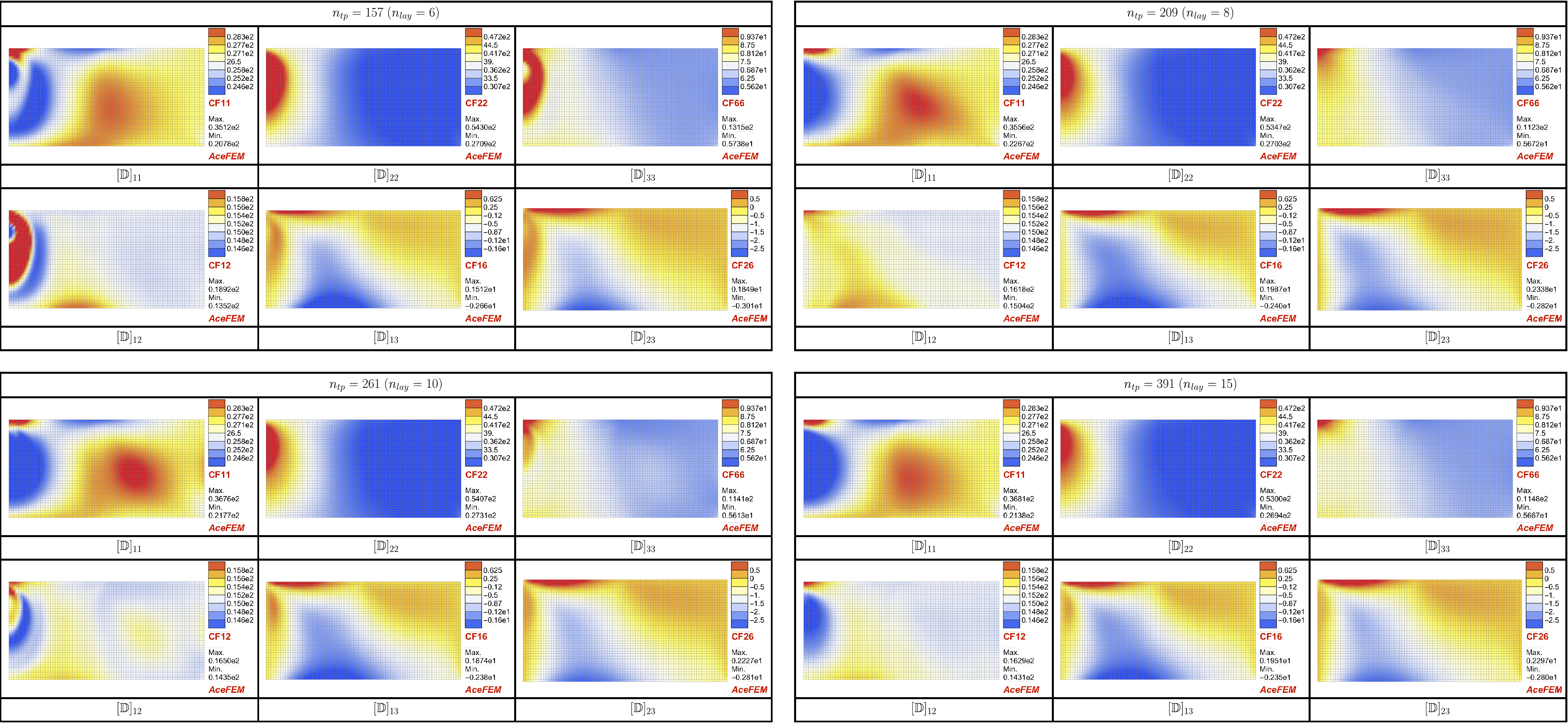}
\caption{Compression test: distribution of components of the material tangent constitutive stiffness tensor $\mathbb{D}$ within the domain obtained with model-data-driven strategies characterized by a training width of $\Delta_{tr}=25\%$ and different densities of sampling points (i.e., different number of layers $n_{lay}$ and training points $n_{tp}$).}
\label{fig:punch_C_comparison_all}
\end{sidewaysfigure}

\begin{figure}[p]
\centering
\subfigure[$n_{tp}=53$  ($n_{lay}=6$)]{\includegraphics[width=0.98\textwidth]{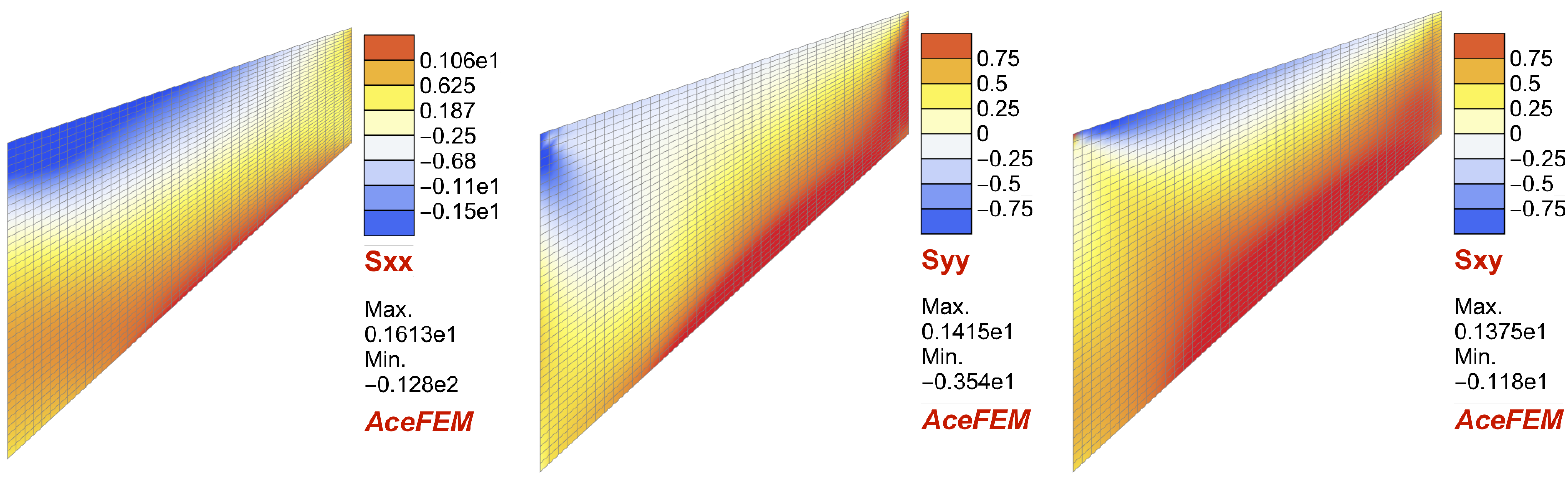}}
\subfigure[$n_{tp}=209$  ($n_{lay}=8$)]{\includegraphics[width=0.98\textwidth]{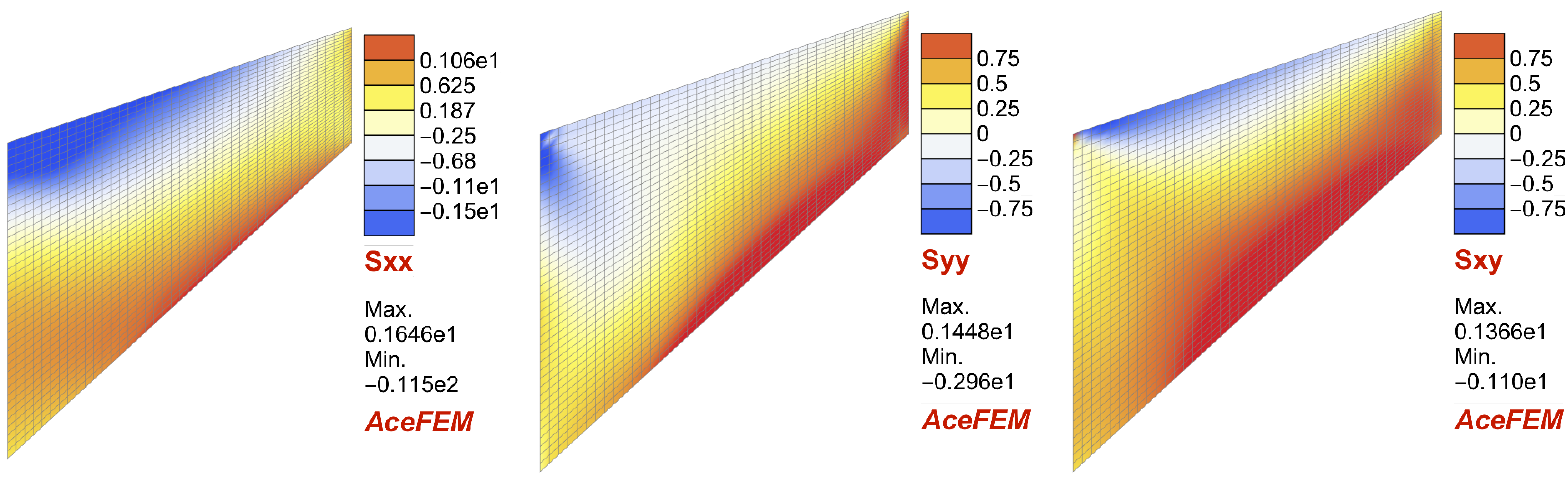}}
\subfigure[$n_{tp}=261$  ($n_{lay}=10$)]{\includegraphics[width=0.98\textwidth]{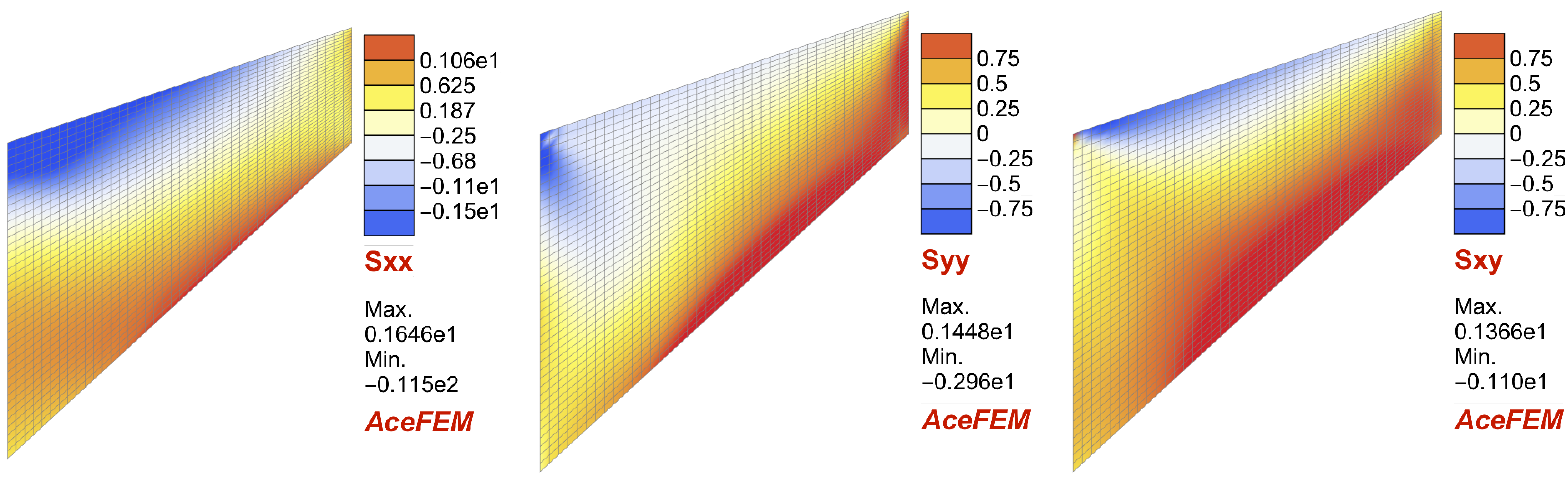}}
\subfigure[$n_{tp}=391$  ($n_{lay}=15$)]{\includegraphics[width=0.98\textwidth]{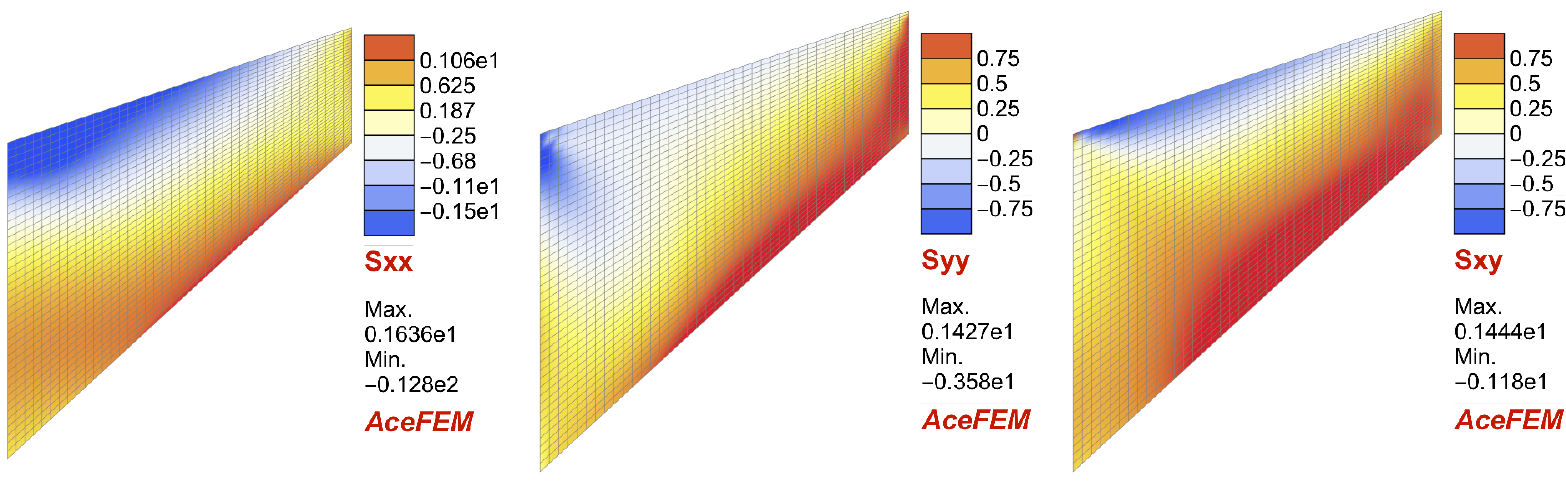}}
\caption{Cook's membrane: distribution of components of the Second Piola-Kirchhoff stress tensor $[{\bf S}]_{11}$ (left), $[{\bf S}]_{22}$ (center) and $[{\bf S}]_{12}$ (right) within the domain obtained with model-data-driven strategies characterized by a training width of $\Delta_{tr}=25\%$ and different densities of sampling points (i.e., different number of layers $n_{lay}$ and training points $n_{tp}$).}
\label{fig:cook_S_comparison_all}
\end{figure}

\begin{sidewaysfigure}[p]
\centering
\includegraphics[width=0.9\textwidth]{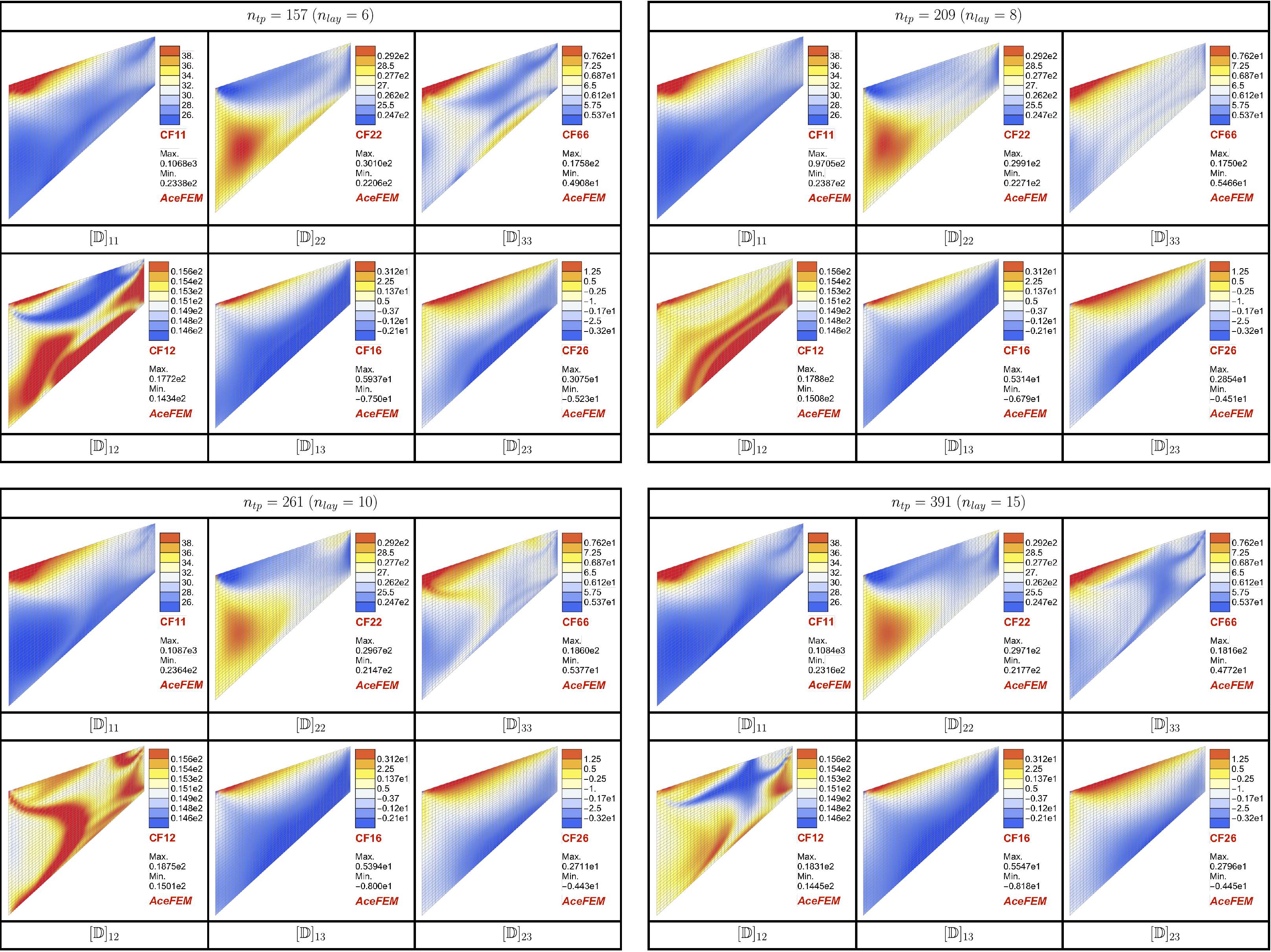}
\caption{Cook's membrane: distribution of components of the material tangent constitutive stiffness tensor $\mathbb{D}$ within the domain obtained with model-data-driven strategies characterized by a training width of $\Delta_{tr}=25\%$ and different densities of sampling points (i.e., different number of layers $n_{lay}$ and training points $n_{tp}$).}
\label{fig:cook_C_comparison_all}
\end{sidewaysfigure}

This Appendix shows additional results obtained with a model-data-driven strategy characterized by different training datasets for the \emph{constant width case} (see Section \ref{par:specs_hybrid}). In detail, for the compression test, the distribution of the components of the second Piola-Kirchhoff stress tensor ${\bf S}$ and of the material tangent constitutive stiffness tensor $\mathbb{D}$ are respectively shown in Figs. \ref{fig:punch_S_comparison_all} and \ref{fig:punch_C_comparison_all}. Analogously, for the Cook's membrane test, the same set of results is shown in Figs. \ref{fig:cook_S_comparison_all} and \ref{fig:cook_C_comparison_all}.

It can be noted that the values and distribution of ${\bf S}$ are practically independent from the number of training points $n_{tp}$. On the other hand, differences in the values of the components of $\mathbb{D}$ can be observed for different densities of training points. This is particularly evident for some components, such as $[\mathbb{D}]_{12}$, and variations are higher for the Cook's membrane problem than for the compression test.

\end{document}